\PassOptionsToPackage{table}{xcolor}
\documentclass[10pt,logo,copyright]{nvidiatechreport}
\usepackage[numbers,sort]{natbib}
\usepackage{subcaption}
\usepackage{adjustbox}
\usepackage{amsmath,amsfonts,bm}

\def\eqref#1{equation~\ref{#1}}

\def\1{\bm{1}}

\DeclareMathAlphabet{\mathsfit}{\encodingdefault}{\sfdefault}{m}{sl}
\SetMathAlphabet{\mathsfit}{bold}{\encodingdefault}{\sfdefault}{bx}{n}

\usepackage{xspace}
\usepackage{tikz}
\usepackage{tabularx}
\usepackage{multirow}
\usepackage{siunitx}
\usepackage{booktabs}
\usepackage{graphicx}
\newlength{\figgroupgap}
\newlength{\figpanelwidth}
\newlength{\figpanelheight}
\usepackage{wrapfig} %
\usepackage{caption}
\usepackage[capitalize]{cleveref}
\newcommand{\inlinesection}[1]{\noindent{\textbf{#1}}}
\newcommand{\MethodName}{LucidLyra\@\xspace}
\newcommand{\MethodNamePlus}{LucidLyra+\@\xspace}
\newcommand{\ProjectPageURL}{https://research.nvidia.com/labs/sil/projects/lucidlyra}
\newcommand{\ProjectPageLink}{%
  \if\relax\detokenize\expandafter{\ProjectPageURL}\relax
    \todo{TODO: public project page URL}%
  \else
    \href{\ProjectPageURL}{\expandafter\nolinkurl\expandafter{\ProjectPageURL}}%
  \fi
}
\newcommand{\todo}[1]{{\textbf{\color{red}#1}}}

\newcolumntype{Z}{S[table-format=2.1]}
\newcolumntype{Y}{>{\centering\arraybackslash}X} %

\renewcommand{\textuparrow}{%
  \raisebox{0.175em}{\hspace{-0.15em}\scalebox{0.75}{\ensuremath{\uparrow}}}%
}
\renewcommand{\textdownarrow}{%
  \raisebox{0.175em}{\hspace{-0.15em}\scalebox{0.75}{\ensuremath{\downarrow}}}%
}

\definecolor{lightgray}{RGB}{230,230,230}
\definecolor{gray}{RGB}{180,180,180}
\definecolor{nv}{RGB}{133,183,55}
\definecolor{my_red}{RGB}{225,49,47}
\definecolor{my_orange}{RGB}{239, 131, 68}
\definecolor{my_blue}{RGB}{6, 25, 245}
\definecolor{my_pink}{RGB}{255,100,169}
\definecolor{my_green}{RGB}{76,166,92}
\definecolor{my_yellow}{RGB}{228, 184, 67}

\definecolor{unlabeled}{RGB}{0,0,0}
\definecolor{wall}{RGB}{120,120,120}
\definecolor{building}{RGB}{180,120,120}
\definecolor{sky}{RGB}{6,230,230}
\definecolor{floor}{RGB}{80,50,50}
\definecolor{tree}{RGB}{4,200,3}
\definecolor{ceiling}{RGB}{120,120,80}
\definecolor{road}{RGB}{140,140,140}
\definecolor{bed}{RGB}{204,5,255}
\definecolor{windowpane}{RGB}{230,230,230}
\definecolor{grass}{RGB}{4,250,7}
\definecolor{cabinet}{RGB}{224,5,255}
\definecolor{sidewalk}{RGB}{235,255,7}
\definecolor{person}{RGB}{150,5,61}
\definecolor{earth}{RGB}{120,120,70}
\definecolor{door}{RGB}{8,255,51}
\definecolor{table}{RGB}{255,6,82}
\definecolor{mountain}{RGB}{143,255,140}
\definecolor{plant}{RGB}{204,255,4}
\definecolor{curtain}{RGB}{255,51,7}
\definecolor{chair}{RGB}{204,70,3}
\definecolor{car}{RGB}{0,102,200}
\definecolor{water}{RGB}{61,230,250}
\definecolor{painting}{RGB}{255,6,51}
\definecolor{sofa}{RGB}{11,102,255}
\definecolor{shelf}{RGB}{255,7,71}
\definecolor{house}{RGB}{255,9,224}
\definecolor{sea}{RGB}{9,7,230}
\definecolor{mirror}{RGB}{220,220,220}
\definecolor{rug}{RGB}{255,9,92}
\definecolor{field}{RGB}{112,9,255}
\definecolor{armchair}{RGB}{8,255,214}
\definecolor{seat}{RGB}{7,255,224}
\definecolor{fence}{RGB}{255,184,6}
\definecolor{desk}{RGB}{10,255,71}
\definecolor{rock}{RGB}{255,41,10}
\definecolor{wardrobe}{RGB}{7,255,255}
\definecolor{lamp}{RGB}{224,255,8}
\definecolor{bathtub}{RGB}{102,8,255}
\definecolor{railing}{RGB}{255,61,6}
\definecolor{cushion}{RGB}{255,194,7}
\definecolor{base}{RGB}{255,122,8}
\definecolor{box}{RGB}{0,255,20}
\definecolor{column}{RGB}{255,8,41}
\definecolor{signboard}{RGB}{255,5,153}
\definecolor{chestofdrawers}{RGB}{6,51,255}
\definecolor{counter}{RGB}{235,12,255}
\definecolor{sand}{RGB}{160,150,20}
\definecolor{sink}{RGB}{0,163,255}
\definecolor{skyscraper}{RGB}{140,140,140}
\definecolor{fireplace}{RGB}{250,10,15}
\definecolor{refrigerator}{RGB}{20,255,0}
\definecolor{grandstand}{RGB}{31,255,0}
\definecolor{path}{RGB}{255,31,0}
\definecolor{stairs}{RGB}{255,224,0}
\definecolor{runway}{RGB}{153,255,0}
\definecolor{case}{RGB}{0,0,255}
\definecolor{pooltable}{RGB}{255,71,0}
\definecolor{pillow}{RGB}{0,235,255}
\definecolor{screendoor}{RGB}{0,173,255}
\definecolor{stairway}{RGB}{31,0,255}
\definecolor{river}{RGB}{11,200,200}
\definecolor{bridge}{RGB}{255,82,0}
\definecolor{bookcase}{RGB}{0,255,245}
\definecolor{blind}{RGB}{0,61,255}
\definecolor{coffeetable}{RGB}{0,255,112}
\definecolor{toilet}{RGB}{0,255,133}
\definecolor{flower}{RGB}{255,0,0}
\definecolor{book}{RGB}{255,163,0}
\definecolor{hill}{RGB}{255,102,0}
\definecolor{bench}{RGB}{194,255,0}
\definecolor{countertop}{RGB}{0,143,255}
\definecolor{stove}{RGB}{51,255,0}
\definecolor{palm}{RGB}{0,82,255}
\definecolor{kitchenisland}{RGB}{0,255,41}
\definecolor{computer}{RGB}{0,255,173}
\definecolor{swivelchair}{RGB}{10,0,255}
\definecolor{boat}{RGB}{173,255,0}
\definecolor{bar}{RGB}{0,255,153}
\definecolor{arcademachine}{RGB}{255,92,0}
\definecolor{hovel}{RGB}{255,0,255}
\definecolor{bus}{RGB}{255,0,245}
\definecolor{towel}{RGB}{255,0,102}
\definecolor{light}{RGB}{255,173,0}
\definecolor{truck}{RGB}{255,0,20}
\definecolor{tower}{RGB}{255,184,184}
\definecolor{chandelier}{RGB}{0,31,255}
\definecolor{awning}{RGB}{0,255,61}
\definecolor{streetlight}{RGB}{0,71,255}
\definecolor{booth}{RGB}{255,0,204}
\definecolor{televisionreceiver}{RGB}{0,255,194}
\definecolor{airplane}{RGB}{0,255,82}
\definecolor{dirttrack}{RGB}{0,10,255}
\definecolor{apparel}{RGB}{0,112,255}
\definecolor{pole}{RGB}{51,0,255}
\definecolor{land}{RGB}{0,194,255}
\definecolor{bannister}{RGB}{0,122,255}
\definecolor{escalator}{RGB}{0,255,163}
\definecolor{ottoman}{RGB}{255,153,0}
\definecolor{bottle}{RGB}{0,255,10}
\definecolor{buffet}{RGB}{255,112,0}
\definecolor{poster}{RGB}{143,255,0}
\definecolor{stage}{RGB}{82,0,255}
\definecolor{van}{RGB}{163,255,0}
\definecolor{ship}{RGB}{255,235,0}
\definecolor{fountain}{RGB}{8,184,170}
\definecolor{conveyerbelt}{RGB}{133,0,255}
\definecolor{canopy}{RGB}{0,255,92}
\definecolor{washer}{RGB}{184,0,255}
\definecolor{plaything}{RGB}{255,0,31}
\definecolor{swimmingpool}{RGB}{0,184,255}
\definecolor{stool}{RGB}{0,214,255}
\definecolor{barrel}{RGB}{255,0,112}
\definecolor{basket}{RGB}{92,255,0}
\definecolor{waterfall}{RGB}{0,224,255}
\definecolor{tent}{RGB}{112,224,255}
\definecolor{bag}{RGB}{70,184,160}
\definecolor{minibike}{RGB}{163,0,255}
\definecolor{cradle}{RGB}{153,0,255}
\definecolor{oven}{RGB}{71,255,0}
\definecolor{ball}{RGB}{255,0,163}
\definecolor{food}{RGB}{255,204,0}
\definecolor{step}{RGB}{255,0,143}
\definecolor{tank}{RGB}{0,255,235}
\definecolor{tradename}{RGB}{133,255,0}
\definecolor{microwave}{RGB}{255,0,235}
\definecolor{pot}{RGB}{245,0,255}
\definecolor{animal}{RGB}{255,0,122}
\definecolor{bicycle}{RGB}{255,245,0}
\definecolor{lake}{RGB}{10,190,212}
\definecolor{dishwasher}{RGB}{214,255,0}
\definecolor{screen}{RGB}{0,204,255}
\definecolor{blanket}{RGB}{20,0,255}
\definecolor{sculpture}{RGB}{255,255,0}
\definecolor{hood}{RGB}{0,153,255}
\definecolor{sconce}{RGB}{0,41,255}
\definecolor{vase}{RGB}{0,255,204}
\definecolor{trafficlight}{RGB}{41,0,255}
\definecolor{tray}{RGB}{41,255,0}
\definecolor{ashcan}{RGB}{173,0,255}
\definecolor{fan}{RGB}{0,245,255}
\definecolor{pier}{RGB}{71,0,255}
\definecolor{crtscreen}{RGB}{122,0,255}
\definecolor{plate}{RGB}{0,255,184}
\definecolor{monitor}{RGB}{0,92,255}
\definecolor{bulletinboard}{RGB}{184,255,0}
\definecolor{shower}{RGB}{0,133,255}
\definecolor{radiator}{RGB}{255,214,0}
\definecolor{glass}{RGB}{25,194,194}
\definecolor{clock}{RGB}{102,255,0}
\definecolor{flag}{RGB}{92,0,255}

\renewcommand{\cf}{cf.\@\xspace}

\titleformat{name=\section,numberless}[block]
  {\Large\bfseries\headingfont\raggedright}
  {}{0pt}{#1}

\title{LucidLyra: Feed-Forward Generative 3D Scene Reconstruction and Joint Understanding via Multi-Task Distillation}
\author{%
  \textbf{Oliver Hahn}\textsuperscript{1,2} \quad
  \textbf{Rodrigo Marcuzzi}\textsuperscript{1} \quad
  \textbf{Sangeetha Grama Srinivasan}\textsuperscript{1} \authorcr
  \textbf{Alessandro Burzio}\textsuperscript{1,3} \quad
  \textbf{Sherwin Bahmani}\textsuperscript{\textdagger} \quad
  \textbf{Jiahui Huang}\textsuperscript{\textdagger} \authorcr
  \textbf{Stefan Roth}\textsuperscript{2,4,5} \quad
  \textbf{Shengyu Huang}\textsuperscript{1} \quad
  \textbf{Qunjie Zhou}\textsuperscript{1} \quad
  \textbf{Laura Leal-Taix\'e}\textsuperscript{1} \\
  {\small \textsuperscript{1}NVIDIA \quad
  \textsuperscript{2}TU Darmstadt} \\
  {\small \textsuperscript{3}Universit\`a degli Studi di Modena e Reggio Emilia} \\
 {\small \textsuperscript{4}ELIZA \quad
  \textsuperscript{5}hessian.AI} \quad
  {\small \textsuperscript{\textdagger}Work performed while at NVIDIA} \\
  \vspace{1em}{\small \ProjectPageLink}
}
\hypersetup{
  pdftitle={LucidLyra: Feed-Forward Generative 3D Scene Reconstruction and Joint Understanding via Multi-Task Distillation},
  pdfauthor={Oliver Hahn, Rodrigo Marcuzzi, Sangeetha Grama Srinivasan, Alessandro Burzio, Sherwin Bahmani, Jiahui Huang, Stefan Roth, Shengyu Huang, Qunjie Zhou, Laura Leal-Taix\'e},
  pdfsubject={LucidLyra preprint}
}

\begin{document}
\maketitle
\begin{center}
  \begingroup%
\newcommand{\GenerationTitle}{Single Image-to-3D Scene Generation}
\newcommand{\ApplicationTitle}{Towards Simulation-Ready Generative 3D Scenes}
\newcommand{\SplitAngle}{20} %
\newcommand{\SplitWidth}{0.25mm}
\newcommand{\BlockBorderWidth}{0.45pt}
\newcommand{\InputFrameWidth}{0.7pt}
\newcommand{\BlockTitleSize}{\small}
\newcommand{\RowLabelSize}{\tiny}
\newcommand{\AppLabelSize}{\tiny}
\newcommand{\TileW}{28.9}
\newcommand{\TileH}{15.9}
\newcommand{\TileGap}{0.35}
\newcommand{\LegendScale}{1.5}
\newcommand{\LegendOpacity}{0.5}
\newcommand{\LabelOpacity}{0.75} %
\definecolor{applicationbg}{HTML}{FFFFFF}
\definecolor{figureink}{HTML}{1A1A1A}
\definecolor{mutedink}{HTML}{555555}
\definecolor{framecolor}{HTML}{DEDEDE}
\ifPDFTeX
  \renewcommand{\sfdefault}{phv}%
  \fontencoding{T1}\selectfont%
\else
  \sffamily
\fi
\newsavebox{\TeaserPanelImage}%

\newcommand{\Tile}[3]{%
  \node[anchor=north west,inner sep=0,outer sep=0] at (#1,#2)
    {\includegraphics[width=\TileW mm,height=\TileH mm]{#3}};
}%
\newcommand{\SplitTile}[4]{%
  \begin{scope}[shift={(#1,#2)}]
    \pgfmathsetmacro{\splitTop}{\TileW/2 + \TileH/2*tan(\SplitAngle)}
    \pgfmathsetmacro{\splitBottom}{\TileW/2 - \TileH/2*tan(\SplitAngle)}
    \begin{scope}
      \clip (0,0)--(\splitTop,0)--(\splitBottom,\TileH)--(0,\TileH)--cycle;
      \Tile{0}{0}{#3}
    \end{scope}
    \begin{scope}
      \clip (\splitTop,0)--(\TileW,0)--(\TileW,\TileH)--(\splitBottom,\TileH)--cycle;
      \Tile{0}{0}{#4}
    \end{scope}
    \begin{scope}
      \clip (0,0) rectangle (\TileW,\TileH);
      \draw[white,line width=\SplitWidth,line cap=butt]
        (\splitTop,0)--(\splitBottom,\TileH);
    \end{scope}
  \end{scope}
}%
\newcommand{\SemanticLegend}[3]{%
  \begin{scope}[shift={(0,\rowC)}]
    \clip (0,0) rectangle (\TileW,\TileH);
    \pgfmathsetmacro{\legendCanvasW}{\TileW*\LegendScale}
    \pgfmathsetmacro{\legendCanvasH}{\TileH*\LegendScale}
    \pgfmathsetmacro{\legendWidth}{(#2-16)/1280*\legendCanvasW}
    \pgfmathsetmacro{\legendHeight}{(688-#3)/704*\legendCanvasH}
    \pgfmathsetmacro{\legendImageX}{0.55-16/1280*\legendCanvasW}
    \pgfmathsetmacro{\legendImageBottom}{\TileH-0.5+16/704*\legendCanvasH}
    \path[fill=white,fill opacity=\LegendOpacity,rounded corners=0.15mm]
      (0.40,{\TileH-0.65-\legendHeight})
      rectangle ({0.70+\legendWidth},{\TileH-0.35});
    \node[anchor=south west,inner sep=0,outer sep=0]
      at (\legendImageX,\legendImageBottom)
      {\includegraphics[width=\legendCanvasW mm,height=\legendCanvasH mm]{#1/semantic_legend.png}};
  \end{scope}
}%
\newcommand{\SceneGrid}[8]{%
  \begin{scope}[shift={(#1,#2)}]
    \pgfmathsetmacro{\colB}{\TileW+\TileGap}
    \pgfmathsetmacro{\colC}{2*(\TileW+\TileGap)}
    \pgfmathsetmacro{\gridWidth}{3*\TileW+2*\TileGap}
    \pgfmathsetmacro{\rowB}{\TileH+\TileGap}
    \pgfmathsetmacro{\rowC}{2*(\TileH+\TileGap)}
    \pgfmathsetmacro{\gridHeight}{3*\TileH+2*\TileGap}
    \begin{scope}
      \clip[rounded corners=0.6mm] (0,0) rectangle (\gridWidth,\gridHeight);
      \Tile{0}{0}{#3/input}
      \Tile{\colB}{0}{#3/traj#8_t120_rgb_recon.jpg}
      \Tile{\colC}{0}{#3/traj#7_t120_rgb_recon.jpg}
      \Tile{0}{\rowB}{#3/traj05_t120_semantic_pca.jpg}
      \SplitTile{\colB}{\rowB}{#3/traj#8_t120_semantic_pca.jpg}{#3/traj#8_t120_instance_pca.jpg}
      \Tile{\colC}{\rowB}{#3/traj#7_t120_instance_pca.jpg}
      \Tile{0}{\rowC}{#3/traj05_t120_open_vocab_seg}
      \SplitTile{\colB}{\rowC}{#3/traj#8_t120_open_vocab_seg}{#3/traj#8_t120_instance_seg}
      \Tile{\colC}{\rowC}{#3/traj#7_t120_instance_seg}
    \end{scope}
    \SemanticLegend{#3}{#5}{#6}
    \ifnum#4=1\relax
    \node[rowlabel] at (0,0) {Input Image};
    \node[rowlabel,anchor=north] at ({\colC-\TileGap/2},0) {3DGS RGB Renderings};
    \node[rowlabel] at (0,\rowB) {3DGS Features (Semantic/Instance)};
    \node[rowlabel] at (0,\rowC) {3DGS Segmentation (Open-Voc./Instance)};
    \fi
    \draw[black,line width=\InputFrameWidth,line join=miter]
      (0,0.6)--(0,\TileH)--(\TileW,\TileH)--(\TileW,0)--(0.6,0)
      .. controls (0.2686,0) and (0,0.2686) .. (0,0.6)--cycle;
  \end{scope}
}%
\newcommand{\ApplicationPanel}[7][]{%
  \begin{scope}[shift={(#2,#3)}]
    \sbox{\TeaserPanelImage}{\includegraphics[width=#4mm,height=#5mm,keepaspectratio]{#7}}%
    \node[inner sep=0,outer sep=0] (applicationbounds) at ({#4/2},{#5/2})
      {\phantom{\usebox{\TeaserPanelImage}}};
    \begin{scope}
      \clip[rounded corners=0.6mm]
        (applicationbounds.south west) rectangle (applicationbounds.north east);
      \node[inner sep=0,outer sep=0] at (applicationbounds.center)
        {\usebox{\TeaserPanelImage}};
      \if\relax\detokenize{#6}\relax\else
        \node[applabel,text height=1.55mm,text depth=0.25mm] at (applicationbounds.north west) {#6};
      \fi
    \end{scope}
    \draw[draw=framecolor,line width=0.22pt,rounded corners=0.6mm,#1]
      (applicationbounds.south west) rectangle (applicationbounds.north east);
  \end{scope}
}%
\newcommand{\ApplicationView}[6]{%
  \ApplicationPanel[draw=none]{#1}{#2}{#3}{#4}{}{#6}
  \begin{scope}[shift={(#1,#2)}]
    \node[applabel,anchor=center] at ({#3/2},0) {#5};
  \end{scope}
}%
\newcommand{\OutlinedSceneLabel}[3]{%
  \foreach \contourangle in {0,22.5,...,337.5}{
    \node[scenelabel,text=white,anchor=#1,
      xshift={0.06mm*cos(\contourangle)},yshift={0.06mm*sin(\contourangle)}]
      at (#2) {#3};
  }
  \node[scenelabel,text=black,anchor=#1] at (#2) {#3};
}%
\newcommand{\FlowArrow}[2]{%
  \draw[figureink,line width=0.8pt,line cap=round]
    (#1,121.819)--({#2-1.3},121.819);
  \path[fill=figureink]
    ({#2-1.4},121.069)--(#2,121.819)--({#2-1.4},122.569)--cycle;
}%

\newcommand{\AssetCrop}[8]{%
  \pgfmathsetmacro{\assetScale}{min(#3/#7,#4/#8)}
  \pgfmathsetmacro{\assetW}{1280*\assetScale}
  \begin{scope}[shift={(#1,#2)}]
    \clip ({(#3-#7*\assetScale)/2},{(#4-#8*\assetScale)/2})
      rectangle ({(#3+#7*\assetScale)/2},{(#4+#8*\assetScale)/2});
    \node[anchor=north west,inner sep=0,outer sep=0]
      at ({(#3-#7*\assetScale)/2-#5*\assetScale},{(#4-#8*\assetScale)/2-#6*\assetScale})
      {\includegraphics[width=\assetW mm]{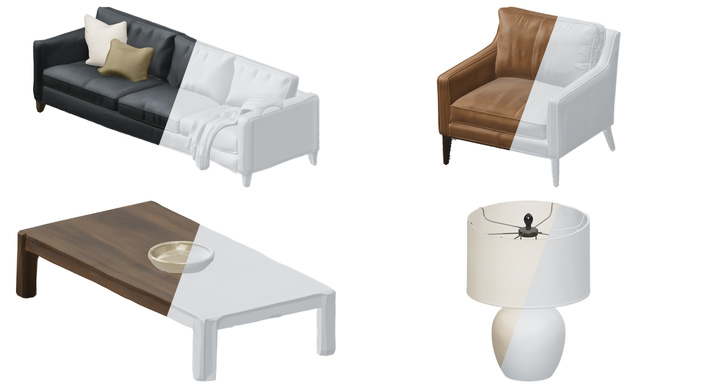}};
  \end{scope}
}%
\resizebox{\linewidth}{!}{%
\begin{tikzpicture}[x=1mm,y=-1mm,
  labelbox/.style={text=figureink,anchor=north west,fill=white,
    fill opacity=\LabelOpacity,text opacity=1,rounded corners=0.25mm,
    inner xsep=0.75mm,inner ysep=0.30mm,outer sep=0},
  rowlabel/.style={labelbox,font=\sffamily\RowLabelSize,
    text height=1.55mm,text depth=0.25mm},
  applabel/.style={labelbox,font=\sffamily\AppLabelSize},
  scenelabel/.style={font=\sffamily\bfseries\BlockTitleSize,align=center,
    inner xsep=0.3mm,inner ysep=0.35mm,outer sep=0},
  sectiontitle/.style={font=\sffamily\mdseries\BlockTitleSize,
    text=black,anchor=center,fill=white,inner xsep=1.8mm,inner ysep=0.25mm,outer sep=0},
  blockframe/.style={draw=black,line width=\BlockBorderWidth,rounded corners=1.2mm}]
  \path[use as bounding box] (0,2.4) rectangle (180,135.675);
  \node[sectiontitle,fill=none] (generationtitle) at (90,4.0) {\GenerationTitle};
  \draw[blockframe] (generationtitle.west)--(0,4.0)--(0,106.2)
    --(180,106.2)--(180,4.0)--(generationtitle.east);
  \SceneGrid{1.8}{6.2}{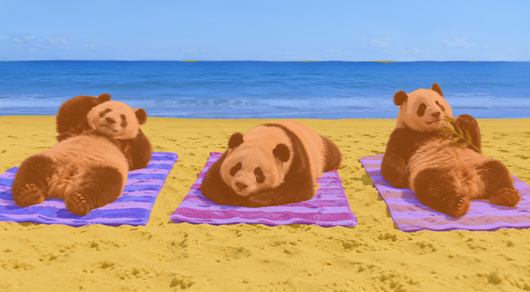}{1}{120}{593}{04}{02}
  \SceneGrid{90.8}{6.2}{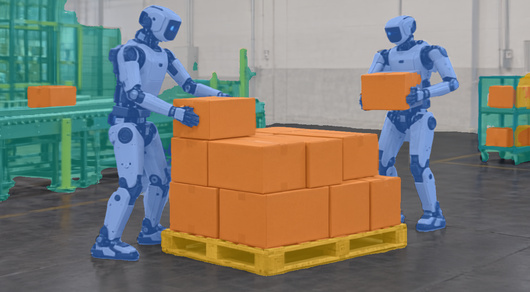}{0}{155}{553}{00}{03}
  \SceneGrid{1.8}{56.0}{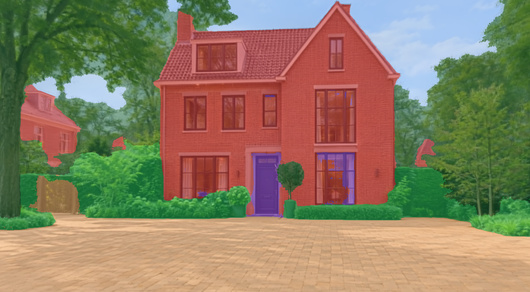}{0}{131}{573}{00}{03}
  \SceneGrid{90.8}{56.0}{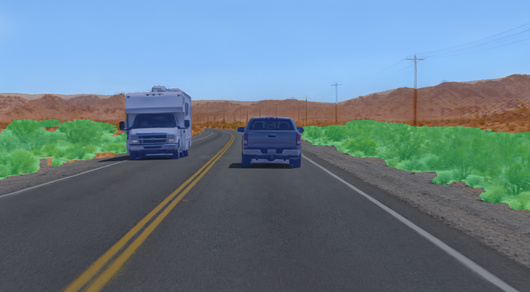}{0}{169}{593}{02}{03}

  \path[blockframe,fill=applicationbg] (0,110.9) rectangle (180,135.675);
  \node[sectiontitle] at (90,110.9) {\ApplicationTitle};
  \ApplicationPanel[draw=black,line width=\InputFrameWidth]{1.8}{115.55}{28.9}{15.9}{Input Image}{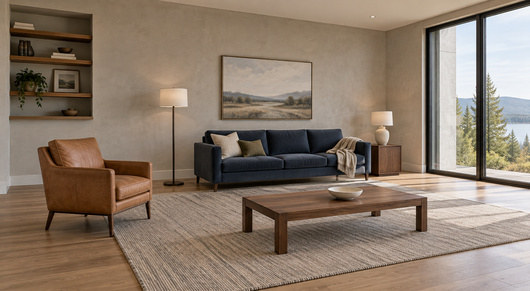}
  \ApplicationPanel[draw=none]{31.1}{112.925}{42}{21.15}{}{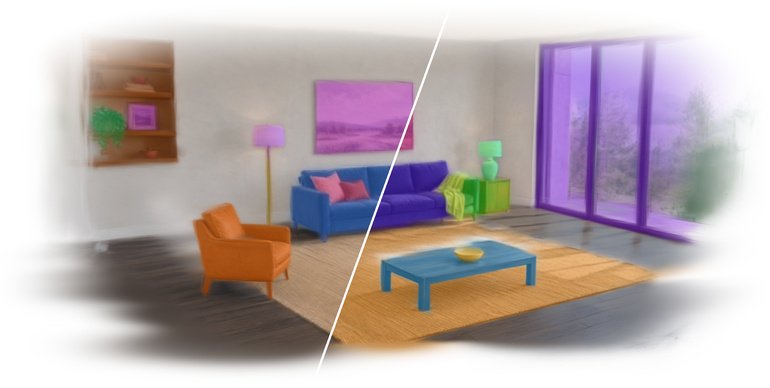}
  \path (applicationbounds.north west)--(applicationbounds.north east)
    coordinate[pos=0.25] (semanticlabel) coordinate[pos=0.75] (instancelabel);
  \node[applabel,anchor=north,align=center] at ([yshift=-1.0mm]semanticlabel) {Open-Voc.\\Segmentation};
  \node[applabel,anchor=north,align=center] at ([yshift=-1.0mm]instancelabel) {Instance\\Segmentation};
  \OutlinedSceneLabel{south}{[yshift=-0.7mm]applicationbounds.south}{LucidLyra 3DGS Scene}
  \ApplicationPanel[draw=none]{74.2}{112.7}{38.5}{21.175}{}{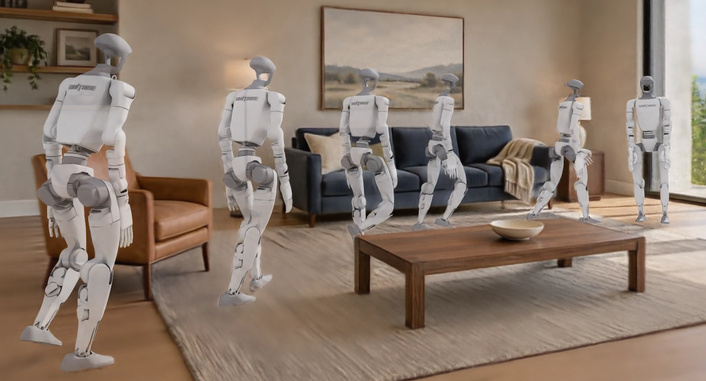}
  \AssetCrop{113.5}{112.7}{16}{10.4}{20}{0}{595}{345}
  \AssetCrop{129.8}{112.7}{9.1}{10.4}{785}{0}{360}{345}
  \AssetCrop{113.5}{123.4}{16}{10.475}{20}{355}{600}{349}
  \AssetCrop{129.8}{123.4}{9.1}{10.475}{835}{355}{260}{349}
  \ApplicationPanel[draw=none]{139.7}{112.7}{38.5}{21.175}{}{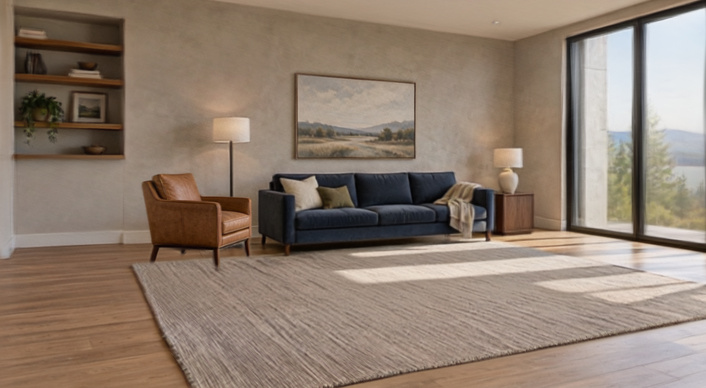}
  \node[applabel,anchor=center] at (93.45,133.875) {\textbf{Robot Simulation}};
  \node[applabel,anchor=center] at (126.2,133.875) {\textbf{3D Asset Mining}};
  \node[applabel,anchor=center] at (158.95,133.875) {\textbf{3D Editing}};
  \begin{scope}[shift={(0,1.681)}]
    \FlowArrow{31.0}{33.7}
    \FlowArrow{71.0}{73.7}
  \end{scope}
\end{tikzpicture}%
}%
\endgroup%

  \captionof{figure}{
    \textbf{Feed-forward semantic- and instance-aware image-to-3D scene generation.}
    \emph{Top:} \MethodName infers a unified 3D Gaussian Splatting (3DGS) representation consisting of appearance, geometry, semantic, and instance features from a single image. The learned feature fields enable open-vocabulary semantic segmentation and instance segmentation. \emph{Bottom:} \MethodName enables downstream applications such as physical-AI simulation, editing, and 3D asset mining.
  }
  \label{fig:teaser}
\end{center}
\clearpage
\begin{abstract}
We present \MethodName, a framework that makes generatively reconstructed 3D scenes directly queryable for semantics and instances in a feed-forward manner from video latents.
While generative 3D reconstruction can synthesize virtual environments from a single image, downstream applications in robotics and physical AI require not only visual realism but also access to scene semantics and object identities.
Existing 3D understanding approaches, however, require captured multi-view scenes, have limited generalization capabilities, or rely on expensive per-scene processing and optimization.
To bridge this gap, our method distills complementary knowledge from a camera-controlled video diffusion model and frozen 2D vision foundation models into a unified model to jointly infer 3D Gaussians encoding geometry, appearance, open-vocabulary semantics, and class-agnostic instance cues.
The resulting scenes support novel-view rendering, open-vocabulary segmentation, and object grouping without per-scene optimization, while training requires neither manual 3D annotations nor captured multi-view data.
Our method outperforms baselines combining state-of-the-art 3DGS segmentation approaches with generative 3D scene reconstruction on curated synthetic and real-image-conditioned benchmarks, while preserving reconstruction quality.
Furthermore, our latent formulation naturally extends to monocular video, reconstructing structured dynamic 3D scenes across time.
By turning renderable but unstructured outputs into queryable 3D worlds, \MethodName enables downstream applications including 3D scene editing, 3D asset mining, and robotic simulation.
\end{abstract}

\abscontent

\section{Introduction}

Recent generative reconstruction methods can turn a single image into a renderable 3D environment. Building upon learned generative models, they can imagine content beyond the observed field of view and consolidate it into a scene that can be viewed from new perspectives \citep{gao:2024:ccai,ren:2025:3wvg,liang:2025:wn3s,
szymanowicz:2025:bg3s,schwarz:2025:ggsg,bahmani:2026:g3sr}.
Generated 3D environments are promising for scalable content generation in interactive media, physical AI simulation, and industrial applications in engineering and manufacturing \citep{he:2026:aso3}.
However, existing generated scene representations primarily capture geometry and appearance, while lacking an explicit understanding of their content. Downstream applications in embodied AI and physical simulation cannot interact with an collection of colored primitives; they require structured scene elements with semantic roles and discrete instance boundaries to support collision reasoning, physical manipulation, and interactive scene editing.

A natural solution might seem to generate an appearance-only scene first and recover semantic and instance representations post-hoc.
However, this two-stage route breaks down in the single-view generative setting. Existing post-hoc methods either attach 2D foundation-model predictions to an existing reconstruction through per-scene optimization~\citep{qin:2024:3lgs,ye:2024:ggsa,tian:2025:cclf,xiong:2026:sfs}, or jointly infer geometry and understanding from multiple captured views~\citep{zhu:2026:eff3,jiang:2026:iaf3,tian:2026:br3s}.
The former requires expensive scene-specific foundation model preprocessing and optimization.
The latter relies on captured multi-view inputs and is constrained by the limited scale and domain diversity of multi-view training data.
Neither route directly yields a structured 3D scene that has been generated from a single image.

Scaling these capabilities is not merely a matter of attaching a feature head to a reconstruction model. Generated regions lack paired ground-truth supervision, so estimated geometry, semantics, and object boundaries can drift and disagree across views. Predictions from 2D vision models compound this, as their semantics and masks are not guaranteed to be view-consistent. The central challenge is to turn complementary but imperfect generative and perceptual signals
into a consistent 3D representation while balancing geometry, appearance, semantics, and instances without compromising generation quality.
An integrated model should let semantic priors and geometry reinforce each other, rather than tagging an isolated reconstruction with semantic or object information.

We introduce \MethodName, a framework that unifies feed-forward generative 3D reconstruction and scene understanding in an explicit Gaussian representation (\cref{fig:teaser}).
Our central insight is that a video generator provides a unified foundation for structured 3D scene generation, serving both as the shared backbone and as the source of training data.
Building on Lyra's latent-to-Gaussian formulation~\citep{bahmani:2026:g3sr}, \MethodName learns a 3D scene representation from camera-controlled video latents through multi-task distillation. Given a single input image, we denoise the corresponding video latents across camera trajectories, and process them through two parallel branches. The teacher branch decodes the latent into an RGB video and applies frozen vision-language, segmentation, and tracking foundation models to obtain semantic and instance supervision. The student branch incorporates vision cues of the input view into the video latents and directly maps the shared latent into a 3D Gaussian scene. We consolidate the teacher predictions across the generated views and distill them into persistent semantic and instance fields on the Gaussian primitives, jointly with geometry and appearance.
A single shared representation can therefore support novel-view rendering,
open-vocabulary segmentation, and object grouping.
As an initial exploration toward generative 4D scene
understanding, we also extend the formulation to monocular video inputs and predict
semantics- and instance-aware dynamic 3D scenes.
For this novel task, we introduce an evaluation protocol with benchmarks covering static scenes, real-image domain transfer, and dynamic scenes.

On our static LucidLyra-SDG benchmark, \MethodName improves semantic mIoU and
class-agnostic instance mTrackPR over the strongest comparable baseline by \num{31.7} and
\num{9.8} percentage points,
while maintaining reconstruction fidelity, with a modest PSNR gain of \num{0.53}\,dB over Lyra.
More broadly, this formulation turns pretrained video generation into a
scalable source of supervision for structured 3D learning, opening a path from
renderable but unstructured outputs to queryable 3D worlds that downstream
systems can reason about and interact with. In summary, our contributions are:
\begin{itemize}
    \item We introduce a unified generative 3D model that produces an explicit
    Gaussian scene representation with geometry, appearance, open-vocabulary semantic, and class-agnostic instance features from a single image in a feed-forward fashion.

    \item We develop a multi-task distillation strategy that transfers
    2D generative, semantic, and object-level knowledge through aligned generated
    views into a consistent 3D representation. This enables task-specific training
    without captured multi-view data or manual 3D annotations, and avoids
    per-scene optimization at inference time.

    \item We jointly evaluate reconstruction and understanding on static and dynamic scenes. \MethodName consistently outperforms two-stage baselines across optimization-based, mask-lifting, and feed-forward 3DGS segmentation methods, while maintaining novel-view reconstruction fidelity on static scenes with a small reconstruction trade-off on dynamic scenes. We further demonstrate downstream applications including robotic simulation, 3D scene editing, and 3D asset mining.
\end{itemize}

\section{Related Work}

\noindent\textbf{From observed reconstruction to generative scene construction.}
Neural radiance fields~\citep{mildenhall:2020:nrsa} and 3D Gaussian splatting (3DGS)~\citep{kerbl:2023:3gsf} fit per-scene representations, while feed-forward methods amortize reconstruction across scenes to predict 3D Gaussians for novel-view synthesis~\citep{charatan:2024:p3gs,zhang:2024:glrm} or estimate dense depth and camera parameters~\citep{Wang:2025:vggt,lin:2025:da3r,burzio:2026:dejaview}. Dynamic formulations underwent a similar evolution from per-sequence fitting~\citep{yang:2024:d3gf,yang:2024:rpds} to feed-forward estimation~\citep{liang:2025:fbro,xu:2025:4la4,kim:2026:lgmw}, yet all remain bounded by observed multi-view coverage.
Generative methods relax this constraint by synthesizing unobserved content. Early works distill diffusion priors into per-scene optimizations or iterative view inpaintings~\citep{poole:2023:dtu2,hollein:2023:tett}. Multi-view generators synthesize dense video trajectories for downstream reconstruction~\citep{gao:2024:ccai,schneider:2025:wtn3}, with later extensions incorporating panoramic scaffolds, geometric alignment, or long-horizon expansion~\citep{wang:2025:vsdv,wang:2026:o2sg,hollein:2026:wrfi,shen:2026:eg3w,lin:2025:da3r,tian:2026:srgs}. However, generation and 3D reconstruction remain decoupled into different stages.

A closer line to our setting couples generation directly with feed-forward 3D prediction. Recent methods decode video latents into Gaussians~\citep{liang:2025:wn3s}, embed Gaussian feature fields within video diffusion~\citep{schwarz:2025:ggsg}, directly generate explicit 3D representations~\citep{szymanowicz:2025:bg3s,shabanov:2026:frgn}, or align generative image, video, and 3D latent spaces~\citep{li:2026:fwhq,huang:2026:g3rf,gao:2026:pu3s}.
Lyra~\citep{bahmani:2026:g3sr} self-distills a feed-forward Gaussian decoder from a camera-controlled video foundation model~\citep{ren:2025:3wvg,cosmos:2025:cwfm}, using synthetic multi-view video trajectories as both the latent input and the supervision target for static and dynamic 3D synthesis. We build upon Lyra's scalable, latent-to-Gaussian formulation, extending its feed-forward output from raw geometry and appearance to joint language-grounded semantics and class-agnostic instances.

\noindent\textbf{From per-scene optimization to amortized 3D scene understanding.}
Early works predict semantics and instances from point clouds~\citep{qi:2017:pdlo,qi:2017:pdhf,choy:2019:4scm,jiang:2020:pdpp}, while subsequent methods lift open-vocabulary 2D foundation features onto reconstructed geometry~\citep{peng:2023:o3su,jatavallabhula:2023:com3,takmaz:2023:oo3i} or embed language fields into neural representations~\citep{zhi:2021:isla,kundu:2022:pnfa,siddiqui:2023:plf3,kerr:2023:ller}.
For 3DGS, several approaches optimize per-Gaussian language fields, aggregate 2D foundation features, or lift multi-view masks into object-aware representations~\citep{li:2025:hd3l,xiong:2026:sfs,zhu:2025:rete,qin:2024:3lgs,tian:2025:cclf,kim:2025:dsdr,cheng:2025:olae,ye:2024:ggsa,wu:2024:otp3,shen:2024:f2t3,xie:2025:epgs}.
Feed-forward methods amortize understanding across scenes, either by encoding pre-existing 3DGS reconstructions~\citep{li:2026:mtph} or by jointly predicting geometry alongside semantic, panoptic, or instance structure from multi-view inputs~\citep{li:2025:iggt,zhu:2026:eff3,jiang:2026:iaf3,tian:2026:br3s,li:2025:gssu,li:2026:fast,koch:2026:ustf,li:2025:sf3s,sheng:2025:ses3,sun:2026:uu3r,liu:2026:f3lg,yoo:2026:saon,lee:2026:eofs,zhang:2026:ssgs}.
Generative understanding approaches synthesize observations for language-aware reconstruction, modular scene assembly, or driving-specific world models~\citep{liu:2025:rg3l,huang:2025:vpmf,yin:2026:ebiv,deng:2026:3gdw}. Yet, these formulations remain confined to domain-specific priors or assume multiple posed observations.

Existing approaches follow two main paradigms: augmenting a scene-specific reconstruction or jointly inferring geometry and segmentation from multi-view inputs.
The former requires a pre-existing reconstruction, precluding single-view generation, while post-hoc lifting decouples semantics from geometry, leaving fusion vulnerable to cross-view inconsistencies. Crucially, both rely on an expensive preprocessing stage with a 2D vision foundation model.
Feed-forward approaches jointly predict geometry and segmentation, but train on captured multi-view datasets with narrow domain coverage limiting generalization.
\MethodName instead unifies scene understanding with generative reconstruction. Conditioned on a single image, its amortized latent-to-Gaussian decoder jointly predicts geometry, appearance, language-aligned semantic features, and class-agnostic instance features in a single feed-forward pass, without prior 3D reconstructions, per-scene optimization, or post-hoc lifting. Distilling frozen foundation models on generated video eliminates the need for captured multi-view data or manual 3D annotations.

\section{Semantic- and Instance-Aware Generative 3D Reconstruction}
\label{sec:method}

\begin{figure}[t]
    \centering
    \includegraphics[width=\linewidth]{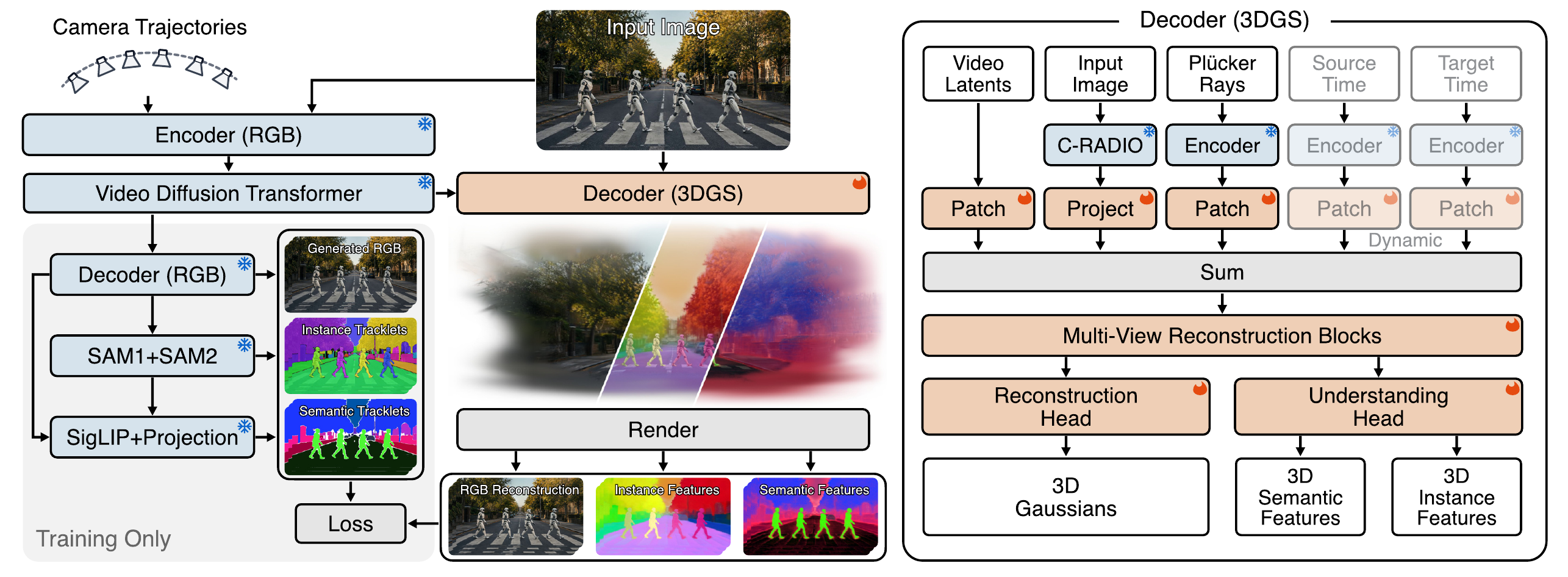}
    \caption{\textbf{\MethodName framework.}
    A frozen camera-controlled video generation model produces multi-view latents. Its RGB
    decoder and frozen 2D foundation models construct reconstruction, semantic,
    and instance targets. The targets supervise our Gaussian decoder which maps the latents, camera
    rays, and input image features to a shared queryable 3D scene. The
    right panel details our 3DGS decoder. Source- and target-time inputs (grayed out) are
    used only for the dynamic extension. Snowflakes and flames denote frozen
    and trainable components, respectively.}
    \label{fig:method-overview}
\end{figure}

Given an input image \(\mathbf I_0\), \MethodName generates a queryable 3D Gaussian Splatting (3DGS)~\citep{kerbl:2023:3gsf} scene representation whose primitives jointly encode geometry, appearance, language-aligned semantics, and class-agnostic instance identity. Rather than supervising reconstruction alone as in Lyra~\citep{bahmani:2026:g3sr}, we use the multi-view outputs of a camera-controlled video generator to construct aligned 2D semantic and instance targets using frozen vision foundation models and distill them into an explicit 3D representation via an amortized feed-forward decoder (\cref{fig:method-overview}).

\subsection{Preliminaries: Lyra Generative 3D Reconstruction}
Lyra~\citep{bahmani:2026:g3sr} amortizes 3D reconstruction by mapping camera-controlled video latents directly into 3D Gaussians. Given an image \(\mathbf I_0\) and \(V\) camera trajectories \(\mathcal C=(\mathcal C^{(1)},\dots,\mathcal C^{(V)})\), a video diffusion model \(\mathcal V\) generates multi-view latents \(\mathbf Z=(\mathbf Z^{(1)},\dots,\mathbf Z^{(V)})\), while camera rays are mapped to Pl\"ucker encodings \(\mathbf E\equiv\mathbf E(\mathcal C)\). A feed-forward decoder \(D^{\text{Lyra}}_{\mathrm{3DGS}}\) predicts a 3DGS scene representation from these inputs:
\begin{equation}
    \mathcal G = D^{\text{Lyra}}_{\mathrm{3DGS}}(\mathbf Z,\mathbf E) = \{g_i\}_{i=1}^{N},\qquad
    g_i = (\boldsymbol\mu_i,\mathbf s_i,\mathbf q_i,\alpha_i,\mathbf c_i),
    \label{eq:Lyra-gaussians}
\end{equation}
where each of $N$ Gaussian primitives \(g_i\) parameterizes position \(\boldsymbol\mu_i\), scale \(\mathbf s_i\), rotation quaternion \(\mathbf q_i\), opacity \(\alpha_i\), and color \(\mathbf c_i\), renderable via differentiable splatting \(\mathcal R\). The predicted Gaussians are aligned to the source-ray grid of the generated views across all trajectories.

To learn this mapping without 3D supervision, Lyra compares two decodings of \(\mathbf Z\): for target camera \(\mathcal C_k\), a frozen VAE decoder outputs a teacher image \(\mathbf I_k = D_{\mathrm{RGB}}(\mathbf Z_k)\), supervising the rendered prediction \(\hat{\mathbf I}_k = \mathcal R(\mathcal G,\mathcal C_k)\) alongside depth targets. At inference time, sampled latents \(\mathbf Z\) yield Gaussians encoding only geometry and appearance.

\subsection{Semantic and Instance Gaussian Decoder}
\label{ssec:method-architecture}
To equip Gaussians with semantic and instance awareness, we introduce additional conditioning using vision features of the input image and dedicated output heads. First, we supplement the video model latents \(\mathbf Z\) with dense input image features \(\mathbf F_0\) extracted with C-RADIOv3~\citep{ranzinger:2024:avfm,heinrich:2025:ibav,ranzinger:2025:yvfs}. 
 We align $\mathbf{F}_0$ with $\mathbf{Z}$ through a learned linear projection to the latent feature dimension. Spatial interpolation matches $\mathbf{F}_0$ to the latent token grid, and broadcasting replicates it across trajectories and latent time. Each latent-time slice is then modulated by a learned scalar weight shared across channels and a learned global gate before being added to the video and ray tokens.
Next, the 3DGS decoder's shared trunk fuses the multi-view tokens into features
\(\mathbf H=\{\mathbf h_i\}_{i=1}^N\), where each feature corresponds to a source ray on the multi-view video latent grid.
Two parallel heads map these shared features to complementary Gaussian attributes:
the reconstruction head $D_{\mathrm{rec}}$ predicts geometry and appearance parameters \(g_i\),
while the understanding head $D_{\mathrm{seg}}$ jointly predicts a semantic descriptor
\(\mathbf f_i^{\mathrm{sem}}\in\mathbb R^{128}\),
an object-level instance embedding
\(\mathbf e_i^{\mathrm{inst}}\in\mathbb R^{12}\),
and a part-level embedding
\(\mathbf e_i^{\mathrm{part}}\in\mathbb R^{4}\):
\begin{equation}
    \widetilde{\mathcal G} = D_{\mathrm{3DGS}}(\mathbf Z,\mathbf E,\mathbf F_0) = \{\tilde g_i\}_{i=1}^{N}, \qquad
    \tilde g_i = [D_{\mathrm{rec}}(\mathbf h_i) \parallel D_{\mathrm{seg}}(\mathbf h_i)].
    \label{eq:lucidlyra-gaussians}
\end{equation}
The shared 3DGS decoder trunk and each task-specific head consist of seven Mamba-2~\citep{dao:2024:tasg} layers followed by one Transformer~\citep{vaswani:20217:aiay} layer.
Operating synchronously on the shared token grid attaches understanding fields directly to the co-located Gaussian \(g_i\), bypassing post-hoc 2D-to-3D association.
For camera \(\mathcal C_k\), a differentiable renderer \(\mathcal R(\widetilde{\mathcal G},\mathcal C_k)\) outputs rendered RGB \(\hat{\mathbf I}_k\), semantic maps \(\hat{\mathbf F}_k^{\mathrm{sem}}\), and instance maps \(\hat{\mathbf F}_k^{\mathrm{inst}}\) via volumetric \(\alpha\)-blending, and depth \(\hat{\mathbf D}_k\) by accumulating primitive center depths.

\subsection{Multi-Task 2D Feature Distillation}
\label{ssec:method-supervision}
\label{ssec:method-training}
We supervise the understanding fields using decoded video frames \(\mathbf I_k\) and frozen 2D vision foundation models---class-agnostic segmentation, video tracking, and vision-language foundation models---generalizing RGB self-distillation to multi-task 2D feature distillation without 3D labels. To resolve cross-view feature jitter and mask inconsistency across imagined views, we first establish cross-view object tracks to associate observations and pool coherent semantic targets.

\inlinesection{Aligned instance targets.}
To associate objects across trajectories, we run SAM's automatic mask generator on the shared first frame \(\mathbf I_0\), filter redundancies by coverage, and use the surviving masks to seed independent SAM2 propagations along each trajectory~\citep{kirillov:2023:sam,ravi:2025:saiv}. The resulting class-agnostic instance tracklets \(Y_k^{\mathrm{inst}}\) preserve trajectory visibility while linking observations from the same initial object. Regions disoccluded late in a trajectory lacking a first-frame seed are excluded from instance supervision.

\inlinesection{Track-consistent semantic targets.}
The instance tracklets also let us aggregate semantic evidence across views.
We use RADSeg~\citep{alama:2026:upce} to extract dense features aligned with the
SigLIP2~\citep{tschannen:2025:mvle} image--text space for every decoded frame \(\mathbf I_k\). 
Storing its native
\num{1152}-dimensional descriptor on every Gaussian is computationally prohibitive, therefore we compress this space with a projection network
\(P_{\mathrm{VL}}:\mathbb R^{1152}\!\rightarrow\mathbb R^{128}\) for both image and text embeddings. 
The projector is trained over frozen SigLIP2 image and text embeddings, using the vanilla SigLIP training objective~\citep{zhai:2024:slli} on the LAION-2B~\citep{schuhmann:2022:laion} dataset and remains frozen afterwards.
Within each camera trajectory, we pool projected features over all pixels and frames
belonging to each SAM track and assign the resulting centroid to every track
pixel. This yields semantic targets \(\mathbf Y_k^{\mathrm{sem}}\), where track pooling suppresses view-dependent feature jitter.

\inlinesection{Joint multi-task objective.}
We initialize the shared trunk and reconstruction head \(D_{\mathrm{rec}}\) from pretrained Lyra and initialize \(D_{\mathrm{seg}}\) from scratch, training the entire Gaussian decoder while keeping all 2D teacher models and \(P_{\mathrm{VL}}\) frozen.
At target camera \(\mathcal C_k\), rendered predictions \((\hat{\mathbf I}_k, \hat{\mathbf D}_k, \hat{\mathbf F}_k^{\mathrm{sem}}, \hat{\mathbf F}_k^{\mathrm{inst}})\) are supervised against targets \((\mathbf I_k, \mathbf D_k, \mathbf Y_k^{\mathrm{sem}}, Y_k^{\mathrm{inst}})\) via
\begin{equation}
    \mathcal L = \mathcal L_{\mathrm{rec}} + \mathcal L_{\mathrm{sem}} + \lambda_{\mathrm{inst}}\mathcal L_{\mathrm{inst}}.
    \label{eq:full-training-objective}
\end{equation}
\(\mathcal L_{\mathrm{rec}}\) mirrors Lyra, combining photometric \(\ell_2\), LPIPS~\citep{zhang:2018:tueo}, depth, and opacity losses to preserve geometry during feature grounding. Schedules and hyperparameters are detailed in \Cref{sec:supp-implementation}.

\inlinesection{Instance distillation.} We learn a class-agnostic instance representation via supervised contrastive learning~\citep{khosla:2020:scl} over tracked pixels across views. For a sample \(u \in \mathcal S\) at pixel \(p_u\) in view \(k_u\) with normalized embedding \(\bar{\mathbf e}_u = \operatorname{norm}_2(\hat{\mathbf F}_{k_u}^{\mathrm{inst}}(p_u))\), we define positive matches sharing track ID as \(\mathcal P(u)=\{v \in \mathcal S\setminus\{u\} \mid Y_{k_v}^{\mathrm{inst}}(p_v) = Y_{k_u}^{\mathrm{inst}}(p_u)\}\). Over valid anchors \(\mathcal U=\{u \in \mathcal S \mid |\mathcal P(u)| > 0\}\), we optimize
\begin{equation}
    \mathcal L_{\mathrm{inst}}
    = -\frac{1}{|\mathcal U|}\sum_{u\in\mathcal U}
    \frac{1}{|\mathcal P(u)|}\sum_{v\in\mathcal P(u)}
    \log\frac{\exp(\bar{\mathbf e}_u^\top\bar{\mathbf e}_v / \tau_{\mathrm{inst}})}
    {\sum_{a\in\mathcal S\setminus\{u\}} \exp(\bar{\mathbf e}_u^\top\bar{\mathbf e}_a / \tau_{\mathrm{inst}})},
    \label{eq:instance-supcon}
\end{equation}
with temperature \(\tau_{\mathrm{inst}} > 0\). This pulls rendered embeddings of the same tracked object together across views while repelling different objects, clustering the underlying 3D Gaussians into distinct instance fields. For part-level supervision, we apply the same objective in parallel to the part embeddings, defining positive and negative pairs from sub-object tracklet IDs within each object tracklet.

\inlinesection{Semantic distillation.}
We supervise rendered semantic features with pooled teacher descriptors. For rendered feature \(\hat{\mathbf f}\equiv\hat{\mathbf F}_k^{\mathrm{sem}}(p)\) and target \(\mathbf y\equiv\mathbf Y_k^{\mathrm{sem}}(p)\) with \(\ell_2\)-normalized counterparts \(\bar{\hat{\mathbf f}}\) and \(\bar{\mathbf y}\), we define the pointwise loss
\begin{equation}
    \ell_{\mathrm{sem}}(\hat{\mathbf f},\mathbf y)
    =-\frac{1}{2}\log\!\left(
      \frac{1+\bar{\hat{\mathbf f}}^\top\bar{\mathbf y}}{2}
      +\epsilon_{\ell}\right)
    +\frac{1}{2d_{\mathrm{sem}}}
      \left\lVert\bar{\hat{\mathbf f}}-\bar{\mathbf y}\right\rVert_1,
    \label{eq:semantic-point-loss}
\end{equation}
with \(d_{\mathrm{sem}}=\num{128}\). The cosine term aligns predictions with the SigLIP2 image--text space, while the \(\ell_1\) term refines coordinate-level precision. Averaging over rendered pixels yields \(\mathcal L_{\mathrm{sem}}^{\mathrm{2D}}\).
Rendered feature losses constrain a blend of several Gaussian features, rather than each primitive individually. We therefore additionally supervise each primitive directly: each Gaussian \(\tilde g_i\) originates from an input ray \((k_i, p_i)\), allowing us to penalize discrepancy against teacher target \(\mathbf Y_{k_i}^{\mathrm{sem}}(p_i)\) as \(\mathcal L_{\mathrm{sem}}^{\mathrm{G}}\), yielding the total semantic loss \(\mathcal L_{\mathrm{sem}} = \lambda_{\mathrm{sem}}^{\mathrm{2D}}\mathcal L_{\mathrm{sem}}^{\mathrm{2D}} + \lambda_{\mathrm{sem}}^{\mathrm{G}}\mathcal L_{\mathrm{sem}}^{\mathrm{G}}\).

\subsection{Inference: Querying the Generated Scene}
\label{ssec:method-inference}
Given generated video latents, our Gaussian decoder predicts the scene
\(\widetilde{\mathcal G}\) in a single forward pass.
Its per-Gaussian semantic and instance features enable open-vocabulary
queries and object grouping directly in 3D, without using the
distillation teachers or lifting predictions from 2D views.

\inlinesection{Open-vocabulary semantics.}
Given categorical text queries \(\mathcal Q=\{s_c\}_{c=1}^{C}\), the frozen SigLIP2 text encoder \(\Phi_{\mathrm T}\) and projection network \(P_{\mathrm{VL}}\) yield normalized embeddings \(\bar{\mathbf t}_c = \operatorname{norm}_2(P_{\mathrm{VL}}(\Phi_{\mathrm T}(s_c)))\). Each Gaussian \(\tilde g_i\) is assigned to the category \(c^*\) that maximizes the cosine similarity with its primitive descriptor:
\begin{equation}
    c^\ast = \arg\max_{c \in \{1,\dots,C\}} \operatorname{norm}_2(\mathbf f_i^{\mathrm{sem}})^\top \bar{\mathbf t}_c.
    \label{eq:semantic-inference}
\end{equation}
Querying new vocabularies needs only text--Gaussian dot products without regeneration.

\inlinesection{Instance grouping.}
Within each predicted semantic class, we spatially smooth instance embeddings and cluster high-opacity Gaussians via HDBSCAN~\citep{campello:2013:dbcb}. Unclustered Gaussians inherit instance IDs from their nearest neighbor in joint embedding and 3D space. For the part-level instance segmentation, we apply the same technique within each object cluster (\cf \cref{ssec:supp-inference}).

\inlinesection{\MethodNamePlus.} Our optional semantic refinement leaves geometry and appearance unchanged. It uses a single RADSeg+ open-vocabulary prediction of $\mathbf{I}_0$ as a 2D anchor, obtained at minimal overhead since RADSeg+ uses the same C-RADIO features.
Our ray-aligned decoder grid links each anchor pixel to its corresponding Gaussian. For every well-supported anchor, we average the semantic features of its Gaussians into a prototype, and relabel any Gaussian whose feature is close enough to that prototype. All remaining Gaussians keep their original semantic labels. We then apply the same instance grouping.

\subsection{Extension to Dynamic Scenes}
\label{ssec:method-dynamic}
For monocular video inputs, we extend \MethodName to dynamic scenes by independently decoding a time-conditioned 3DGS state following Lyra's bullet-time formulation~\citep{bahmani:2026:g3sr,liang:2025:fbro}. The same video model provides multi-view latents preserving input motion; the decoder predicts the full scene state at time \(\tau\), conditioned on source/target time encodings and query-time C-RADIO input-frame features \(\mathbf F_\tau\):
\begin{equation}
    \widetilde{\mathcal G}_{\tau} = D_{\mathrm{3DGS}}^{\mathrm{dyn}}(\mathbf Z,\mathbf E,\mathbf F_\tau, \mathbf T^{\mathrm{src}},\mathbf T_{\tau}^{\mathrm{tgt}}).
\end{equation}

Initialized from the static model, the dynamic decoder trains with the same multi-task losses on \(\tau\)-synchronized views. 
Dynamic scenes can only be supervised at the exact queried time, due to the changing scene.
Following \citet{bahmani:2026:g3sr}, we create six additional trajectories by processing the time-reversed input video and flipping it back. Each target frame is then paired with a same-time observation from a different viewpoint increasing multi-view supervision.
Since each query time yields an independently decoded Gaussian state, we implement temporal instance tracking via an online prototype codebook inspired by IGGT4D~\citep{zou:2026:is4i}.
At each query time, we represent each cluster by its normalized mean object
embedding and greedily match cluster prototypes to codebook entries one-to-one
above a cosine-similarity threshold. Matched clusters inherit persistent IDs
and update the stored prototypes through averaging weighted by Gaussian count,
while unmatched clusters initialize new IDs.

\section{Experiments}
\label{sec:experiments}
We evaluate \MethodName's joint generative 3D reconstruction and scene understanding on synthetic and real-image-conditioned scenes, followed by an analysis of its key components. Dynamic experiments, additional quantitative and qualitative results, downstream applications,
and implementation details are provided in the appendix.

\subsection{Experimental Setup}

\inlinesection{Datasets.}
We augment Lyra-SDG~\citep{bahmani:2026:g3sr} with semantic and instance teacher targets for training (\cref{ssec:method-supervision}) and annotations for evaluation. Lyra-SDG uses GEN3C~\citep{ren:2025:3wvg} to expand images and videos generated from diverse text prompts into multiple camera trajectories per scene. Its \num{59031} static and \num{7378} dynamic scenes contain RGB videos, camera poses, depth maps, and video latents. We preprocess teacher targets for all \num{53} million frames.
A single image admits multiple plausible 3D completions. We evaluate agreement with fixed generated trajectories and their annotations.
Our three curated \num{25}-scene benchmarks use propagated reference annotations with human review and refinement: \emph{LucidLyra-SDG} and \emph{LucidLyra-SDG Dynamic} cover static and dynamic generated scenes, while \emph{LucidLyra-ADE20K} uses real ADE20K validation images~\citep{zhou:2019:sust} to condition generated completions. All data and benchmarks will be released.

For qualitative results, we generate input images with GPT Image or FLUX \citep{openai_image_generation, flux2024}. For dynamic examples, we extend these to videos using Cosmos~3~\citep{nvidia2026cosmos3}. We find that our method is agnostic to the choice of image-generation model.

\inlinesection{Evaluation.}
Following Lyra~\citep{bahmani:2026:g3sr}, we report PSNR and LPIPS~\citep{zhang:2018:tueo} for reconstruction.
Following \citet{li:2025:hd3l,xiong:2026:sfs,li:2026:mtph,li:2025:iggt}, we evaluate semantic segmentation using mIoU and mAcc over ADE20K classes present in each reference scene. 
Building on IGGT and video instance segmentation~\citep{li:2025:iggt,yang:2019:vis}, we evaluate cross-view instance consistency using the mean track precision--recall product (mTrackPR) and TrackPR\textsubscript{25} at IoU \num{0.25} within ground-truth thing regions (\cref{sec:supp-evaluation}).

\inlinesection{Baselines.}
We benchmark against existing 3D understanding methods combined with Lyra's generative pipeline. \emph{Per-scene optimization}: LangSplatV2, SFS, and Unified-Lift~\citep{li:2025:hd3l,xiong:2026:sfs,zhu:2025:rete} fit feature fields on frozen Lyra Gaussians. \emph{Feed-forward}: Chorus~\citep{li:2026:mtph} processes frozen Lyra Gaussians, while IGGT~\citep{li:2025:iggt} reconstructs geometry and instances directly from generated RGB videos. \emph{References}: Reconstruction-only Lyra, alongside a teacher-lifting reference that directly assigns teacher features and instance IDs to frozen Lyra ray-aligned-Gaussians.
The latter contextualizes the quality of our segmentation training signals.

\subsection{Evaluating Generative 3D Reconstruction and Understanding}

\begin{table}[t]
    \centering
    \caption{\textbf{Static evaluation on LucidLyra-SDG and LucidLyra-ADE20K.} We report PSNR (dB), LPIPS (--), and mIoU, mAcc, mTrackPR, and TrackPR\textsubscript{25} (all in $\%$). $\dagger$: uses video model RGB output.}

    \scriptsize
\sisetup{
  table-number-alignment=center,
  detect-weight=true,
  mode=text
}
\setlength{\tabcolsep}{1.0pt}
\renewcommand{\arraystretch}{0.95}

\begin{tabularx}{\textwidth}{
    >{\hspace{-\tabcolsep}\raggedright\arraybackslash
      \columncolor{white}[\tabcolsep][\tabcolsep]}
    X
    S[table-format=2.2,table-column-width=3.8em]
    S[table-format=1.3,table-column-width=3.8em]
    S[table-format=2.1,table-column-width=3.8em]
    S[table-format=2.1,table-column-width=3.8em]
    S[table-format=2.1,table-column-width=3.8em]
    S[table-format=2.1,table-column-width=3.8em]
    S[table-format=2.2,table-column-width=3.8em]
    S[table-format=1.3,table-column-width=3.8em]
    S[table-format=2.1,table-column-width=3.8em]
    S[table-format=2.1,table-column-width=3.8em]
    S[table-format=2.1,table-column-width=3.8em]
    S[table-format=2.1,table-column-width=3.8em]
}
    \toprule

    &
    \multicolumn{6}{c}{\textbf{\MethodName-SDG}~\textit{Static}} &
    \multicolumn{6}{c}{\textbf{\MethodName-ADE20K}} \\[-1pt]
    \cmidrule(lr){2-7}
    \cmidrule(lr){8-13}

    &
    \multicolumn{2}{c}{{\emph{Reconstruction}}} &
    \multicolumn{2}{c}{{\emph{Semantics}}} &
    \multicolumn{2}{c}{{\emph{Instance}}} &
    \multicolumn{2}{c}{{\emph{Reconstruction}}} &
    \multicolumn{2}{c}{{\emph{Semantics}}} &
    \multicolumn{2}{c}{{\emph{Instance}}} \\[-1pt]
    \cmidrule(lr){2-3}
    \cmidrule(lr){4-5}
    \cmidrule(lr){6-7}
    \cmidrule(lr){8-9}
    \cmidrule(lr){10-11}
    \cmidrule(lr){12-13}

    \multirow{-3}{*}{\vspace{1em}\strut\textbf{Method}} 
    &
    \multicolumn{1}{c}{\textbf{{PSNR~\textuparrow}}} &
    \multicolumn{1}{c}{\textbf{{LPIPS~\textdownarrow}}} &
    \multicolumn{1}{c}{\textbf{{mIoU~\textuparrow}}} &
    \multicolumn{1}{c}{\textbf{{mAcc~\textuparrow}}} &
    \multicolumn{1}{c}{\bfseries\begin{tabular}[c]{@{}c@{}}mTrack\\PR~\textuparrow\end{tabular}} &
    \multicolumn{1}{c}{\bfseries\begin{tabular}[c]{@{}c@{}}Track\\PR\textsubscript{25}\textuparrow\end{tabular}} &
    \multicolumn{1}{c}{\textbf{{PSNR~\textuparrow}}} &
    \multicolumn{1}{c}{\textbf{{LPIPS~\textdownarrow}}} &
    \multicolumn{1}{c}{\textbf{{mIoU~\textuparrow}}} &
    \multicolumn{1}{c}{\textbf{{mAcc~\textuparrow}}} &
    \multicolumn{1}{c}{\bfseries\begin{tabular}[c]{@{}c@{}}mTrack\\PR~\textuparrow\end{tabular}} &
    \multicolumn{1}{c}{\bfseries\begin{tabular}[c]{@{}c@{}}Track\\PR\textsubscript{25}\textuparrow\end{tabular}} \\
    \midrule

    \mbox{\color{gray}Lyra~+~Lifted Teachers} &
    \color{gray} 24.49 & \color{gray} 0.211 &
    \color{gray} 60.5 & \color{gray} 75.3 &
    \color{gray} 22.8 & \color{gray} 33.2 &
    \color{gray} 24.59 & \color{gray} 0.217 &
    \color{gray} 50.1 & \color{gray} 64.5 &
    \color{gray} 17.3 & \color{gray} 26.8 \\
    \midrule

    Lyra &
    24.49 & 0.211 
    & {\text{--}} & {\text{--}} 
    & {\text{--}} & {\text{--}} 
    & 24.59 & 0.217  
    & {\text{--}} & {\text{--}} 
    & {\text{--}} & {\text{--}} \\

    Lyra~+~LangSplatV2 &
    24.49 & 0.211 
    & 32.4 & 52.7 
    & {\text{--}} & {\text{--}} 
    & 24.59 & 0.217 
    & 37.1	& 53.7 
    & {\text{--}} & {\text{--}} \\

    Lyra~+~SFS &
    24.49 & 0.211 
    & 31.9 & 49.5 
    & {\text{--}} & {\text{--}} 
    & 24.59 & 0.217 
    & 35.2 & 51.3 
    & {\text{--}} & {\text{--}} \\

    Lyra~+~Unified-Lift &
    24.49 & 0.211 
    & {\text{--}} & {\text{--}} 
    & 11.7 & 17.0
    & 24.59 & 0.217 
    & {\text{--}} & {\text{--}} 
    & 7.5 & 11.5 \\

    Lyra~+~Chorus &
    24.49 & 0.211 
    & 11.5	& 19.3
    & {\text{--}} & {\text{--}} 
    & 24.59 & 0.217 
    & 9.3 & 16.1 
    & {\text{--}} & {\text{--}} \\

    Lyra\textsuperscript{$\dagger$}+~IGGT &
    {\text{--}} & {\text{--}}
    & 29.2	& 38.3
    & 5.2 & 9.3 
    & {\text{--}} & {\text{--}} 
    & 32.7	& 41.2 
    & 9.5 & 16.6 \\
    \midrule

    \rowcolor{lightgray}
    {\MethodName \textit{(Ours)}} &
    \bfseries 25.02   & \bfseries 0.186 
    & 64.1  & 79.2 
    &  21.5  & 29.3 
    & \bfseries 24.72 & \bfseries 0.199  
    & 49.8  & 65.6 
    & \bfseries 18.5  & \bfseries 29.7 \\

    \rowcolor{lightgray}
    {\MethodNamePlus\textit{(Ours)}}
    & \bfseries25.02 &\bfseries0.186 
    & \bfseries67.9 & \bfseries81.7 
    & \bfseries22.0 & \bfseries31.1 
    & \bfseries24.72 & \bfseries0.199  
    & \bfseries53.7 & \bfseries71.6 
    & \bfseries18.5 & 28.5 \\[-1pt]
    \bottomrule
\end{tabularx}

    \label{tab:static_main}

\end{table}

\MethodName preserves reconstruction fidelity while adding understanding fields (\cref{tab:static_main}). Relative to Lyra, PSNR improves by \num{0.53}/\num{0.13}\,dB and LPIPS improves by \num{0.025}/\num{0.018} across benchmarks, confirming that our multi-task training does not degrade novel-view rendering quality.

\MethodName outperforms all two-stage baselines without per-scene optimization (\cref{tab:static_main}) in semantics, achieving \num{64.1}/\num{49.8} mIoU with synthetic and real-image conditioning, respectively. Plain \MethodName exceeds the 3D-lifted teacher reference on SDG (\num{64.1} vs.\ \num{60.5}) and performs comparably on ADE20K (\num{49.8} vs.\ \num{50.1}). \MethodNamePlus input-image 3D anchoring further boosts mIoU to \num{67.9}/\num{53.7}.

\begin{figure}[t]
    \centering
    \input{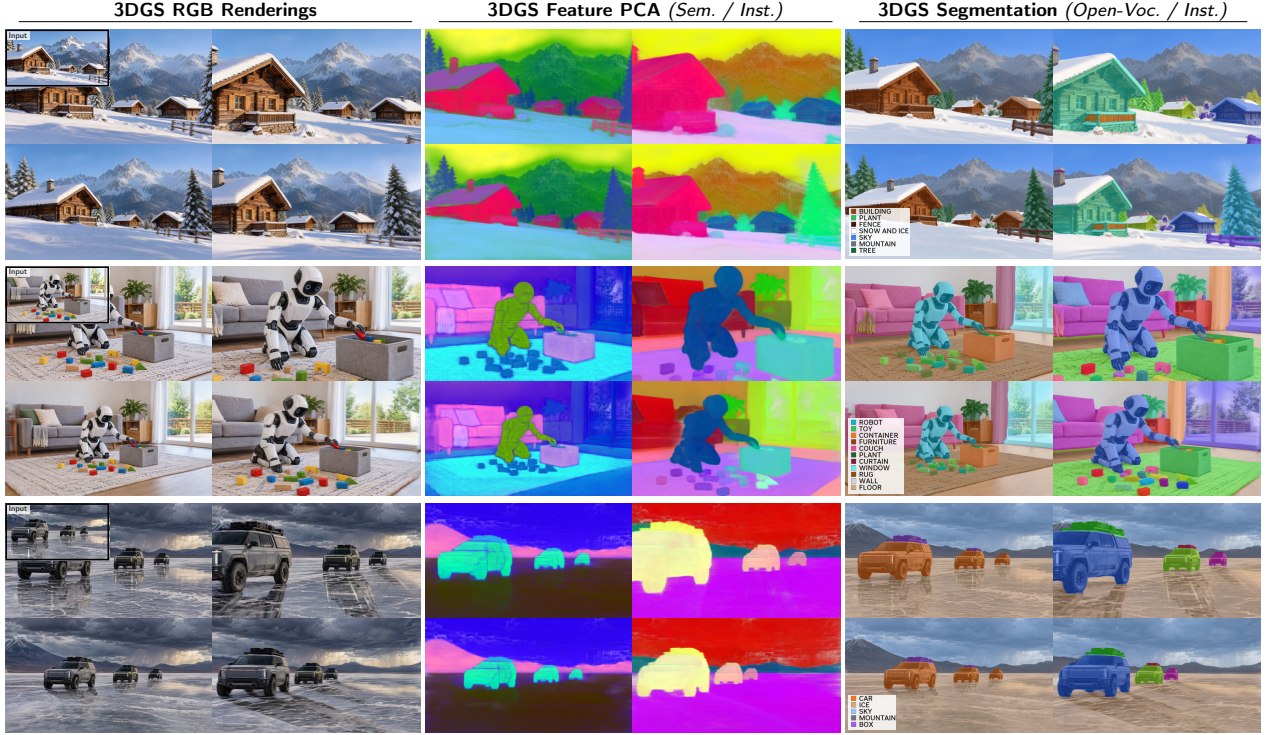}
    \vspace{-0.75em}
    \caption{\textbf{\MethodName qualitative results.} We visualize input- and novel-view 3DGS~\citep{kerbl:2023:3gsf} renderings across RGB, semantic/instance feature PCAs, and open-vocabulary/instance segmentation using generated input images.}
    \label{fig:static-qualitative}
\end{figure}

\MethodName achieves \num{21.5}/\num{18.5} instance mTrackPR (\cref{tab:static_main}), outperforming the compared segmentation baselines in separating individual objects across generated viewpoints. Compared to the lifted teacher reference, mTrackPR is lower on SDG (\num{21.5} vs.\ \num{22.8}) and higher on ADE20K (\num{18.5} vs.\ \num{17.3}). \MethodNamePlus gives a small SDG mTrackPR gain. Qualitative multi-view segmentations and feature PCAs are visualized in \cref{fig:static-qualitative}.

On a single NVIDIA A100 \num{80}\,GB, teacher lifting, LangSplatV2, SFS, and Unified-Lift require \num{64}, \num{697}, \num{168}, and \num{216}\,min to segment a single generated scene, respectively. This is dominated by foundation-model preprocessing.
Feed-forward methods CHORUS and IGGT require \num{1.0} and \num{3.6}\,min, respectively, while \MethodName and \MethodNamePlus require \num{2.1} and \num{2.3}\,min. We measure wall-clock time from Lyra generated scenes (baselines) or video model latents (\MethodName) to the final rgb, open-vocabulary semantic segmentation, and instance segmentation prediction.

\WFclear
\subsection{\MethodName Analysis}
\inlinesection{Losses and regularization.}
Removing C-RADIO~\citep{heinrich:2025:ibav} conditioning causes the largest drop in understanding: \num{7.2} mIoU and \num{6.6} mTrackPR points (\cref{tab:analysis}).
Direct Gaussian semantic supervision adds \num{0.7} mIoU and \num{2.7} mTrackPR points beyond rendered supervision. Reconstruction masking and opacity pruning add \num{1.3} and \num{2.1} mTrackPR points, respectively, with small changes in reconstruction quality.
Part-level supervision slightly improves mIoU and enables a more fine-grained scene interaction.
The full model also retains reconstruction quality, slightly improving PSNR (\num{25.02} vs.\ \num{24.88}) and 
\begin{wraptable}{r}{0.53\textwidth}
    \centering
    \caption{\textbf{Ablations on \MethodName-SDG.} PSNR (dB), LPIPS (–), and mIoU, mAcc, mTrackPR, and TrackPR25 (all in $\%$).}
    \vspace{-1em}
    \begingroup
\scriptsize
\sisetup{
  table-number-alignment=center,
  detect-weight=true,
  mode=text
}
\setlength{\tabcolsep}{0.8pt}
\renewcommand{\arraystretch}{0.95}
\newcommand{\analysismetric}[1]{%
    \multicolumn{1}{c}{\bfseries\begin{tabular}[c]{@{}c@{}}#1\end{tabular}}%
}

\begin{tabularx}{\linewidth}{
    >{\hspace{-\tabcolsep}\raggedright\arraybackslash
      \columncolor{white}[\tabcolsep][\tabcolsep]}X
    >{\centering\arraybackslash}
    S[table-format=2.2]
    S[table-format=1.3]
    S[table-format=2.1]
    S[table-format=2.1]
    S[table-format=2.1]
    S[table-format=2.1]
}
    \toprule
    &
    \multicolumn{2}{c}{{\emph{Reconstruction}}} &
    \multicolumn{2}{c}{{\emph{Semantics}}} &
    \multicolumn{2}{c}{{\emph{Instance}}}\\[-1pt]
    
    \cmidrule(lr){2-3}
    \cmidrule(lr){4-5}
    \cmidrule(lr){6-7}
    
    \multirow{-3}{*}{\strut\textbf{Method}} &
    \analysismetric{PSNR~\textuparrow} &
    \analysismetric{LPIPS~\textdownarrow} &
    \analysismetric{mIoU~\textuparrow} &
    \analysismetric{mAcc~\textuparrow} &
    \analysismetric{mTrack\\PR~\textuparrow} &
    \analysismetric{Track\\PR\textsubscript{25}\textuparrow} \\
    \midrule

    \rowcolor{lightgray}
    {\MethodName \textit{(Ours)}} &
    25.02 & 0.186 & 64.1  & 79.2 & 21.5 & 29.3  \\
    \midrule
    \multicolumn{7}{l}{\textbf{Loss}}\\
    w/o 3D Loss & 24.89 & 0.190 & 63.4 & 77.0 & 18.8 & 26.2 \\
    w/o C-RADIO & 24.77 & 0.190 & 56.9 & 70.2 & 14.9 & 24.5 \\
    w/o part-level supervision & 25.00 & 0.187 & 63.8 & 79.0 & 21.0 & 29.6 \\
    w/o opacity pruning & 25.03 & 0.187 & 63.7	& 78.7	& 19.4 & 27.7 \\
    w/o ignore mask recon. & 24.84 & 0.189 & 64.0 & 78.8 & 20.2 & 28.7 \\
    w/o \MethodName & 24.88 & 0.189 & \text{--} & \text{--} & \text{--} & \text{--}\\
    \midrule
    \multicolumn{7}{l}{\textbf{Semantic Feature Dim.}}\\
    \num{64} & 24.83 & 0.189	& 62.7 & 76.2 & 20.5 & 30.2 \\
    \num{256} & \multicolumn{6}{c}{OOM}\\
    \midrule
    \multicolumn{7}{l}{\textbf{Instance Feature Dim.}}\\
    \num{8} & 24.88 & 0.188 & 64.1 & 78.6 & 19.3 & 28.7 \\
    \num{16} & 24.98 & 0.187 & 63.1 & 77.6 & 18.1 & 27.0 \\
    \midrule
    \multicolumn{7}{l}{\textbf{Shared 3DGS Decoder Layers}} \\
    \num{4} & 25.03 & 0.184 & 64.3 & 79.1 & 21.0 & 29.8 \\
    \num{12} & 24.83 & 0.189 & 63.6 & 78.2 & 18.6 & 27.9 \\

    \bottomrule
\end{tabularx}
\endgroup

    \label{tab:analysis}
    \vspace{-1em}
\end{wraptable}
LPIPS (\num{0.186} vs.\ \num{0.189}) over the reconstruction-only control run. This control initializes from vanilla Lyra checkpoint and is trained identical to \MethodName excluding our changes to the loss terms and architecture.

\inlinesection{Feature dimensions.}
Reducing the semantic feature dimension from \num{128} to \num{64} lowers mIoU by \num{1.4} points and mTrackPR by \num{1.0} point. A \num{256}-dimensional representation exceeds the GPU memory budget. The chosen \num{128}D achieves the best accuracy--memory trade-off among the tested settings.
The \num{12}-dimensional object-level instance representation achieves the highest mTrackPR among the tested settings: \num{21.5}, compared with \num{19.3} at \num{8} dimensions and \num{18.1} at \num{16} dimensions. Semantic mIoU is unchanged at \num{8} dimensions and \num{1.0} point lower at \num{16} dimensions.

\clearpage
\inlinesection{Shared 3DGS decoder layers.}
Increasing the shared decoder depth from eight to twelve layers degrades reconstruction and understanding. Four shared layers remain competitive, while eight shared layers achieve the highest mTrackPR among the tested depths.

\subsection{Downstream Applications}
\MethodName's structured, queryable scenes directly enable downstream tasks including robotic simulation, 3D scene editing, and 3D asset mining (details and qualitative results in \cref{sec:supp-downstream-applications}).

\section{Conclusion}
We presented \MethodName, which turns generated scenes into queryable 3D worlds by distilling generative and perceptual knowledge into a feed-forward Gaussian decoder.
A camera-controlled video model supplies both latent representations and training data, enabling joint reconstruction and understanding without captured multi-view data, manual 3D annotations, or per-scene optimization.
Experiments demonstrate improved semantic and instance understanding over the compared two-stage baselines while preserving Lyra's reconstruction quality.
Furthermore, we demonstrate initial results on dynamic scenes and applications in 3D scene editing, 3D asset mining, and physical-AI simulation.
Future work could scale our approach by accelerated video generation~\citep{li2026solvideoinferenceengine} with larger-scale joint training for generation and scene understanding~\citep{gabeur:2026:igag}.

\clearpage
\section*{Acknowledgements}
We sincerely thank Tianchang Shen, Xuanchi Ren, and Zan Gojcic for their support. We also thank Riccardo de Lutio, Nicholas Sharp, and Benjamin Missaoui for insightful discussions, and Simon Kiefhaber and Jan-Martin Steitz for their assistance with the LAION dataset.
Stefan Roth was partially supported by the ERC under the European
Union’s Horizon 2020 research and innovation programme
(grant agreement No. 866008) and the Deutsche Forschungsgemeinschaft (German Research Foundation, DFG) under Germany’s Excellence Strategy (EXC
3066/1 “The Adaptive Mind”, Project No. 533717223). 
Finally, we acknowledge the support of the
European Laboratory for Learning and Intelligent Systems.

{\small
\bibliographystyle{plainnat}
\bibliography{bibliography/short,bibliography/papers}
}
\clearpage
\appendix
\crefalias{section}{appendix}
\crefalias{subsection}{subappendix}

\section*{Appendix}
\label{sec:supp}
The appendix covers implementation details (\cref{sec:supp-implementation}), additional experiments (\cref{sec:supp-additional-experiments,sec:supp-qualitative}), application experiments (\cref{sec:supp-downstream-applications}), and datasets (\cref{sec:supp-datasets}).

\section{Implementation Details}
\label{sec:supp-implementation}
We report the hyperparameters and implementation choices that determine
teacher-target construction, projector pretraining, \MethodName training, and
inference.

\subsection{Teacher Targets and Feature Compression}
\label{ssec:supp-target-construction}

\inlinesection{Instance supervision.}
We use SAM1 ViT-H~\citep{kirillov:2023:sam} to initialize SAM2.1 Hiera-Base+~\citep{ravi:2025:saiv} on the shared conditioning
frame \num{0}, followed by independent propagation through each \num{121}-frame
trajectory. We use the SAM automatic mask generator with \num{32} points per side,
predicted-IoU and stability thresholds of \num{0.84} and \num{0.86}, one crop layer,
a crop-grid downscaling factor of two, box and crop NMS thresholds of \num{0.7},
and a minimum mask area of \num{100} pixels. We do not threshold SAM2 logits and
merge redundant or nested tracks at a mask-coverage threshold of \num{0.5}. Part-level targets retain SAM multimask
candidates \num{0} and \num{1} and use the same propagation and consolidation procedure.

\inlinesection{Semantic supervision.}
We build upon RADSeg~\citep{alama:2026:upce} and C-RADIOv3-B~\citep{heinrich:2025:ibav} with its \num{1152}-dimensional SigLIP2~\citep{tschannen:2025:mvle} adapter, a crop
size of \num{336}, stride \num{224}, and SCRA and SCGA scales of \num{10}. RADSeg class
prediction and SAM refinement are not used. We project the dense tokens to
\num{128} dimensions, normalize them, and resize them bilinearly. We pool all pixels and frames
belonging to a track, which yields one pixel-weighted semantic centroid per track and trajectory. 
This is performed in the object-level instance tracks as well as the part-level tracks.

Constructing semantic and instance targets for the full corpus
required an aggregate of \num{1580} NVIDIA A100 \num{80}\,GB GPU-days offline. This is a one-time training-data preprocessing cost. We will make the preprocessed data available.

\inlinesection{Vision-language feature dimensionality reduction.}
$P_{\mathrm{VL}}$ sums a $\num{1152}{\to}\num{128}$ linear projection and a
$\num{1152}{\to}\num{4608}{\to}\num{128}$ two-layer ReLU MLP. Image and text outputs share the same projection and are $\ell_2$-normalized afterward. We train the projector with frozen SigLIP2
SO400M/16 NaFlex features and the SigLIP alignment objective~\citep{zhai:2024:slli} on filtered
LAION-2B~\citep{schuhmann:2022:laion}. During training, we use a global batch size of \num{16384} across eight GPUs, AdamW~\citep{loshchilov:2019:dwdr} optimizer with learning rate
\num{1e-3}, $\beta=(\num{0.9},\num{0.999})$, $\epsilon=\num{1e-8}$, matrix-only weight decay
\num{0.2}, \num{10000} warm-up steps followed by cosine decay, and gradient clipping at
\num{1}. The SigLIP2 logit bias remains frozen, its scale is capped at \num{100}, and
images use the native processor without additional augmentation. We train for \num{32} epochs and keep the projection network frozen afterward. Training takes \num{2} days using \num{8} NVIDIA A100 \num{80}\,GB GPUs.

\subsection{\MethodName Training}
\label{ssec:supp-curriculum}
\label{ssec:supp-loss-implementation}

We use eight shared layers in the decoder, followed by eight separate layers for reconstruction and understanding each. We initialize the C-RADIO injection gate to \num{0.05}, and its
learnable latent-time weights as
$w_\ell=(1-\ell/15)^2$. We predict raw Gaussian parameters using the reconstruction head.
Each static Gaussian additionally carries a \num{128}-dimensional semantic feature,
a \num{12}-dimensional object-level instance feature, and a \num{4}-dimensional part-level instance feature. We clamp semantic and instance outputs to $[\num{-10},\num{10}]$, with object and part channels normalized separately.
We initialize the shared trunk and reconstruction head \(D_{\mathrm{rec}}\) from pretrained Lyra and initialize \(D_{\mathrm{seg}}\) from scratch.
We optimize all decoder parameters at $\num{704}\times\num{1280}$ resolution using
AdamW with $\beta=(\num{0.9},\num{0.999})$, $\epsilon=\num{1e-8}$, weight decay \num{1e-2}, and gradient clipping at \num{1}. We train on eight GPUs, processing one scene each, without gradient accumulation. We set the learning rates to \num{1e-5} for the shared trunk and \num{1e-4} for the understanding branch, heads, and conditioning modules. We keep the learning rate constant after \num{100} warm-up steps. \Cref{tab:supp-training-schedule} summarizes the training curriculum.

\begin{table}[t]
    \centering
    \caption{\textbf{\MethodName training setup.}}
    \label{tab:supp-training-schedule}
    \scriptsize
    \setlength{\tabcolsep}{5pt}
    \begin{tabular}{@{}ccccccc@{}}
        \toprule
        \textbf{Stage} & $\mathbf{H\times W}$ & $\mathbf{L}$ & $\mathbf{V}$ & $\mathbf{S}$ & $\mathbf{B}$ & \textbf{Steps} \\
        \midrule
        \multicolumn{7}{@{}l}{Static} \\
        \num{1} & $\num{704}\times\num{1280}$ & \num{121} & \num{1} & \num{9} & \num{1} & \num{10000} \\
        \num{2} & $\num{704}\times\num{1280}$ & \num{121} & \num{1} & \num{9} & \num{1} & \num{20000} \\
        \num{3} & $\num{704}\times\num{1280}$ & \num{121} & \num{1} & \num{9} & \num{1} & \num{20000} \\
        \num{4} & $\num{704}\times\num{1280}$ & \num{121} & \num{1}--\num{6} & $\num{3}V$ & \num{1} & \num{10000}  \\
        \midrule
        \multicolumn{7}{@{}l}{Dynamic} \\
        \num{5} & $\num{704}\times\num{1280}$ & \num{121} & \num{1}--\num{6} & \num{12} & \num{1}
          & \num{10000} \\
        \bottomrule
    \end{tabular}
\end{table}

Here, $L$ is the number of source frames per trajectory, $V$ is the number
of input trajectories, $S$ is the number of supervision views per scene, and
$B$ is the per-GPU batch size. The corresponding global batch size is eight.
Stages \num{1}--\num{4} respectively denote auxiliary initialization, single-trajectory
training with object targets, single-trajectory training with object- and part-level targets,
and multi-trajectory training with object- and part-level targets. Opacity pruning is applied during the last \num{1000} steps of Stage \num{4}.
Stage \num{5} initializes from stage \num{4} checkpoint before opacity pruning.
For Stage \num{5}, $S=\num{12}$ for reconstruction, depth, and semantic supervision,
whereas the instance loss uses a random six-trajectory (forward or backward) block at a time
($S=\num{6}$). The opacity pruning loss is applied for the last \num{1000} steps.

We set the weights of the RGB, LPIPS~\citep{zhang:2018:tueo}, and depth losses to \num{1}, \num{0.5}, and \num{0.05}. For the rendered and direct per-Gaussian semantic losses, we use weight \num{0.5} during stage \num{1} and \num{4} thereafter. We set the weights of the object and part contrastive losses to \num{0.05}. For the static model, from multi-trajectory training onward, we restrict the reconstruction to valid object-track pixels,
and gradients from rendered feature losses also update Gaussian geometry.
We use the reconstruction mask to exclude untracked background via ignored SAM pixels.
During the pruning stage, we retain the highest-opacity \num{20}\% of Gaussians for rendering and direct semantic supervision, and add an opacity regularizer equal to \num{0.1} times their mean activated opacity.
For object and part losses independently, we sample at most \num{4096} tracked pixels per scene across the target views, using replacement when subsampling, and use a temperature of \num{0.05}. Anchors must have at least one positive, and part anchors must additionally have a negative from another part of the same object. We use $\epsilon=\num{1e-8}$ for feature normalization, and \num{1e-6} for logarithms.

For dynamic training, we zero-initialize the time-conditioning weights and sample one to six original source trajectories and a query time uniformly.
All \num{12} original and time-reversed target trajectories provide reconstruction,
depth, and semantic supervision. The instance loss uses one randomly selected
six-trajectory block with consistent instance IDs (\cref{ssec:method-dynamic}).
Since the preprocessing of the reversed trajectories is seeded with the last frame of the original video, the instance IDs are different from those of the forward trajectories.
C-RADIO conditioning uses the queried-time frame, and reconstruction uses valid
object-track support. Temporal prototype matching uses a
cosine-similarity threshold of \num{0.2}. Training the static and dynamic Gaussian decoders takes \num{10} days using \num{8} NVIDIA A100 \num{80}\,GB GPUs.

\subsection{Inference and Evaluation Protocol}
\label{sec:supp-evaluation}
\label{ssec:supp-inference}

\inlinesection{Inference.}
We use batch size one and all six input trajectories.
At inference, we prune the lowest-opacity \num{80}\% of Gaussians.
\MethodNamePlus uses the same checkpoint and retained Gaussians, changing only semantic readout and the resulting instance grouping.
We $\ell_2$-normalize Gaussian features and \num{128}-D text embeddings obtained by
projecting the SigLIP2 features from \num{1152} dimensions.
The learned instance embeddings are class-agnostic, but inference clusters them within predicted semantic classes.
For each class, we select the $\lceil\num{0.5}N\rceil$ highest-opacity Gaussians,
and average each normalized object embedding with $k=\num{20}$.
For all datasets, object clustering uses Euclidean HDBSCAN~\citep{campello:2013:dbcb} with minimum cluster size \num{1000}, minimum samples \num{5}, $\epsilon=\num{0.06}$, leaf selection, and a single cluster allowed.
Unassigned Gaussians are assigned to the nearest neighbor.
For part-level instance grouping, we cluster the final four instance channels using an opacity threshold of \num{0.1}, Euclidean HDBSCAN, minimum cluster size \num{100}, minimum samples \num{5}, $\epsilon=\num{0.01}$, EOM selection, and allow for a single cluster.
Static \MethodNamePlus uses the RADSeg+ prediction of the input image; dynamic \MethodNamePlus applies RADSeg+ anchoring to the input frame at each queried time.
Labels are nearest-neighbor resized to the source-ray grid and gathered using
the retained ray indices.
For the prototype assignments, we use cosine similarity $\ge\num{0.95}$.
For dynamic inference, we decode and cluster each queried state independently using the same object-clustering settings, a top-opacity ratio of \num{0.5}, assignment $k=\num{20}$, and a temporal similarity threshold of \num{0.2}.

\inlinesection{Metrics and Comparison Protocol}

We perform the static evaluation of all trajectories at frames
$\{\num{0},\num{20},\num{40},\num{60},\num{80},\num{100},\num{120}\}$, and the dynamic evaluation of all six trajectories at queried times $\{\num{0},\num{40},\num{80},\num{120}\}$.
We compute PSNR on RGB values clamped to $[\num{0},\num{1}]$, SSIM~\citep{wang:2004:iqaf} with an
$\num{11}{\times}\num{11}$ Gaussian window, and VGG-LPIPS~\citep{simonyan:2015:vdcn} with normalized inputs.
Semantic evaluation reports mIoU and mAcc using the ADE20K~\citep{zhou:2019:sust} classes present
in each reference scene. Instance evaluation clips predictions and references
to annotated thing regions. These reference-conditioned scores measure
classification within a known vocabulary and grouping within known foreground.

We evaluate instance consistency across
views. We follow video instance segmentation~\citep{yang:2019:vis}
in representing identities as tubes and computing tube IoU as the ratio
of summed intersections to summed unions across observations within each scene.
At each threshold $\tau\in\mathcal T=\{\num{0.25},\num{0.30},\ldots,\num{0.75}\}$,
Hungarian assignment maximizes the number of one-to-one matches with
tube IoU $\ge\tau$. Matched pairs, unmatched predictions, and unmatched
references define TP, FP, and FN, respectively. We pool these counts
across scenes and define the track precision--recall product:
\begin{equation}
\mathrm{TrackPR}_{\tau}=
\frac{\mathrm{TP}_{\tau}}{\mathrm{TP}_{\tau}+\mathrm{FP}_{\tau}}
\frac{\mathrm{TP}_{\tau}}{\mathrm{TP}_{\tau}+\mathrm{FN}_{\tau}},
\qquad
\mathrm{mTrackPR}=\frac{1}{11}\sum_{\tau\in\mathcal T}\mathrm{TrackPR}_{\tau}.
\end{equation}
We report mTrackPR and TrackPR\textsubscript{25} as percentages, defining zero-denominator
ratios as zero. 
Our clustering produces masks without confidence scores. TrackPR therefore uses an unranked precision--recall product over tube-IoU matches.
In contrast, IGGT~\citep{li:2025:iggt} averages tracking mIoU and tracking success rate over reference objects, without directly penalizing unmatched predicted instances. TrackPR counts unmatched predicted tubes as false positives, penalizing over-segmentation within the evaluated thing regions.

\inlinesection{Baseline implementations.}
LangSplatV2~\citep{li:2025:hd3l}, SFS~\citep{xiong:2026:sfs}, Unified-Lift~\citep{zhu:2025:rete}, and Chorus~\citep{li:2026:mtph} use a frozen  Lyra~\citep{bahmani:2026:g3sr} reconstruction. Our lifted-teacher reference changes only the feature fields, assigning our compressed SigLIP2 semantic features and deterministic \num{12}-D Hamming codes to RGB-derived SAM tracks via the ray-to-gaussian alignment. LangSplatV2~\citep{li:2025:hd3l} independently optimizes three language levels for \num{10000} steps using a \num{64}-entry VQ codebook. SFS~\citep{xiong:2026:sfs} lifts SAM ViT-H/OpenCLIP ViT-B/16~\citep{cherti:2023:rslc} features with Tikhonov coefficient \num{1.2} and temperature \num{100}. Unified-Lift~\citep{zhu:2025:rete} trains a \num{16}-D object field and \num{256}-entry codebook for \num{30000} steps with Adam~\citep{kingma:2015:amfs} learning rates \num{2.5e-3} and \num{5e-4}; and Chorus~\citep{li:2026:mtph} uses its released language--DINO model. IGGT~\citep{li:2025:iggt} reconstructs geometry from \num{11} calibrated \(\num{518}{\times}\num{280}\) views and applies \(k{=}\num{20}\) feature smoothing followed by HDBSCAN \((\epsilon{=}\num{0.01},\ \mathrm{minPts}{=}\num{100},\ \mathrm{MCS}{=}\num{500})\). Using more input views results in OOM errors and significantly increased inference time due to the HDBSCAN clustering while not improving the instance segmentation metrics. For dynamic baselines, 4D LangSplat~\citep{li:2025:4l4l} fits two \num{100}-epoch autoencoders and four \num{24000}-step fields to all \num{726} observations; TRASE~\citep{li:2026:tt4s} trains a \num{32}-D field for \num{30000} steps. IGGT4D~\citep{zou:2026:is4i} streams all views through its fixed ViT-G/\num{8}-D model in rewrite mode, using \num{0.90}/\num{0.85} match/grow thresholds and LSeg~\citep{li:2022:lss} semantics.

\section{Additional Results}
\label{sec:supp-additional-experiments}

\subsection{Dynamic Reconstruction and Understanding}
\label{ssec:supp-dynamic-results}

Next, we demonstrate that our method can be extended to dynamic 3D scene generation from a monocular input video.
\Cref{tab:dynamic_main} compares \MethodName queried-time scene states with dynamic
reconstruction and understanding baselines~\citep{bahmani:2026:g3sr,li:2025:4l4l,li:2026:tt4s,zou:2026:is4i} in the \MethodName-SDG Dynamic evaluation dataset.

\begin{table}[t]
    \centering
    \caption{\textbf{Dynamic reconstruction and understanding on LucidLyra-SDG Dynamic.} We report PSNR (dB), LPIPS (--), and mIoU, mAcc, mTrackPR, and TrackPR\textsubscript{25} (all in $\%$).
    $\dagger$ denotes using the video diffusion model's RGB output.}
    \begingroup
\scriptsize
\sisetup{
  table-number-alignment=center,
  detect-weight=true,
  mode=text
}
\setlength{\tabcolsep}{1.0pt}
\renewcommand{\arraystretch}{0.95}

\begin{tabularx}{\linewidth}{
    >{\hspace{-\tabcolsep}\raggedright\arraybackslash
      \columncolor{white}[\tabcolsep][\tabcolsep]}X
    S[table-format=2.2,table-column-width=3.8em]
    S[table-format=1.3,table-column-width=3.8em]
    S[table-format=2.1,table-column-width=3.8em]
    S[table-format=2.1,table-column-width=3.8em]
    S[table-format=2.1,table-column-width=3.8em]
    S[table-format=2.1,table-column-width=3.8em]
}
    \toprule
    &
    \multicolumn{2}{c}{\emph{Reconstruction}} &
    \multicolumn{2}{c}{\emph{Semantics}} &
    \multicolumn{2}{c}{\emph{Instance}}\\[-1pt]

    \cmidrule(lr){2-3}
    \cmidrule(lr){4-5}
    \cmidrule(lr){6-7}
    
    \multirow{-3}{*}{\strut\textbf{Method}} &
    \multicolumn{1}{c}{\textbf{PSNR~\textuparrow}} &
    \multicolumn{1}{c}{\textbf{LPIPS~\textdownarrow}} &
    \multicolumn{1}{c}{\textbf{mIoU~\textuparrow}} &
    \multicolumn{1}{c}{\textbf{mAcc~\textuparrow}} &
    \multicolumn{1}{c}{\bfseries\begin{tabular}[c]{@{}c@{}}mTrack\\PR~\textuparrow\end{tabular}} &
    \multicolumn{1}{c}{\bfseries\begin{tabular}[c]{@{}c@{}}Track\\PR\textsubscript{25}\textuparrow\end{tabular}} \\
    \midrule

    \textcolor{gray}{Lyra + Lifted Teachers} &
    {\color{gray}\num{22.16}} &
    {\bfseries\color{gray}\num{0.248}} &
    {\color{gray}\num{53.2}} &
    {\color{gray}\num{69.4}} &
    {\color{gray}\num{12.1}} &
    {\color{gray}\num{15.1}} \\
    \midrule

    Lyra &
    22.16 & \bfseries 0.248 & {\text{--}} & {\text{--}} & {\text{--}} & {\text{--}} \\

    Lyra\textsuperscript{$\dagger$}+~4D~LangSplat &
    \bfseries 23.03 & 0.354 & 30.3 & 47.8 & {\text{--}} & {\text{--}} \\

    Lyra\textsuperscript{$\dagger$}+~TRASE &
    22.77 & 0.309 & {\text{--}} & {\text{--}} & 7.7 & 12.5 \\

    Lyra\textsuperscript{$\dagger$}+~IGGT~4D &
    {\text{--}} & {\text{--}}
    & 36.1 & 46.0 & 7.8 & 12.9 \\

    \midrule

    \rowcolor{lightgray}
    {\MethodName \textit{(Ours)}}
    & 21.88 & 0.252 &
    53.2 & 69.3 & 11.7 & 19.7 \\[-1pt]

    \rowcolor{lightgray}
    {\MethodNamePlus\textit{(Ours)}} &
    21.88 & 0.252
    & \bfseries 62.6 & \bfseries 79.3 & \bfseries 12.3 & \bfseries 20.4\\[-1pt]
    \bottomrule
\end{tabularx}
\endgroup

    \label{tab:dynamic_main}
\end{table}

\inlinesection{Semantic and instance understanding.}
\MethodName achieves \num{53.2} mIoU and \num{69.3} mAcc; \MethodNamePlus increases
these to \num{62.6} and \num{79.3}, respectively. Their mTrackPR scores are
\num{11.7} and \num{12.3}, with TrackPR\textsubscript{25} scores of \num{19.7} and \num{20.4}.

\inlinesection{Reconstruction.}
\MethodName obtains \num{21.88}\,dB PSNR and \num{0.252} LPIPS~\citep{zhang:2018:tueo},
compared with Lyra's \num{22.16}\,dB and \num{0.248}. This represents a minimal reconstruction trade-off for dynamic scene understanding.

Our dynamic extension represents an initial step toward structured 4D scene generation and understanding, with encouraging results that motivate further development in this direction.

\subsection{Compact Vision-Language Representation}
\label{ssec:supp-projection-quality}
Since the semantic feature dimension directly affects the cost of representing each
Gaussian, we train a projection network inferring a lower-dimensional compressed version of the SigLIP2~\citep{tschannen:2025:mvle} latent space. Next, we evaluate the projector independently of \MethodName on two tasks: zero-shot ImageNet-1K~\citep{russakovsky:2015:ilsv} classification using SigLIP2
SO400M/16 NaFlex, and RADSeg~\citep{alama:2026:upce} segmentation on the \num{2000}-image ADE20K~\citep{zhou:2019:sust} validation
set. In both tasks, we project image and text descriptors into the same
compact space. The segmentation analysis uses three inference resolutions following \citet{alama:2026:upce}.

\begin{table}[t]

    \centering
    \caption{\textbf{$P_{\mathrm{VL}}$ projection quality.}
    SigLIP2 ImageNet val zero-shot classification accuracy and RADSeg open-vocabulary semantic segmentation on ADE20K eval mIoU. Both metrics in \%~\textuparrow.}

    \begingroup
    \scriptsize
    \sisetup{
      table-number-alignment=center,
      detect-weight=true,
      mode=text
    }
    \setlength{\tabcolsep}{2.0pt}
    \renewcommand{\arraystretch}{0.95}

    \begin{tabularx}{\linewidth}{
        >{\hspace{-\tabcolsep}\raggedright\arraybackslash
          \columncolor{white}[\tabcolsep][\tabcolsep]}Xc
        S[table-format=2.2,table-column-width=3em]
        S[table-format=2.2,table-column-width=3em]
        S[table-format=2.2,table-column-width=3em]
        S[table-format=2.2,table-column-width=3em]
        S[table-format=2.2,table-column-width=3em]
    }
        \toprule
        & &
        \multicolumn{2}{c}{{\emph{ImageNet}}} &
        \multicolumn{3}{c}{{\emph{RADSeg on ADE20K}}} \\[-1pt]

        \cmidrule(lr){3-4}
        \cmidrule(lr){5-7}

        \multirow{-2}{*}{\vspace{0.75em}\textbf{{Method}}}&
        \multirow{-2}{*}{\vspace{0.75em}\textbf{{Feature Dim.}}} &
        \multicolumn{1}{c}{\textbf{{Top-1~\textuparrow}}} &
        \multicolumn{1}{c}{\textbf{{Top-5~\textuparrow}}} &
        \multicolumn{1}{c}{\textbf{{Low~\textuparrow}}} &
        \multicolumn{1}{c}{\textbf{{Mid~\textuparrow}}} &
        \multicolumn{1}{c}{\textbf{{High~\textuparrow}}} \\
        \midrule

        SigLIP2       & \num{1152} & 83.50 & 96.86 & 27.16 & 28.96 & 28.94 \\
        + Projection $P_{\mathrm{VL}}$  & \num{256}  & 80.69 & 95.89 & 25.55 & 27.06 & 27.09 \\
        \rowcolor{lightgray} + Projection $P_{\mathrm{VL}}$ & \num{128}  & 76.61 & 94.02 & 24.22 & 25.34 & 25.14 \\
        + Projection $P_{\mathrm{VL}}$ & \num{64}   & 70.16 & 90.43 & 22.13 & 23.66 & 23.61 \\
        \bottomrule
    \end{tabularx}
    \endgroup
    \label{tab:siglip-projection}

\end{table}

\Cref{tab:siglip-projection} shows the accuracy--capacity trade-off.
The \num{256}-dimensional projector reduces descriptor width by $\num{4.5}\times$,
with a \num{2.81}-point decrease in ImageNet top-1 accuracy and a \num{1.79}-point
decrease in ADE20K mIoU averaged over the three resolutions. Our selected
\num{128}-dimensional representation reduces width by $\num{9}\times$ and achieves
\num{76.61}\% top-1 accuracy and \num{24.90} mean mIoU, decreasing \num{6.89} and
\num{3.45} points, respectively. Although \num{256} dimensions better preserve the
native representation, that configuration exceeds the GPU budget when training on \num{80}\,GB GPUs. Further reducing the feature dimension to \num{64} substantially degrades classification and segmentation accuracy. Hence, we use \num{128} dimensional semantic representations.

\section{Qualitative Results}
\label{sec:supp-qualitative}
We show static and dynamic examples, compare baseline predictions, and examine object- and part-level instance fields. We further provide qualitative video results on out project website.

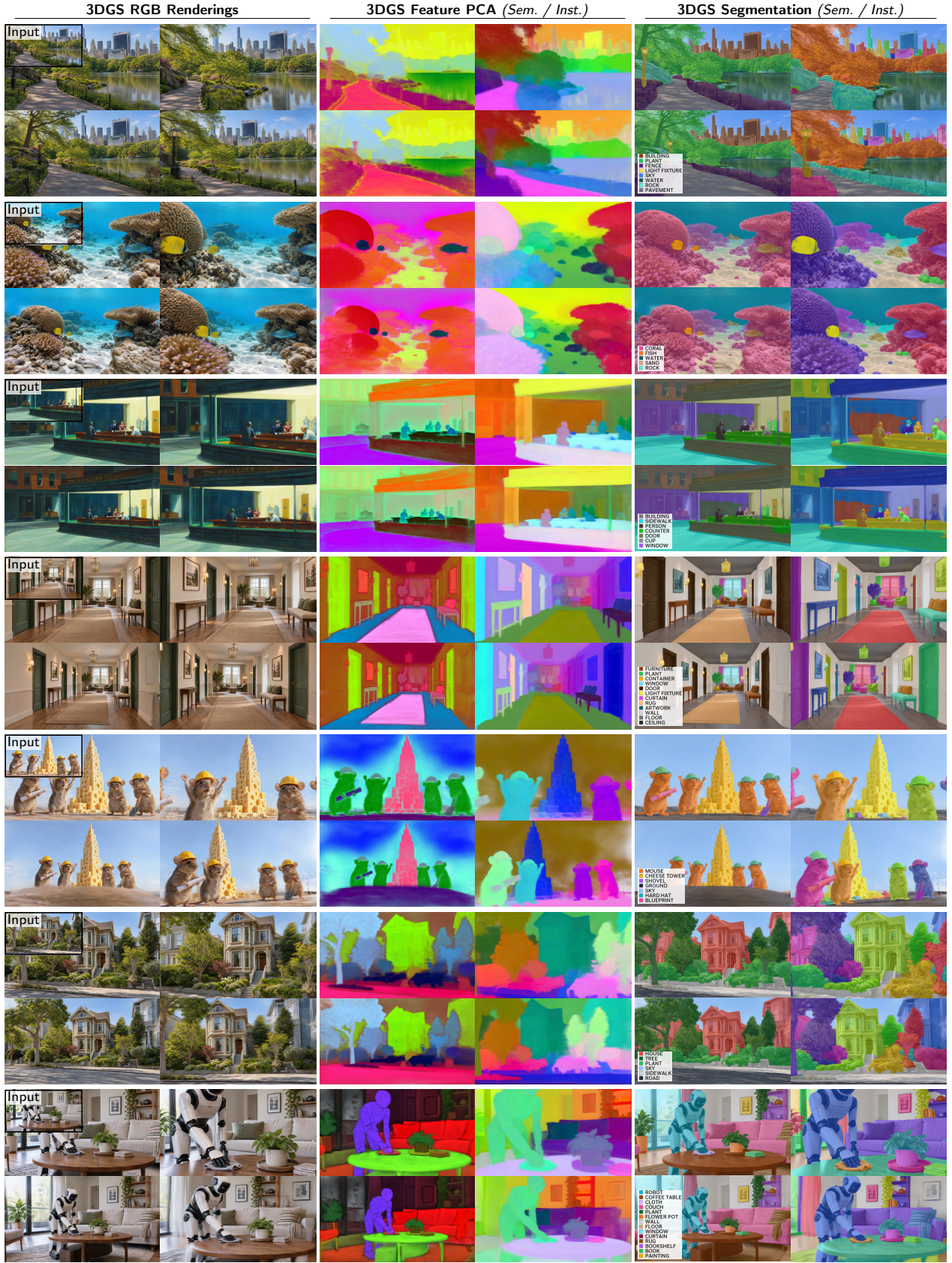
\begin{figure}[p]
    \centering
    \scriptsize
\sffamily

\newcommand{\datapath}{%
    figures/supplement/static_examples/data/%
}

\setlength{\figgroupgap}{2pt}
\setlength{\figpanelwidth}{%
    \dimexpr\linewidth/6-\figgroupgap/3\relax
}
\setlength{\figpanelheight}{0.55\figpanelwidth}

\newcommand{\fullimage}[1]{%
    \includegraphics[width=\figpanelwidth]{\datapath#1}%
}

\newcommand{\imagewithinput}[2]{%
    \begin{tikzpicture}[baseline=0pt]
        \node[anchor=south west,inner sep=0,outer sep=0] (rendering) at (0,0) {%
            \fullimage{#1}%
        };
        \begin{scope}[overlay]
            \node[anchor=north west,inner sep=0,outer sep=0] (inputview)
                at (rendering.north west) {%
                    \includegraphics[width=0.5\figpanelwidth]{\datapath#2}%
                };
            \node[anchor=north west,inner xsep=0.7pt,inner ysep=0.35pt,
                outer sep=0,fill=white,fill opacity=0.75,text opacity=1,
                text=black,font=\normalfont\sffamily\mdseries\upshape\scriptsize]
                at ([xshift=0.6pt,yshift=-0.6pt]inputview.north west) {Input};
            \draw[black,line width=0.6pt]
                ([xshift=0.3pt,yshift=-0.3pt]inputview.north west)
                rectangle ([xshift=-0.3pt,yshift=0.3pt]inputview.south east);
        \end{scope}
    \end{tikzpicture}%
}

\newcommand{\splitimage}[2]{%
    \begin{tikzpicture}[
        x=\figpanelwidth,
        y=\figpanelheight,
        baseline=0pt
    ]
        \path[use as bounding box] (0,0) rectangle (1,1);

        \begin{scope}
            \clip
                (0,0) --
                (0.45,0) --
                (0.55,1) --
                (0,1) -- cycle;

            \node[anchor=south west,inner sep=0] at (0,0) {%
                \includegraphics[width=\figpanelwidth]{\datapath#1}%
            };
        \end{scope}

        \begin{scope}
            \clip
                (0.45,0) --
                (1,0) --
                (1,1) --
                (0.55,1) -- cycle;

            \node[anchor=south west,inner sep=0] at (0,0) {%
                \includegraphics[width=\figpanelwidth]{\datapath#2}%
            };
        \end{scope}

        \draw[white,line width=0.5pt]
            (0.45,0) -- (0.55,1);
    \end{tikzpicture}%
}

\setlength{\tabcolsep}{0pt}

\begin{tabular}{
    @{}c@{}c
    @{\hspace{\figgroupgap}}
    c@{}c
    @{\hspace{\figgroupgap}}
    c@{}c@{}
}
    \multicolumn{2}{c}{\textbf{3DGS RGB Renderings}} &
    \multicolumn{2}{c}{\textbf{3DGS Feature PCA}~\emph{(Sem.~/~Inst.)}} &
    \multicolumn{2}{c}{\textbf{3DGS Segmentation}~\emph{(Sem.~/~Inst.)}}
    \\[-2.5pt]
    \cmidrule(lr){1-2}
    \cmidrule(lr){3-4}
    \cmidrule(lr){5-6}

    \imagewithinput{central_park_13/traj05_t120_rgb_recon.jpg}{central_park_13/input.jpg} &
    \fullimage{central_park_13/traj04_t120_rgb_recon.jpg} &
    \fullimage{central_park_13/traj05_t120_semantic_pca.jpg} &
    \fullimage{central_park_13/traj04_t120_instance_pca.jpg} &
    \fullimage{central_park_13/traj05_t120_open_vocab_seg} &
    \fullimage{central_park_13/traj04_t120_instance_seg}
    \\[-2.5pt]

    \fullimage{central_park_13/traj03_t120_rgb_recon.jpg} &
    \fullimage{central_park_13/traj00_t120_rgb_recon.jpg} &
    \fullimage{central_park_13/traj03_t120_semantic_pca.jpg} &
    \fullimage{central_park_13/traj00_t120_instance_pca.jpg} &
    \begin{tikzpicture}[x=\figpanelwidth,y=\figpanelheight,baseline=0pt]
        \node[anchor=south west,inner sep=0] at (0,0) {%
            \fullimage{central_park_13/traj03_t120_open_vocab_seg}%
        };
        \begin{scope}[overlay,scale=2,transform shape]
            \fill[white,fill opacity=0.85] (0.00937,0.01705) rectangle (0.15234,0.24858);
            \node[anchor=south west,inner sep=0] at (0,0) {%
                \fullimage{central_park_13/semantic_legend.png}%
            };
        \end{scope}
    \end{tikzpicture} &
    \fullimage{central_park_13/traj00_t120_instance_seg} \\

    \imagewithinput{coral_reef_07/traj05_t120_rgb_recon.jpg}{coral_reef_07/input.jpg} &
    \fullimage{coral_reef_07/traj04_t120_rgb_recon.jpg} &
    \fullimage{coral_reef_07/traj05_t120_semantic_pca.jpg} &
    \fullimage{coral_reef_07/traj04_t120_instance_pca.jpg} &
    \fullimage{coral_reef_07/traj05_t120_open_vocab_seg} &
    \fullimage{coral_reef_07/traj04_t120_instance_seg}
    \\[-2.5pt]

    \fullimage{coral_reef_07/traj03_t120_rgb_recon.jpg} &
    \fullimage{coral_reef_07/traj00_t120_rgb_recon.jpg} &
    \fullimage{coral_reef_07/traj03_t120_semantic_pca.jpg} &
    \fullimage{coral_reef_07/traj00_t120_instance_pca.jpg} &
    \begin{tikzpicture}[x=\figpanelwidth,y=\figpanelheight,baseline=0pt]
        \node[anchor=south west,inner sep=0] at (0,0) {%
            \fullimage{coral_reef_07/traj03_t120_open_vocab_seg}%
        };
        \begin{scope}[overlay,scale=2,transform shape]
            \fill[white,fill opacity=0.85] (0.00937,0.01705) rectangle (0.09141,0.16335);
            \node[anchor=south west,inner sep=0] at (0,0) {%
                \fullimage{coral_reef_07/semantic_legend.png}%
            };
        \end{scope}
    \end{tikzpicture} &
    \fullimage{coral_reef_07/traj00_t120_instance_seg} \\

    \imagewithinput{Hopper_Nighthawks-1136x619/traj05_t120_rgb_recon.jpg}{Hopper_Nighthawks-1136x619/input.jpg} &
    \fullimage{Hopper_Nighthawks-1136x619/traj04_t120_rgb_recon.jpg} &
    \fullimage{Hopper_Nighthawks-1136x619/traj05_t120_semantic_pca.jpg} &
    \fullimage{Hopper_Nighthawks-1136x619/traj04_t120_instance_pca.jpg} &
    \fullimage{Hopper_Nighthawks-1136x619/traj05_t120_open_vocab_seg} &
    \fullimage{Hopper_Nighthawks-1136x619/traj04_t120_instance_seg}
    \\[-2.5pt]

    \fullimage{Hopper_Nighthawks-1136x619/traj03_t120_rgb_recon.jpg} &
    \fullimage{Hopper_Nighthawks-1136x619/traj00_t120_rgb_recon.jpg} &
    \fullimage{Hopper_Nighthawks-1136x619/traj03_t120_semantic_pca.jpg} &
    \fullimage{Hopper_Nighthawks-1136x619/traj00_t120_instance_pca.jpg} &
    \begin{tikzpicture}[x=\figpanelwidth,y=\figpanelheight,baseline=0pt]
        \node[anchor=south west,inner sep=0] at (0,0) {%
            \fullimage{Hopper_Nighthawks-1136x619/traj03_t120_open_vocab_seg}%
        };
        \begin{scope}[overlay,scale=2,transform shape]
            \fill[white,fill opacity=0.85] (0.00937,0.01705) rectangle (0.11875,0.22017);
            \node[anchor=south west,inner sep=0] at (0,0) {%
                \fullimage{Hopper_Nighthawks-1136x619/semantic_legend.png}%
            };
        \end{scope}
    \end{tikzpicture} &
    \fullimage{Hopper_Nighthawks-1136x619/traj00_t120_instance_seg} \\

    \imagewithinput{hotel_hallway_02/traj05_t120_rgb_recon.jpg}{hotel_hallway_02/input.jpg} &
    \fullimage{hotel_hallway_02/traj04_t120_rgb_recon.jpg} &
    \fullimage{hotel_hallway_02/traj05_t120_semantic_pca.jpg} &
    \fullimage{hotel_hallway_02/traj04_t120_instance_pca.jpg} &
    \fullimage{hotel_hallway_02/traj05_t120_open_vocab_seg} &
    \fullimage{hotel_hallway_02/traj04_t120_instance_seg}
    \\[-2.5pt]

    \fullimage{hotel_hallway_02/traj03_t120_rgb_recon.jpg} &
    \fullimage{hotel_hallway_02/traj00_t120_rgb_recon.jpg} &
    \fullimage{hotel_hallway_02/traj03_t120_semantic_pca.jpg} &
    \fullimage{hotel_hallway_02/traj00_t120_instance_pca.jpg} &
    \begin{tikzpicture}[x=\figpanelwidth,y=\figpanelheight,baseline=0pt]
        \node[anchor=south west,inner sep=0] at (0,0) {%
            \fullimage{hotel_hallway_02/traj03_t120_open_vocab_seg}%
        };
        \begin{scope}[overlay,scale=2,transform shape]
            \fill[white,fill opacity=0.85] (0.00937,0.01705) rectangle (0.15234,0.36222);
            \node[anchor=south west,inner sep=0] at (0,0) {%
                \fullimage{hotel_hallway_02/semantic_legend.png}%
            };
        \end{scope}
    \end{tikzpicture} &
    \fullimage{hotel_hallway_02/traj00_t120_instance_seg} \\

    \imagewithinput{mouse_cheese_skyscraper_02/traj05_t120_rgb_recon.jpg}{mouse_cheese_skyscraper_02/input.jpg} &
    \fullimage{mouse_cheese_skyscraper_02/traj04_t120_rgb_recon.jpg} &
    \fullimage{mouse_cheese_skyscraper_02/traj05_t120_semantic_pca.jpg} &
    \fullimage{mouse_cheese_skyscraper_02/traj04_t120_instance_pca.jpg} &
    \fullimage{mouse_cheese_skyscraper_02/traj05_t120_open_vocab_seg} &
    \fullimage{mouse_cheese_skyscraper_02/traj04_t120_instance_seg}
    \\[-2.5pt]

    \fullimage{mouse_cheese_skyscraper_02/traj03_t120_rgb_recon.jpg} &
    \fullimage{mouse_cheese_skyscraper_02/traj00_t120_rgb_recon.jpg} &
    \fullimage{mouse_cheese_skyscraper_02/traj03_t120_semantic_pca.jpg} &
    \fullimage{mouse_cheese_skyscraper_02/traj00_t120_instance_pca.jpg} &
    \begin{tikzpicture}[x=\figpanelwidth,y=\figpanelheight,baseline=0pt]
        \node[anchor=south west,inner sep=0] at (0,0) {%
            \fullimage{mouse_cheese_skyscraper_02/traj03_t120_open_vocab_seg}%
        };
        \begin{scope}[overlay,scale=2,transform shape]
            \fill[white,fill opacity=0.85] (0.00937,0.01705) rectangle (0.16094,0.22017);
            \node[anchor=south west,inner sep=0] at (0,0) {%
                \fullimage{mouse_cheese_skyscraper_02/semantic_legend.png}%
            };
        \end{scope}
    \end{tikzpicture} &
    \fullimage{mouse_cheese_skyscraper_02/traj00_t120_instance_seg} \\

    \imagewithinput{san_francisco_victorian_house_12/traj05_t120_rgb_recon.jpg}{san_francisco_victorian_house_12/input.jpg} &
    \fullimage{san_francisco_victorian_house_12/traj04_t120_rgb_recon.jpg} &
    \fullimage{san_francisco_victorian_house_12/traj05_t120_semantic_pca.jpg} &
    \fullimage{san_francisco_victorian_house_12/traj04_t120_instance_pca.jpg} &
    \fullimage{san_francisco_victorian_house_12/traj05_t120_open_vocab_seg} &
    \fullimage{san_francisco_victorian_house_12/traj04_t120_instance_seg}
    \\[-2.5pt]

    \fullimage{san_francisco_victorian_house_12/traj03_t120_rgb_recon.jpg} &
    \fullimage{san_francisco_victorian_house_12/traj00_t120_rgb_recon.jpg} &
    \fullimage{san_francisco_victorian_house_12/traj03_t120_semantic_pca.jpg} &
    \fullimage{san_francisco_victorian_house_12/traj00_t120_instance_pca.jpg} &
    \begin{tikzpicture}[x=\figpanelwidth,y=\figpanelheight,baseline=0pt]
        \node[anchor=south west,inner sep=0] at (0,0) {%
            \fullimage{san_francisco_victorian_house_12/traj03_t120_open_vocab_seg}%
        };
        \begin{scope}[overlay,scale=2,transform shape]
            \fill[white,fill opacity=0.85] (0.00937,0.01705) rectangle (0.11875,0.19176);
            \node[anchor=south west,inner sep=0] at (0,0) {%
                \fullimage{san_francisco_victorian_house_12/semantic_legend.png}%
            };
        \end{scope}
    \end{tikzpicture} &
    \fullimage{san_francisco_victorian_house_12/traj00_t120_instance_seg} \\

    \imagewithinput{wiping_coffee_table_06/traj05_t120_rgb_recon.jpg}{wiping_coffee_table_06/input.jpg} &
    \fullimage{wiping_coffee_table_06/traj04_t120_rgb_recon.jpg} &
    \fullimage{wiping_coffee_table_06/traj05_t120_semantic_pca.jpg} &
    \fullimage{wiping_coffee_table_06/traj04_t120_instance_pca.jpg} &
    \fullimage{wiping_coffee_table_06/traj05_t120_open_vocab_seg} &
    \fullimage{wiping_coffee_table_06/traj04_t120_instance_seg}
    \\[-2.5pt]

    \fullimage{wiping_coffee_table_06/traj03_t120_rgb_recon.jpg} &
    \fullimage{wiping_coffee_table_06/traj00_t120_rgb_recon.jpg} &
    \fullimage{wiping_coffee_table_06/traj03_t120_semantic_pca.jpg} &
    \fullimage{wiping_coffee_table_06/traj00_t120_instance_pca.jpg} &
    \begin{tikzpicture}[x=\figpanelwidth,y=\figpanelheight,baseline=0pt]
        \node[anchor=south west,inner sep=0] at (0,0) {%
            \fullimage{wiping_coffee_table_06/traj03_t120_open_vocab_seg}%
        };
        \begin{scope}[overlay,scale=2,transform shape]
            \fill[white,fill opacity=0.85] (0.00937,0.01705) rectangle (0.15078,0.41903);
            \node[anchor=south west,inner sep=0] at (0,0) {%
                \fullimage{wiping_coffee_table_06/semantic_legend.png}%
            };
        \end{scope}
    \end{tikzpicture} &
    \fullimage{wiping_coffee_table_06/traj00_t120_instance_seg} \\

\end{tabular}

    \caption{\textbf{\MethodName qualitative results.} We visualize input- and novel-view 3DGS~\citep{kerbl:2023:3gsf} renderings across RGB, semantic/instance feature PCAs, and open-vocabulary/instance segmentation for generated input images.}
    \label{fig:supp-static-examples}
\end{figure}

\begin{figure}[p]
    \centering
    \input{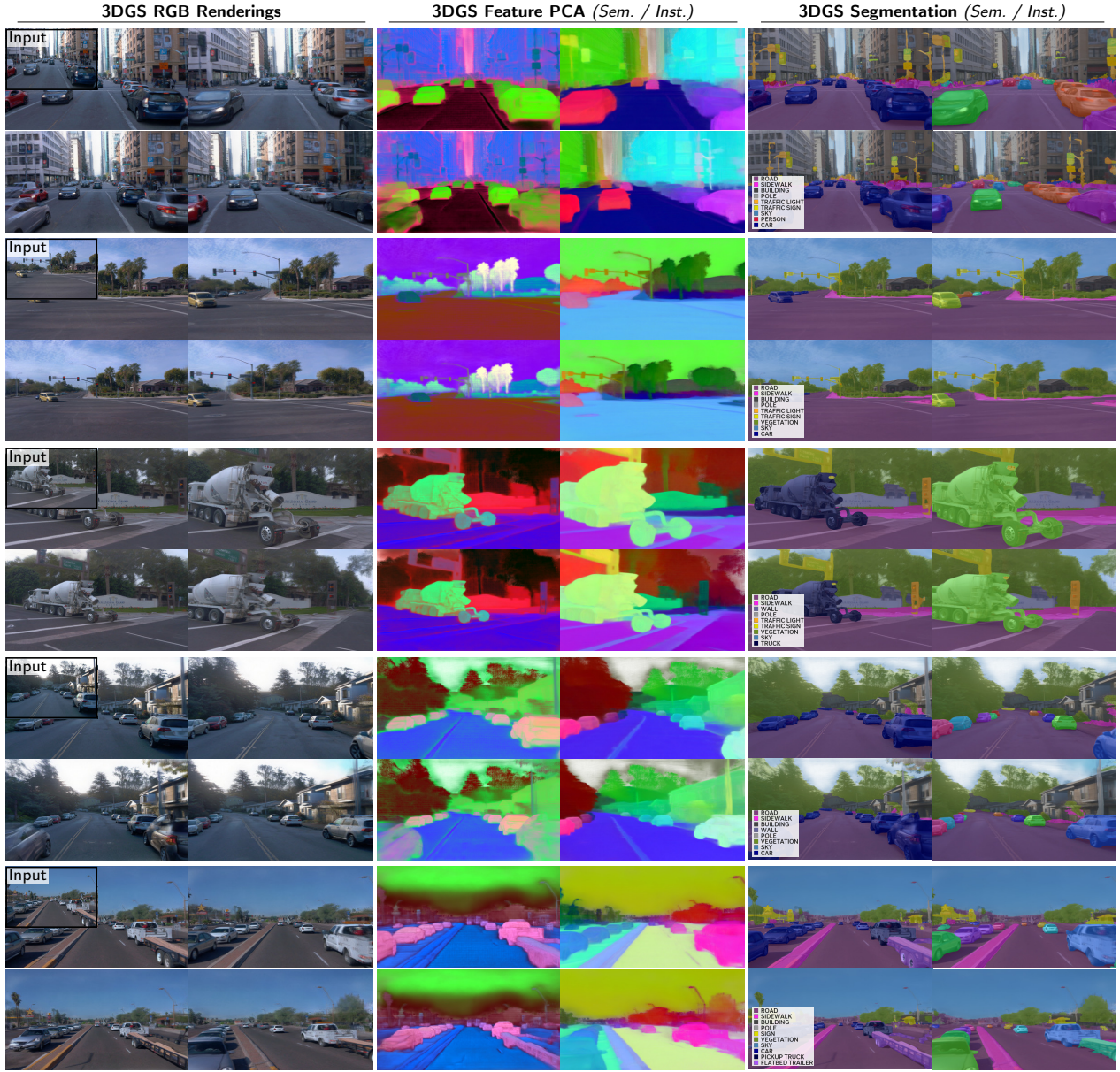}
    \caption{\textbf{Waymo qualitative examples.}
    Each two-row block shows one scene from four viewpoints. Paired columns
    show RGB, semantic/instance feature PCA, and semantic/instance segmentation.
    Insets show the original input images.}
    \label{fig:supp-qualitative-waymo}
\end{figure}

\Cref{fig:supp-static-examples} shows qualitative \MethodName scene generation and understanding across diverse generated examples. 
The qualitative results in \cref{fig:supp-static-examples,fig:supp-dynamic-examples} show crisp reconstructions and coherent open-vocabulary and instance segmentation across diverse scenes and queried categories, while the feature PCA visualizations suggest clear separation of semantic concepts and object instances in the learned representation.
\Cref{fig:supp-dynamic-examples} shows dynamic qualitative \MethodName results from generated inputs. Dynamic \MethodName is able to reconstruct the motion and provides temporally consistent open-vocabulary and instance segmentation.
\Cref{fig:supp-qualitative-waymo} shows qualitative transfer to driving scenes conditioned on real images from the Waymo Open Dataset \citep{sun:2020:sipf}. \MethodName generalizes well to the real-image conditioned driving scenes.

\begin{figure}[!t]
    \centering
    \input{figures/supplement/dynamic_examples/figure}
    \caption{\textbf{Dynamic \MethodName qualitative results.} We visualize input- and novel-view 3DGS renderings across RGB, semantic/instance feature PCAs, and open-vocabulary/instance segmentation for generated input videos.}
    \label{fig:supp-dynamic-examples}
\end{figure}

\subsection{Qualitative Baseline Comparison}
\label{ssec:supp-qualitative-comparison}

\begin{figure}[p]
    \centering
    
\scriptsize
\sffamily
\setlength{\tabcolsep}{0pt}
\newcommand{\imgwidth}{0.133}

\begin{tabular}{
>{\centering\arraybackslash} m{0.05\textwidth}
@{\hspace{1pt}}  
>{\centering\arraybackslash} m{\imgwidth\textwidth}
>{\centering\arraybackslash} m{\imgwidth\textwidth}
>{\centering\arraybackslash} m{\imgwidth\textwidth}
@{\hspace{4pt}}                                    
>{\centering\arraybackslash} m{\imgwidth\textwidth}
>{\centering\arraybackslash} m{\imgwidth\textwidth}
>{\centering\arraybackslash} m{\imgwidth\textwidth}
}

&\multicolumn{6}{c}{\rule[0.5ex]{3.6cm}{0.4pt}\textbf{~Open-Vocabulary Semantic Segmentation~}\rule[0.5ex]{3.6cm}{0.4pt}}\\

\rotatebox[origin=lB]{90}{\hspace{-0.25em}{RGB}}
& \includegraphics[width=\linewidth]{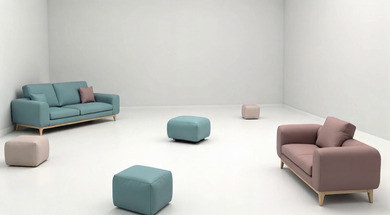}
& \includegraphics[width=\linewidth]{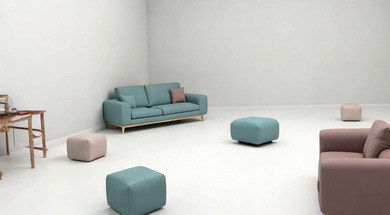}
& \includegraphics[width=\linewidth]{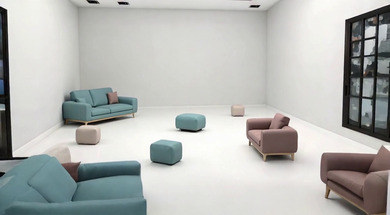}
& \includegraphics[width=\linewidth]{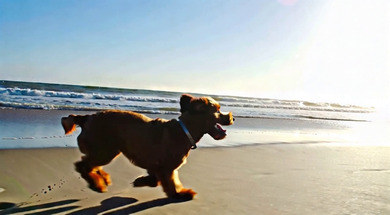}
& \includegraphics[width=\linewidth]{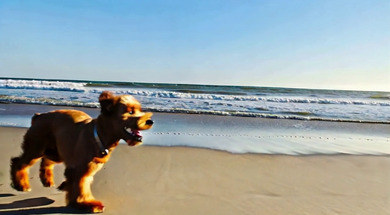}
& \includegraphics[width=\linewidth]{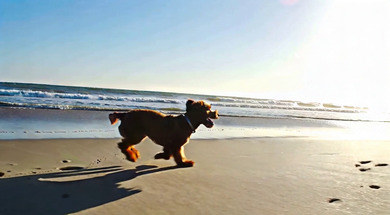}\\[-2.25pt]

\rotatebox[origin=lB]{90}{\hspace{-0.25em}{\shortstack{Ground\\[-1pt]Truth}}}
& \includegraphics[width=\linewidth]{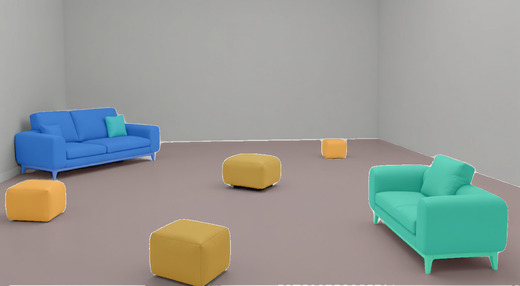}
& \includegraphics[width=\linewidth]{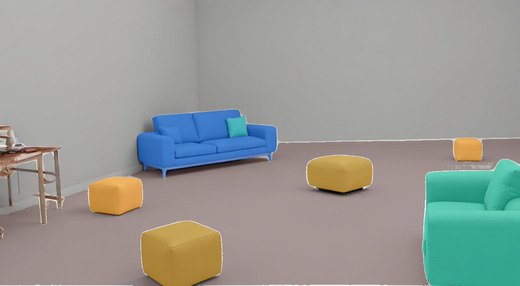}
& \includegraphics[width=\linewidth]{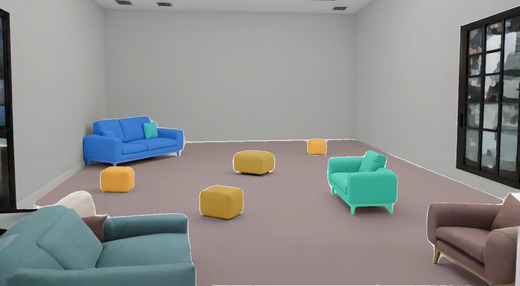}
& \includegraphics[width=\linewidth]{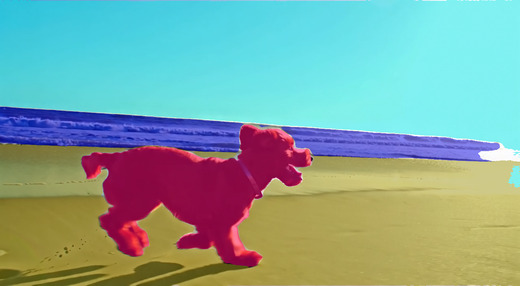}
& \includegraphics[width=\linewidth]{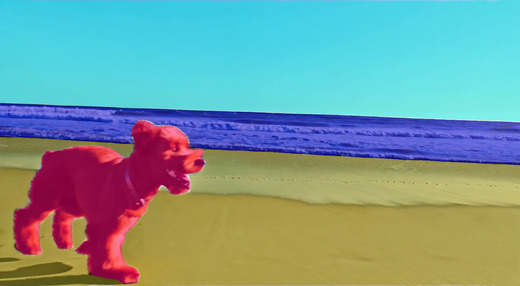}
& \includegraphics[width=\linewidth]{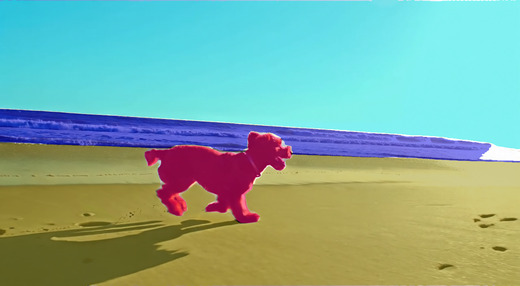}\\[-2.25pt]

\rotatebox[origin=lB]{90}{\hspace{-0.25em}{\shortstack{Lyra~+\\[-1pt]Lifted\\[-1pt]Teachers}}}
& \includegraphics[width=\linewidth]{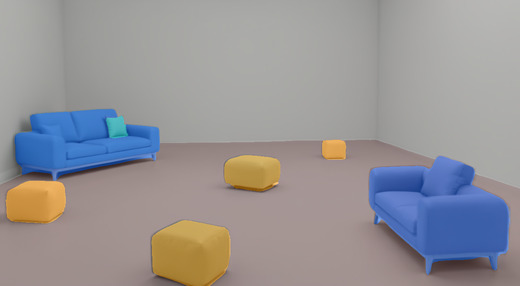}
& \includegraphics[width=\linewidth]{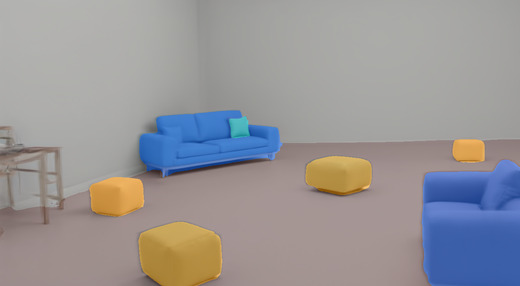}
& \includegraphics[width=\linewidth]{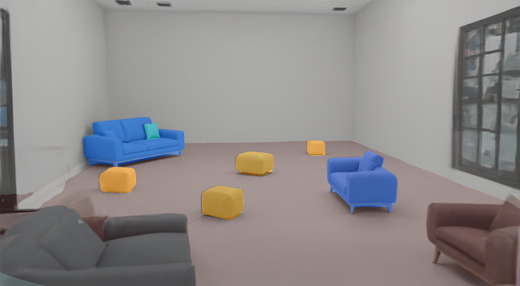}
& \includegraphics[width=\linewidth]{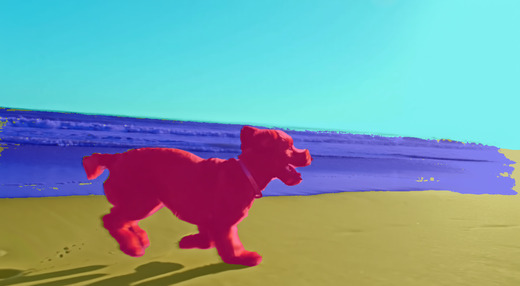}
& \includegraphics[width=\linewidth]{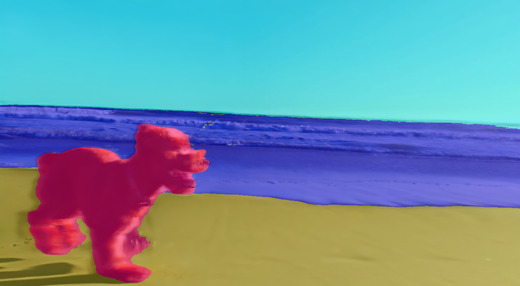}
& \includegraphics[width=\linewidth]{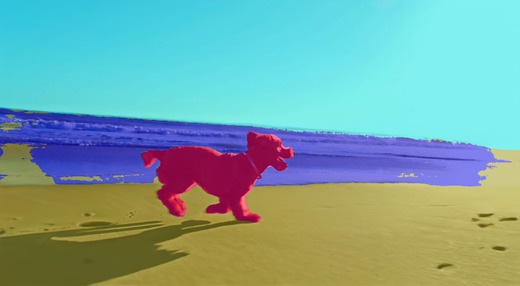}\\[-2.25pt]

\rotatebox[origin=lB]{90}{\hspace{-0.25em}{\shortstack{Lyra~+\\[-1pt]LangSplat\\[-1pt]V2}}}
& \includegraphics[width=\linewidth]{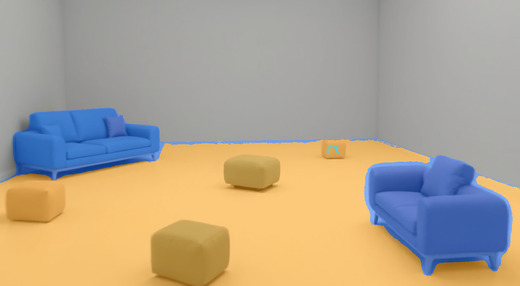}
& \includegraphics[width=\linewidth]{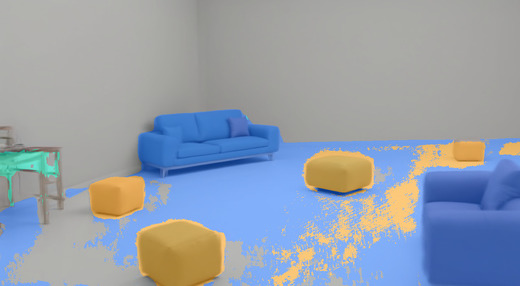}
& \includegraphics[width=\linewidth]{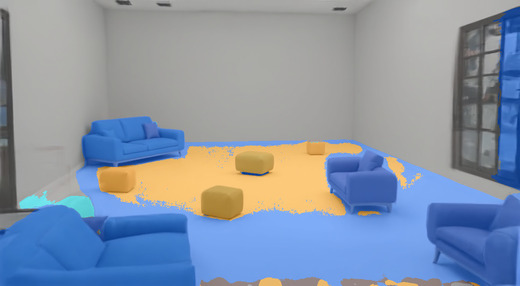}
& \includegraphics[width=\linewidth]{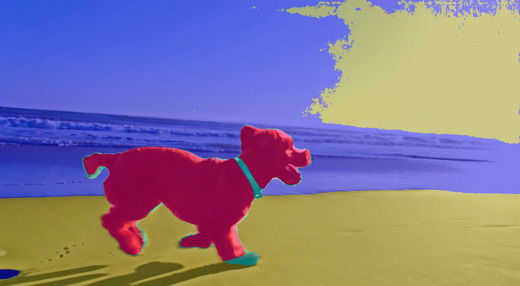}
& \includegraphics[width=\linewidth]{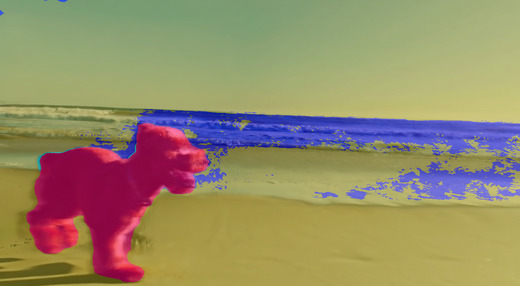}
& \includegraphics[width=\linewidth]{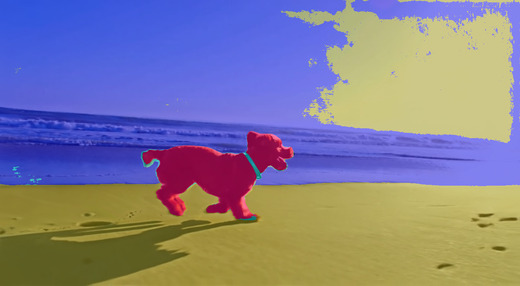}\\[-2.25pt]

\rotatebox[origin=lB]{90}{\hspace{-0.25em}{\shortstack{Lyra~+\\[-1pt]SFS}}}
& \includegraphics[width=\linewidth]{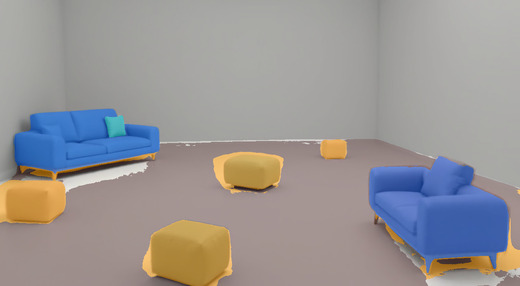}
& \includegraphics[width=\linewidth]{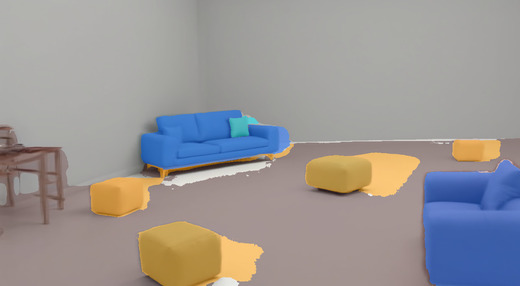}
& \includegraphics[width=\linewidth]{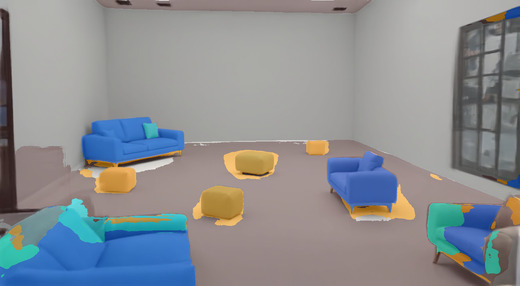}
& \includegraphics[width=\linewidth]{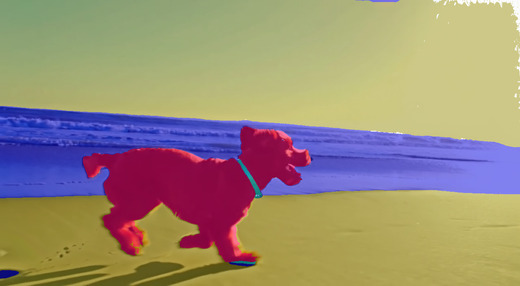}
& \includegraphics[width=\linewidth]{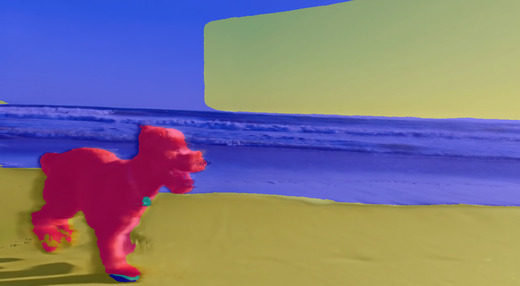}
& \includegraphics[width=\linewidth]{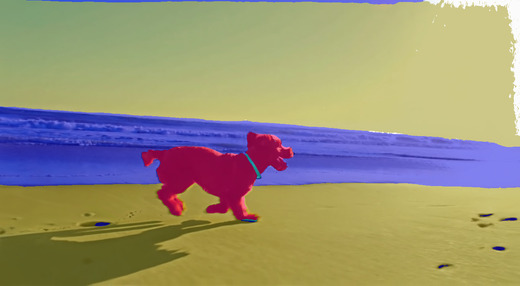}\\[-2.25pt]

\rotatebox[origin=lB]{90}{\hspace{-0.25em}{\shortstack{Lyra~+\\[-1pt]Chorus}}}
& \includegraphics[width=\linewidth]{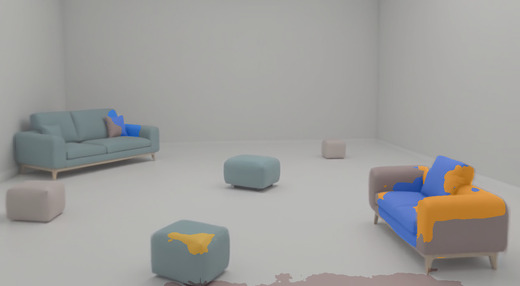}
& \includegraphics[width=\linewidth]{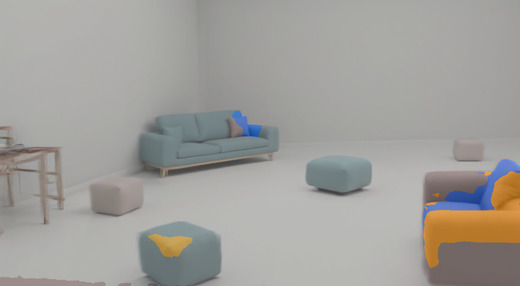}
& \includegraphics[width=\linewidth]{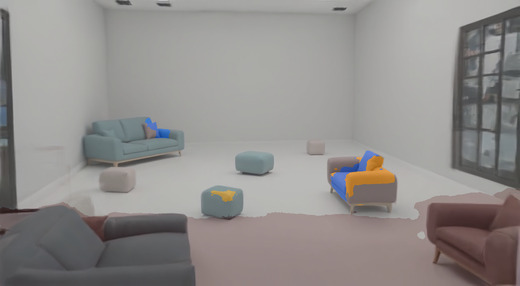}
& \includegraphics[width=\linewidth]{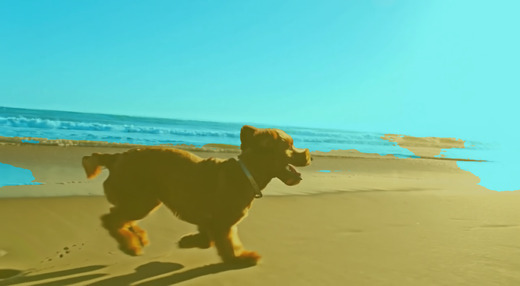}
& \includegraphics[width=\linewidth]{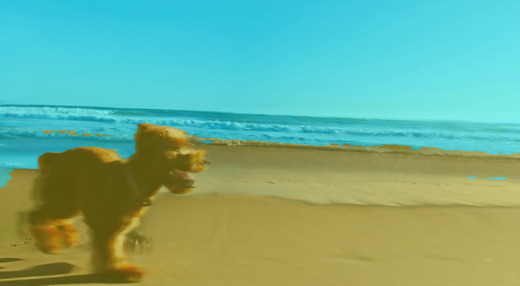}
& \includegraphics[width=\linewidth]{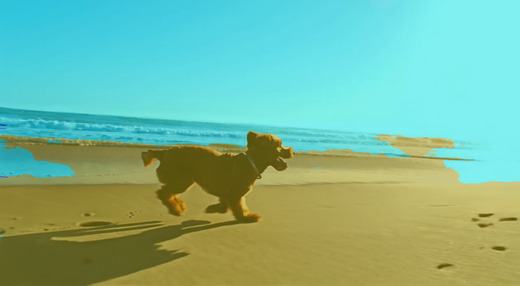}\\[-2.25pt]

\rotatebox[origin=lB]{90}{\hspace{-0.25em}{\shortstack{Lyra~+\\[-1pt]IGGT}}}
& \includegraphics[width=\linewidth]{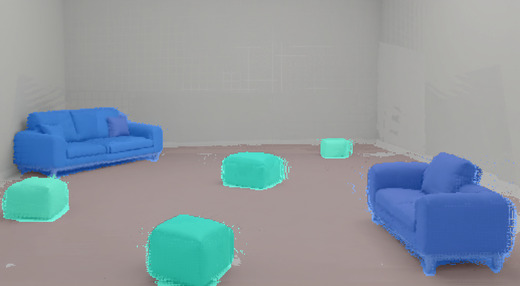}
& \includegraphics[width=\linewidth]{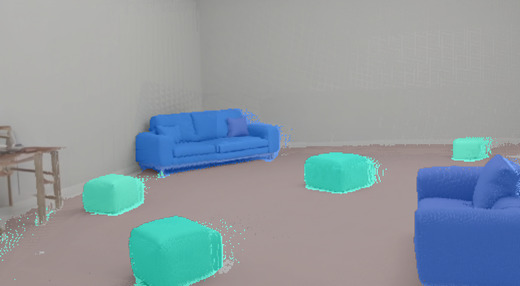}
& \includegraphics[width=\linewidth]{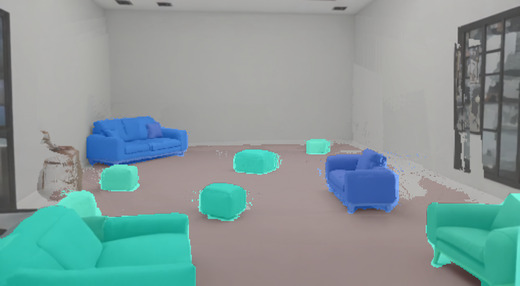}
& \includegraphics[width=\linewidth]{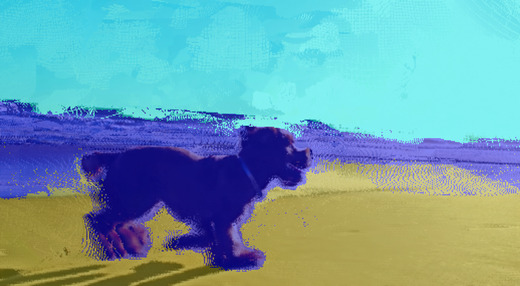}
& \includegraphics[width=\linewidth]{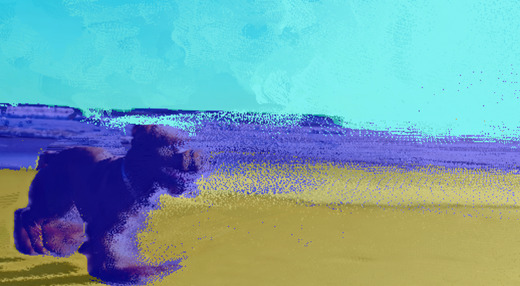}
& \includegraphics[width=\linewidth]{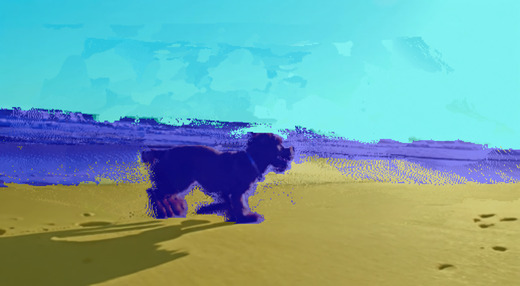}\\[-2.25pt]

\rotatebox[origin=lB]{90}{\hspace{-0.25em}{\MethodName}}
& \includegraphics[width=\linewidth]{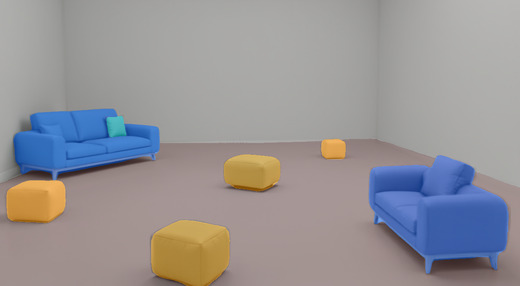}
& \includegraphics[width=\linewidth]{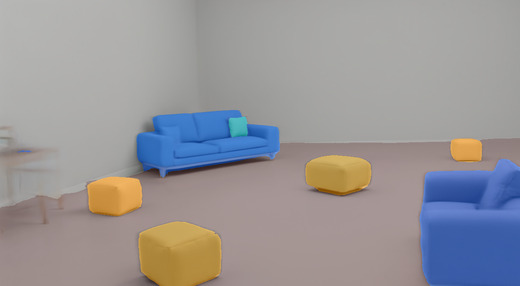}
& \includegraphics[width=\linewidth]{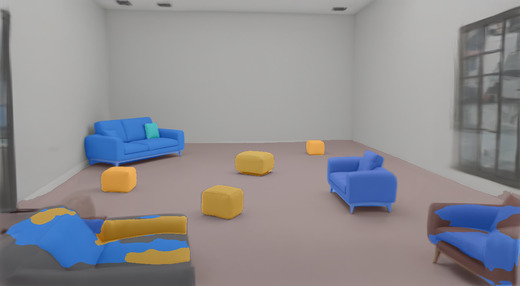}
& \includegraphics[width=\linewidth]{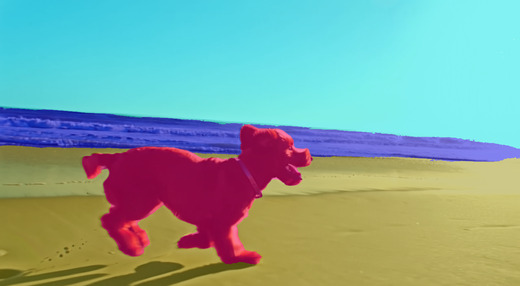}
& \includegraphics[width=\linewidth]{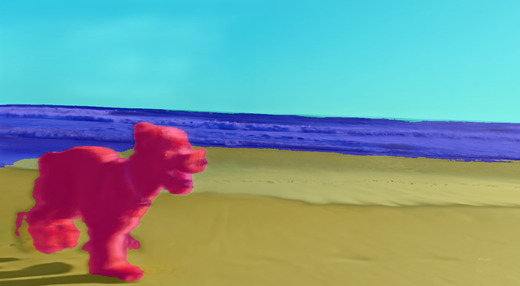}
& \includegraphics[width=\linewidth]{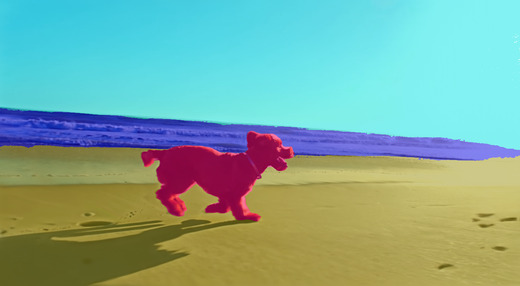}\\[-2.25pt]

\rotatebox[origin=lB]{90}{\hspace{-0.25em}{\MethodNamePlus}}
& \includegraphics[width=\linewidth]{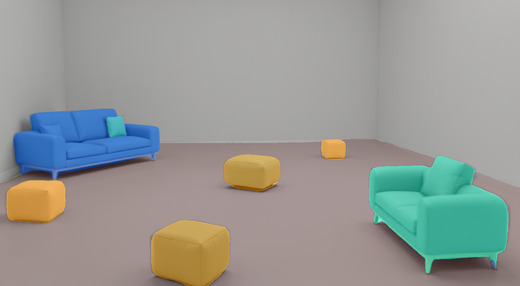}
& \includegraphics[width=\linewidth]{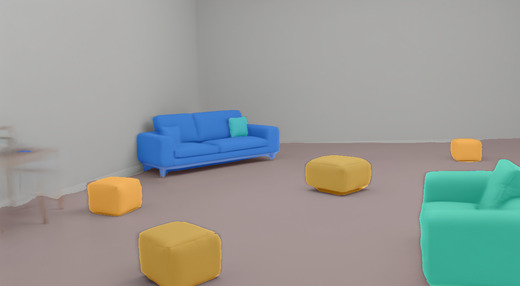}
& \includegraphics[width=\linewidth]{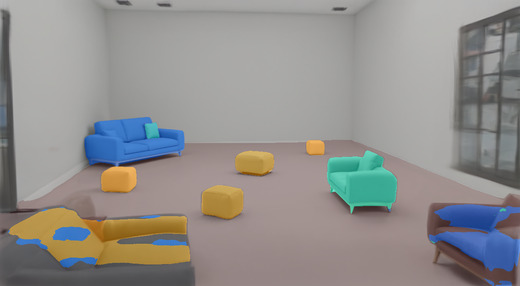}
& \includegraphics[width=\linewidth]{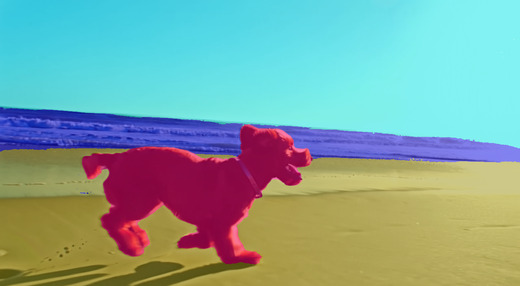}
& \includegraphics[width=\linewidth]{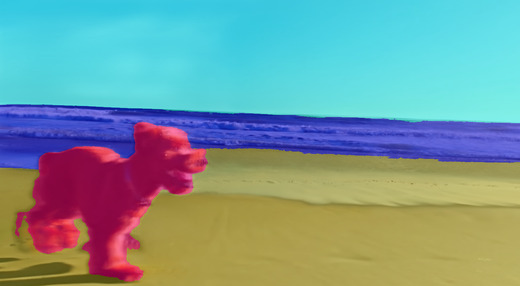}
& \includegraphics[width=\linewidth]{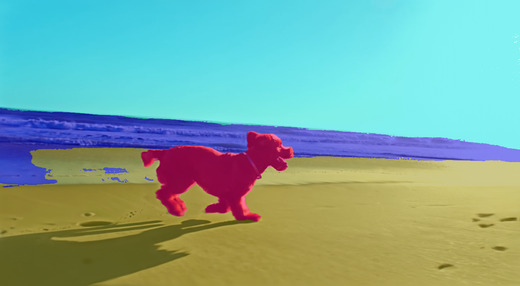}\\[1pt]

& \multicolumn{3}{@{}c@{\hspace{4pt}}}{%
    \begin{minipage}[t]{0.399\textwidth}%
    \vspace{0pt}%
    \begingroup
    \scriptsize
    \sffamily
    \setlength{\tabcolsep}{0pt}%
    \renewcommand{\arraystretch}{0.9}%
    \begin{tabularx}{\linewidth}{*{6}{>{\centering\arraybackslash}X}}
        \cellcolor{wall}\textcolor{white}{Wall} &
        \cellcolor{floor}\textcolor{white}{Floor} &
        \cellcolor{sofa}\textcolor{white}{Sofa} &
        \cellcolor{armchair}\textcolor{black}{Armchair} &
        \cellcolor{pillow}\textcolor{black}{Pillow} &
        \cellcolor{ottoman}\textcolor{black}{Ottoman} \\
    \end{tabularx}%
    \endgroup
    \end{minipage}%
}
& \multicolumn{3}{@{}c@{}}{%
    \begin{minipage}[t]{0.399\textwidth}%
    \vspace{0pt}%
    \begingroup
    \scriptsize
    \sffamily
    \setlength{\tabcolsep}{0pt}%
    \renewcommand{\arraystretch}{0.9}%
    \begin{tabularx}{\linewidth}{*{4}{>{\centering\arraybackslash}X}}
        \cellcolor{sky}\textcolor{black}{Sky} &
        \cellcolor{sea}\textcolor{white}{Sea} &
        \cellcolor{sand}\textcolor{white}{Sand} &
        \cellcolor{animal}\textcolor{white}{Animal} \\
    \end{tabularx}%
    \endgroup
    \end{minipage}
}\\

&\multicolumn{6}{c}{\rule[0.5ex]{4.5cm}{0.4pt}\textbf{~Instance Segmentation~}\rule[0.5ex]{4.5cm}{0.4pt}}\\

\rotatebox[origin=lB]{90}{\hspace{-0.25em}{RGB}}
& \includegraphics[width=\linewidth]{figures/supplement/lyra_eval_static_qualitative/3_0_00359/traj05_t120_rgb.jpg}
& \includegraphics[width=\linewidth]{figures/supplement/lyra_eval_static_qualitative/3_0_00359/traj01_t120_rgb.jpg}
& \includegraphics[width=\linewidth]{figures/supplement/lyra_eval_static_qualitative/3_0_00359/traj03_t120_rgb.jpg}
& \includegraphics[width=\linewidth]{figures/supplement/lyra_eval_static_qualitative/0_000026_353405e57b9649bcab936a1f3a8c269a/traj05_t120_rgb.jpg}
& \includegraphics[width=\linewidth]{figures/supplement/lyra_eval_static_qualitative/0_000026_353405e57b9649bcab936a1f3a8c269a/traj01_t120_rgb.jpg}
& \includegraphics[width=\linewidth]{figures/supplement/lyra_eval_static_qualitative/0_000026_353405e57b9649bcab936a1f3a8c269a/traj03_t120_rgb.jpg}\\[-2.25pt]

\rotatebox[origin=lB]{90}{\hspace{-0.25em}{\shortstack{Ground\\[-1pt]Truth}}}
& \includegraphics[width=\linewidth]{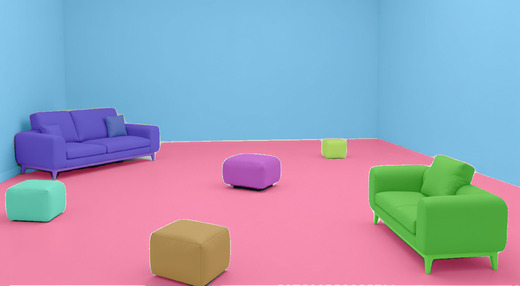}
& \includegraphics[width=\linewidth]{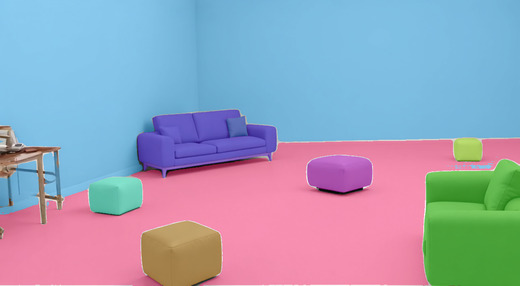}
& \includegraphics[width=\linewidth]{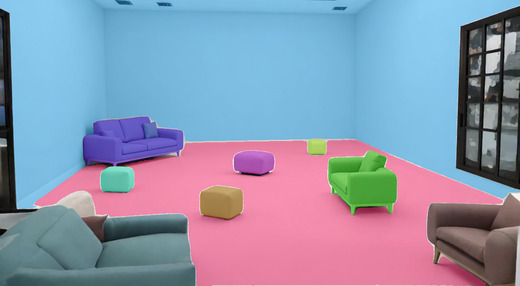}
& \includegraphics[width=\linewidth]{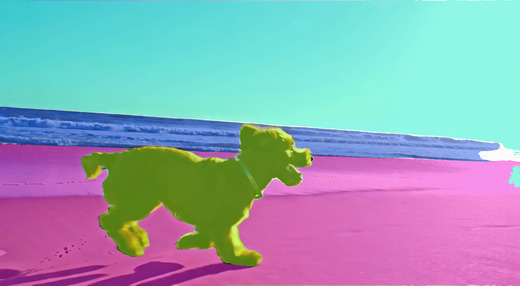}
& \includegraphics[width=\linewidth]{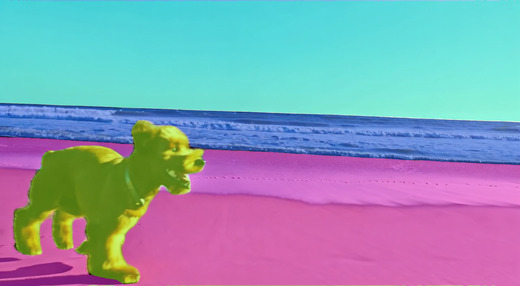}
& \includegraphics[width=\linewidth]{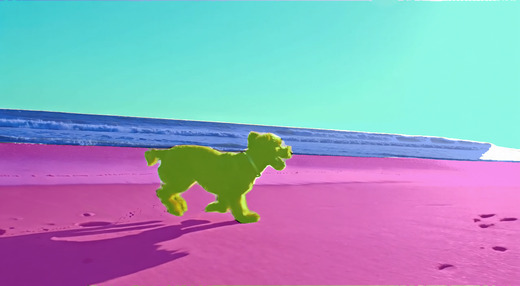}\\[-2.25pt]

\rotatebox[origin=lB]{90}{\hspace{-0.25em}{\shortstack{Lyra~+\\[-1pt]Lifted\\[-1pt]Teachers}}}
& \includegraphics[width=\linewidth]{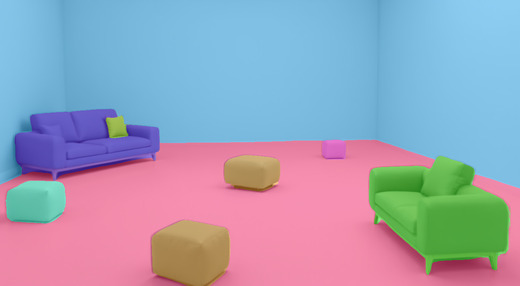}
& \includegraphics[width=\linewidth]{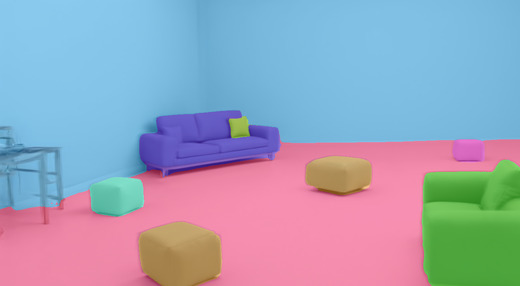}
& \includegraphics[width=\linewidth]{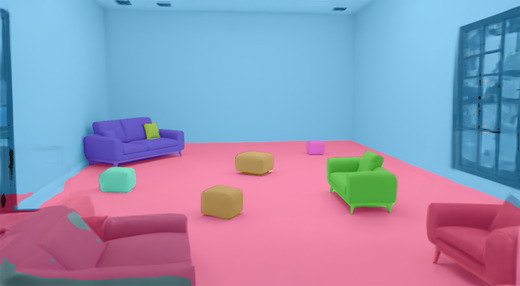}
& \includegraphics[width=\linewidth]{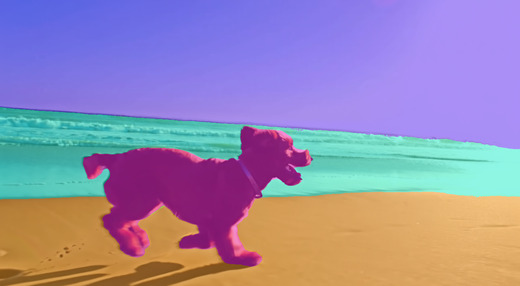}
& \includegraphics[width=\linewidth]{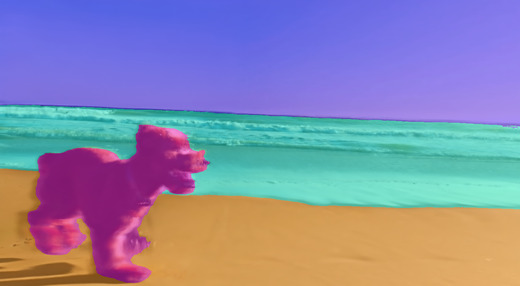}
& \includegraphics[width=\linewidth]{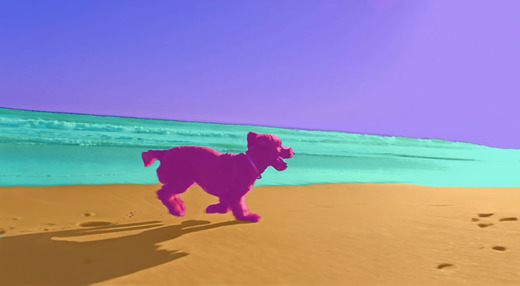}\\[-2.25pt]

\rotatebox[origin=lB]{90}{\hspace{-0.25em}{\shortstack{Lyra~+\\[-1pt]Unified-Lift}}}
& \includegraphics[width=\linewidth]{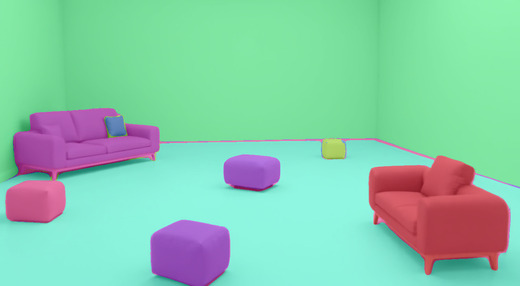}
& \includegraphics[width=\linewidth]{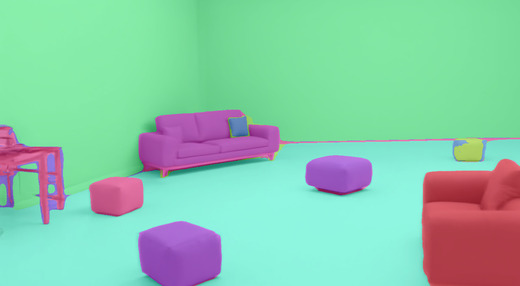}
& \includegraphics[width=\linewidth]{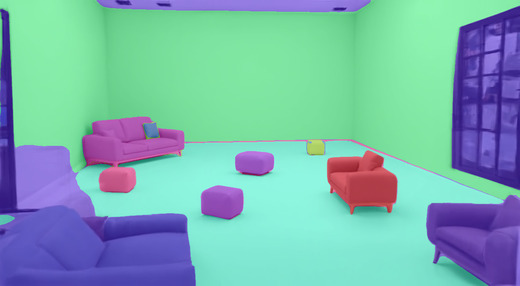}
& \includegraphics[width=\linewidth]{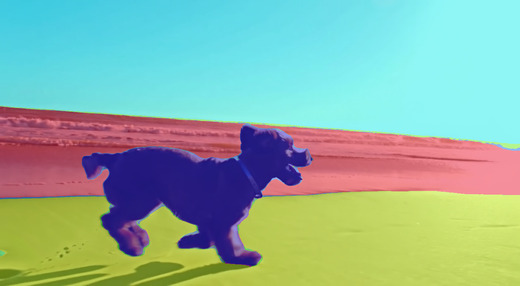}
& \includegraphics[width=\linewidth]{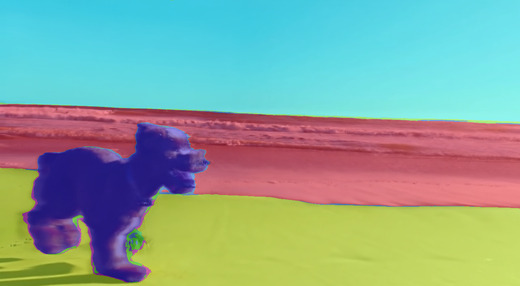}
& \includegraphics[width=\linewidth]{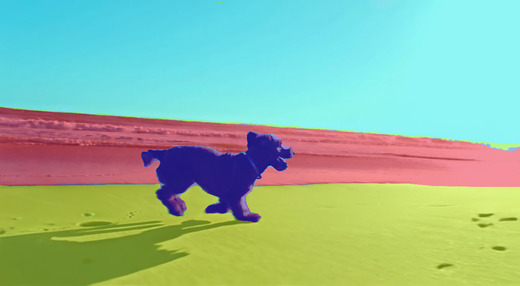}\\[-2.25pt]

\rotatebox[origin=lB]{90}{\hspace{-0.25em}{\shortstack{Lyra\textsuperscript{$\dagger$}~+\\[-1pt]IGGT}}}
& \includegraphics[width=\linewidth]{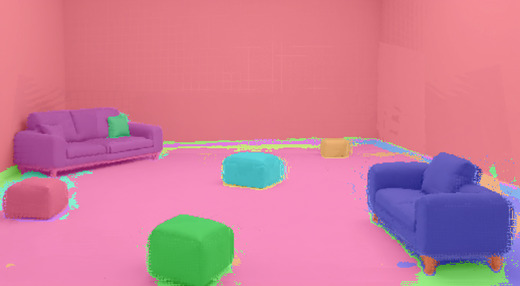}
& \includegraphics[width=\linewidth]{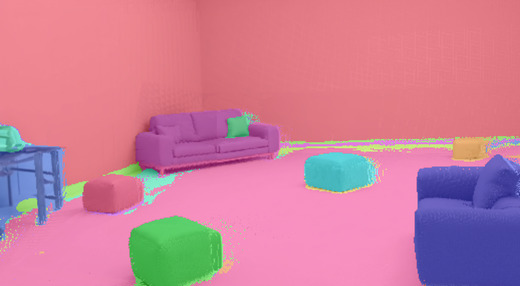}
& \includegraphics[width=\linewidth]{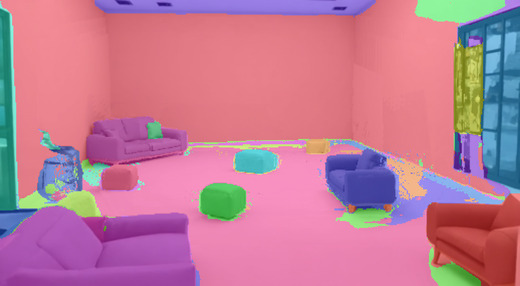}
& \includegraphics[width=\linewidth]{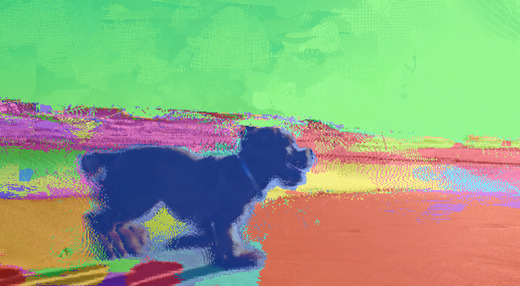}
& \includegraphics[width=\linewidth]{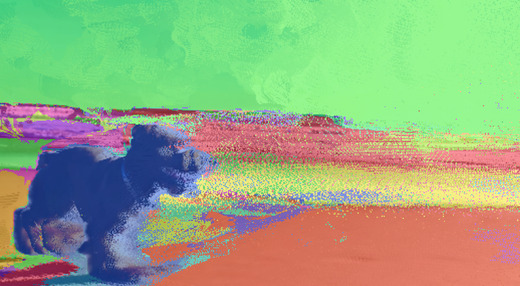}
& \includegraphics[width=\linewidth]{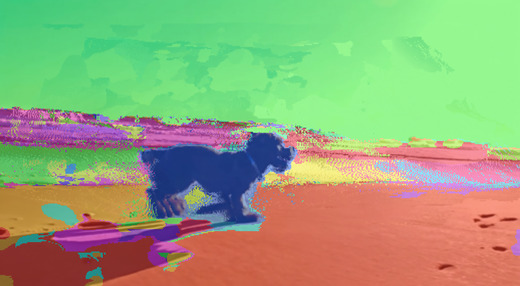}\\[-2.25pt]

\rotatebox[origin=lB]{90}{\hspace{-0.25em}{\MethodName}}
& \includegraphics[width=\linewidth]{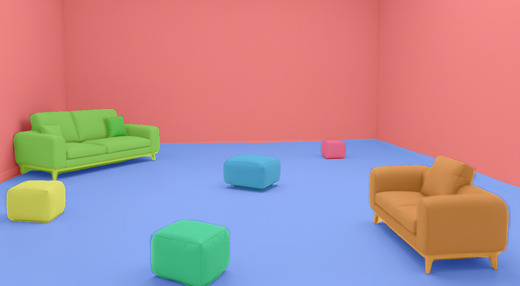}
& \includegraphics[width=\linewidth]{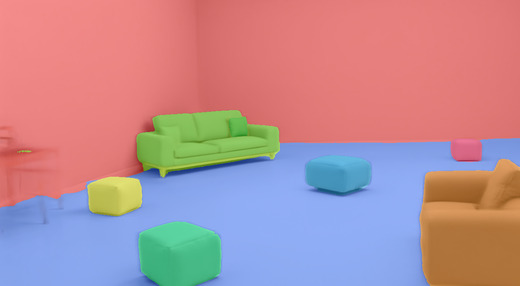}
& \includegraphics[width=\linewidth]{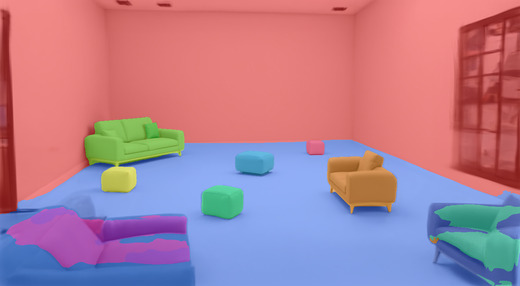}
& \includegraphics[width=\linewidth]{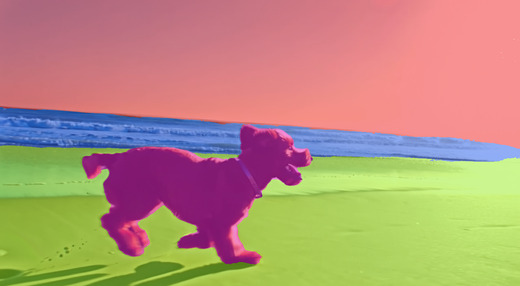}
& \includegraphics[width=\linewidth]{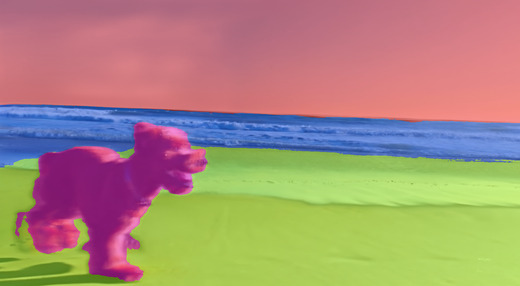}
& \includegraphics[width=\linewidth]{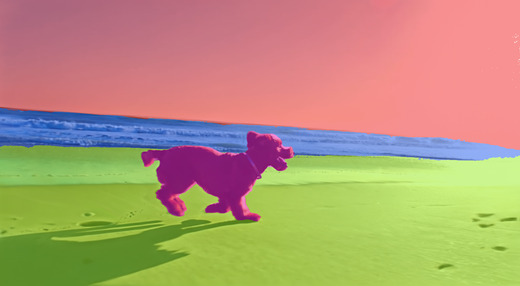}\\[-2.25pt]

\rotatebox[origin=lB]{90}{\hspace{-0.25em}{\MethodNamePlus}}
& \includegraphics[width=\linewidth]{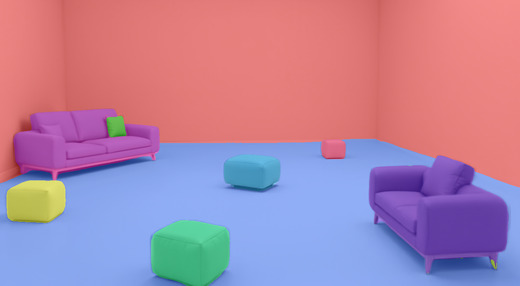}
& \includegraphics[width=\linewidth]{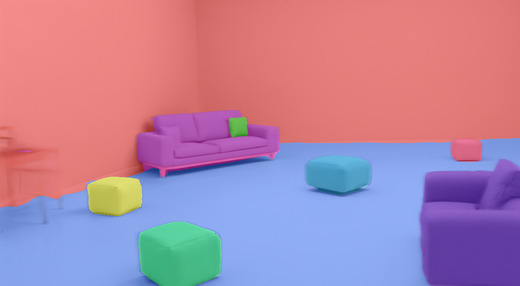}
& \includegraphics[width=\linewidth]{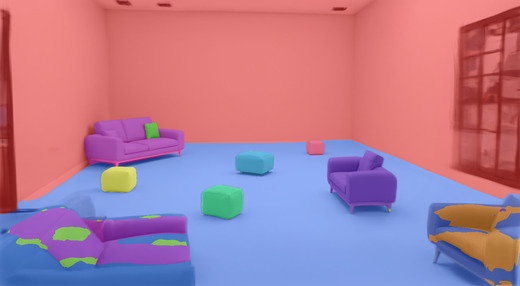}
& \includegraphics[width=\linewidth]{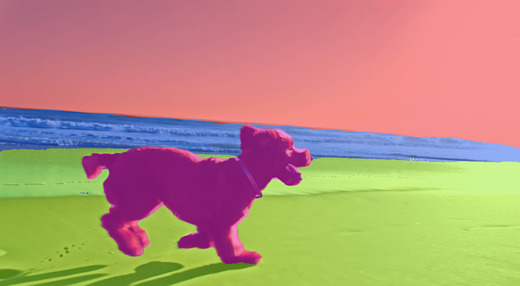}
& \includegraphics[width=\linewidth]{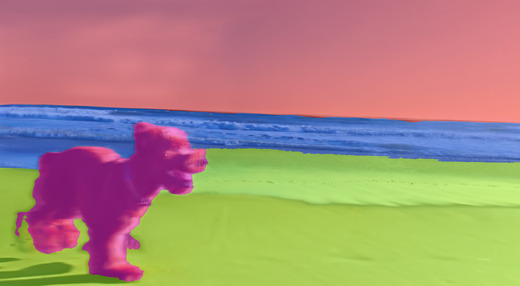}
& \includegraphics[width=\linewidth]{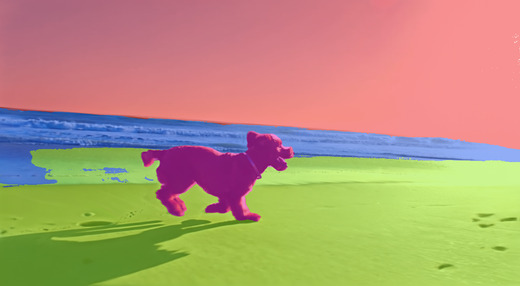}\\[-2.25pt]

\end{tabular}

    \caption{\textbf{Qualitative comparison on \MethodName-SDG Static eval.} We show three viewpoints per scene, with semantic predictions above and instance
    predictions below. $\dagger$ denotes
    directly using generated multi-view RGB~\citep{bahmani:2026:g3sr}.}
    \label{fig:supp-qualitative-sdg}
\end{figure}

\begin{figure}[p]
    \centering
    \scriptsize
\sffamily
\setlength{\tabcolsep}{0pt}
\newcommand{\imgwidth}{0.133}

\begin{tabular}{
>{\centering\arraybackslash} m{0.05\textwidth}
>{\centering\arraybackslash} m{\imgwidth\textwidth}
>{\centering\arraybackslash} m{\imgwidth\textwidth}
>{\centering\arraybackslash} m{\imgwidth\textwidth}
@{\hspace{4pt}}
>{\centering\arraybackslash} m{\imgwidth\textwidth}
>{\centering\arraybackslash} m{\imgwidth\textwidth}
>{\centering\arraybackslash} m{\imgwidth\textwidth}
}

&\multicolumn{6}{c}{\rule[0.5ex]{4.0cm}{0.4pt}\textbf{~Open-Vocabulary Semantic Segmentation~}\rule[0.5ex]{4.0cm}{0.4pt}}\\

\rotatebox[origin=lB]{90}{\hspace{-0.25em}{RGB}}
& \includegraphics[width=\linewidth]{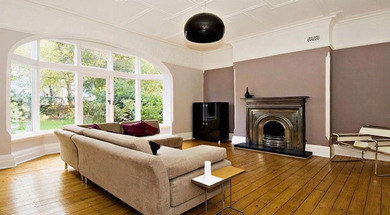}
& \includegraphics[width=\linewidth]{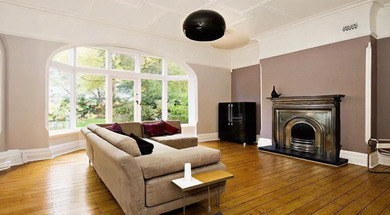}
& \includegraphics[width=\linewidth]{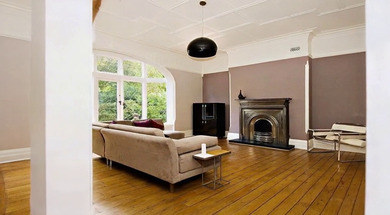}
& \includegraphics[width=\linewidth]{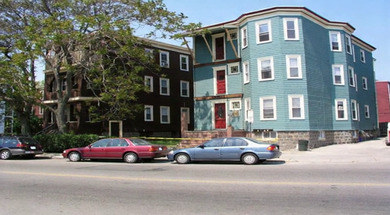}
& \includegraphics[width=\linewidth]{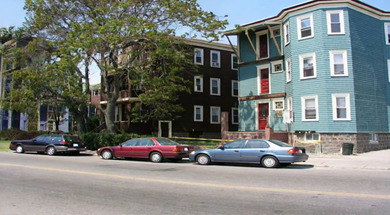}
& \includegraphics[width=\linewidth]{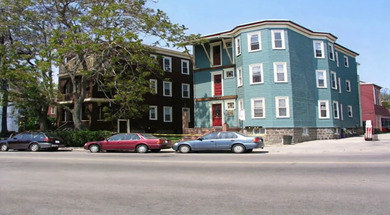}\\[-2.25pt]

\rotatebox[origin=lB]{90}{\hspace{-0.25em}{\shortstack{Ground\\[-1pt]Truth}}}
& \includegraphics[width=\linewidth]{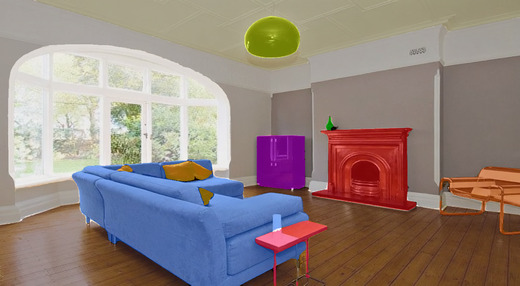}
& \includegraphics[width=\linewidth]{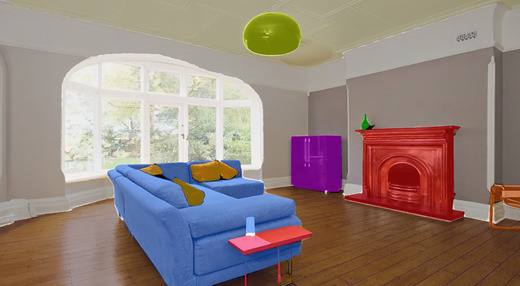}
& \includegraphics[width=\linewidth]{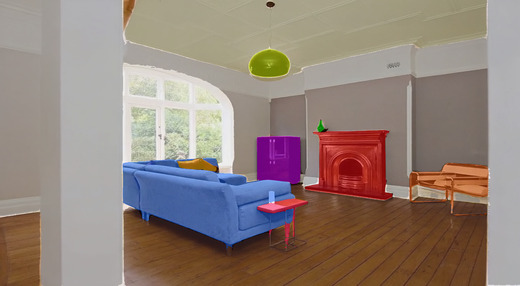}
& \includegraphics[width=\linewidth]{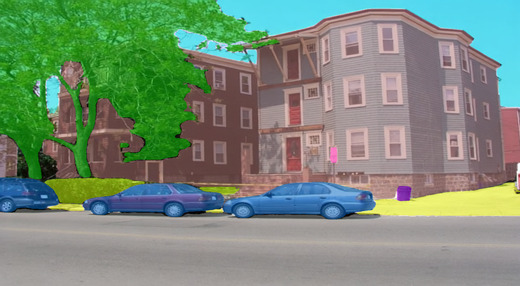}
& \includegraphics[width=\linewidth]{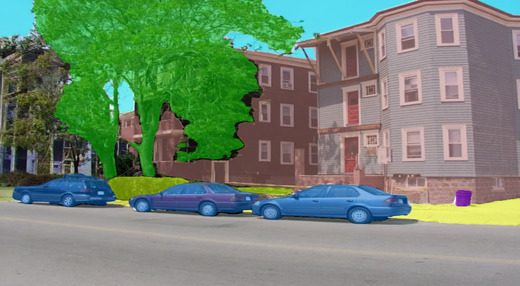}
& \includegraphics[width=\linewidth]{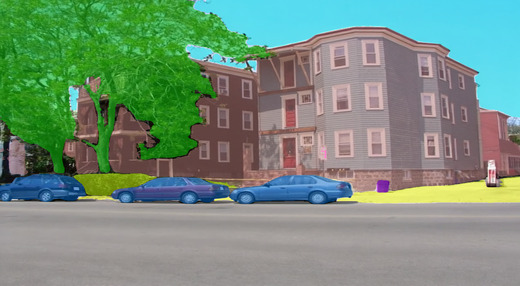}\\[-2.25pt]

\rotatebox[origin=lB]{90}{\hspace{-0.25em}{\shortstack{Lyra~+\\[-1pt]Lifted\\[-1pt]Teachers}}}
& \includegraphics[width=\linewidth]{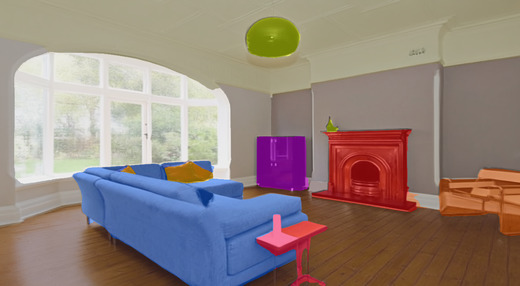}
& \includegraphics[width=\linewidth]{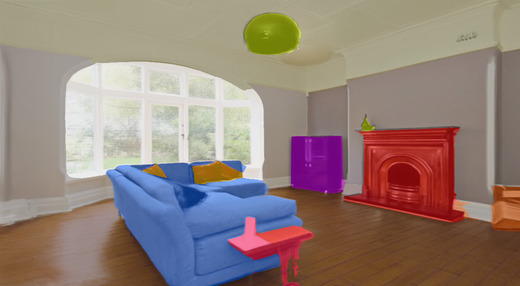}
& \includegraphics[width=\linewidth]{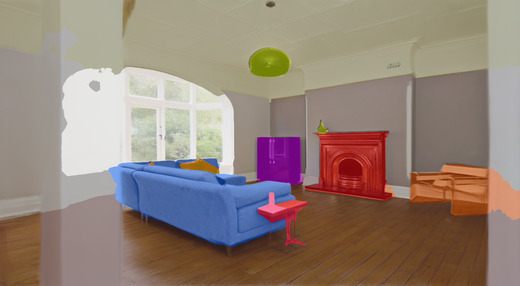}
& \includegraphics[width=\linewidth]{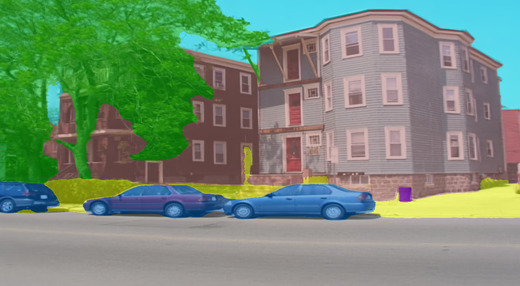}
& \includegraphics[width=\linewidth]{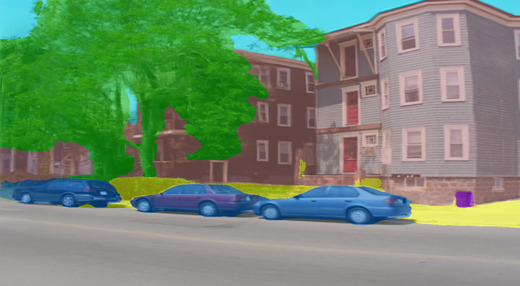}
& \includegraphics[width=\linewidth]{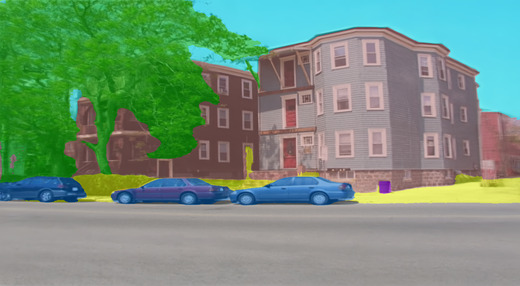}\\[-2.25pt]

\rotatebox[origin=lB]{90}{\hspace{-0.25em}{\shortstack{Lyra~+\\[-1pt]LangSplat\\[-1pt]V2}}}
& \includegraphics[width=\linewidth]{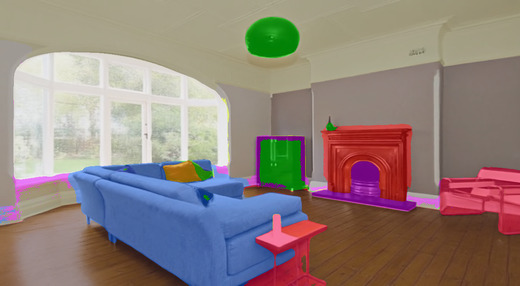}
& \includegraphics[width=\linewidth]{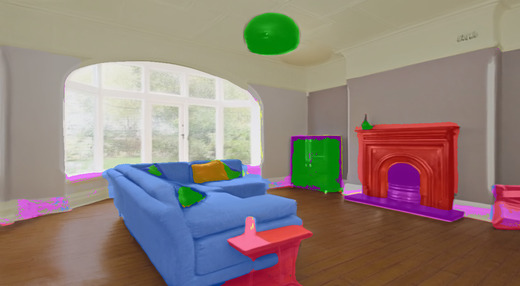}
& \includegraphics[width=\linewidth]{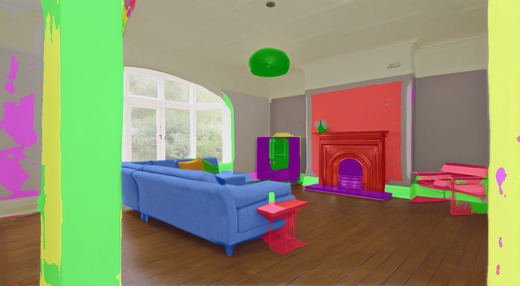}
& \includegraphics[width=\linewidth]{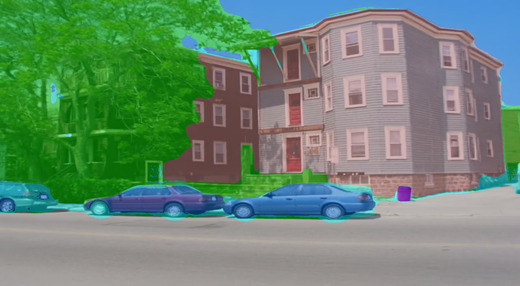}
& \includegraphics[width=\linewidth]{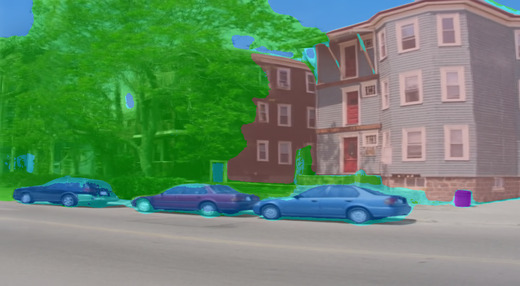}
& \includegraphics[width=\linewidth]{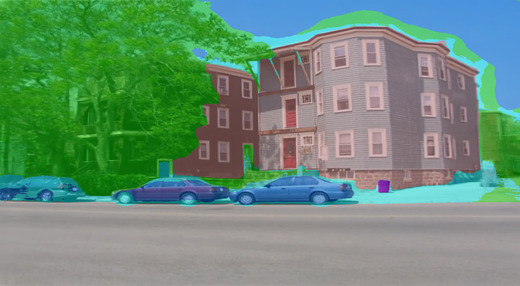}\\[-2.25pt]

\rotatebox[origin=lB]{90}{\hspace{-0.25em}{\shortstack{Lyra~+\\[-1pt]SFS}}}
& \includegraphics[width=\linewidth]{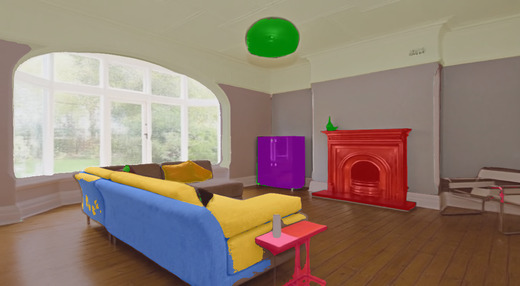}
& \includegraphics[width=\linewidth]{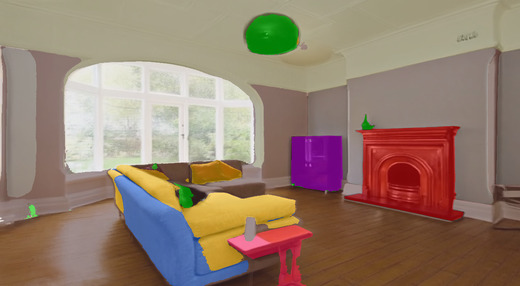}
& \includegraphics[width=\linewidth]{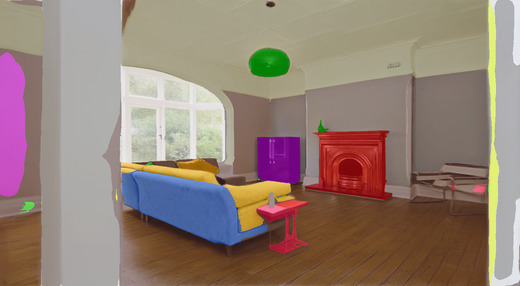}
& \includegraphics[width=\linewidth]{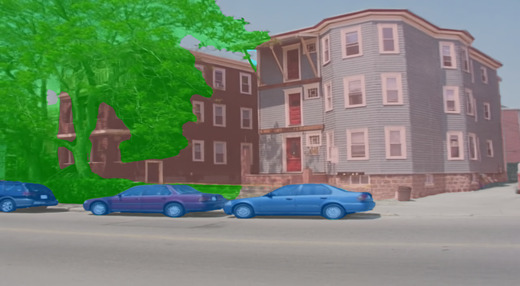}
& \includegraphics[width=\linewidth]{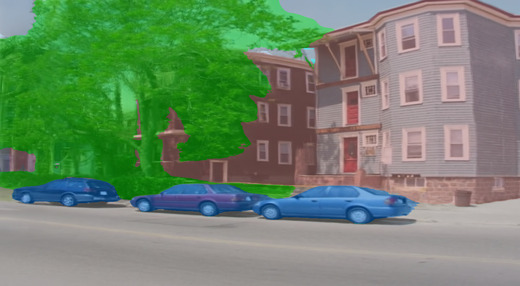}
& \includegraphics[width=\linewidth]{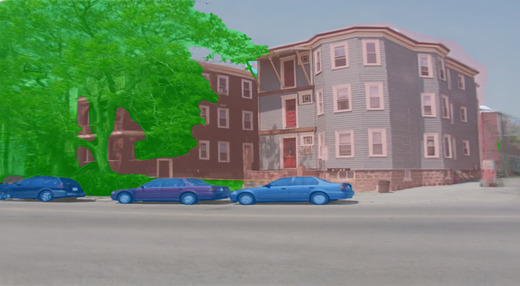}\\[-2.25pt]

\rotatebox[origin=lB]{90}{\hspace{-0.25em}{\shortstack{Lyra~+\\[-1pt]Chorus}}}
& \includegraphics[width=\linewidth]{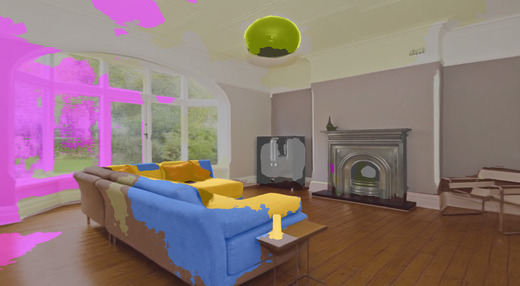}
& \includegraphics[width=\linewidth]{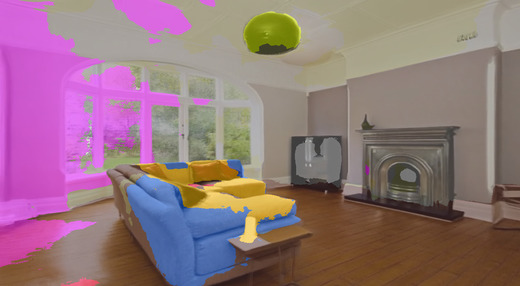}
& \includegraphics[width=\linewidth]{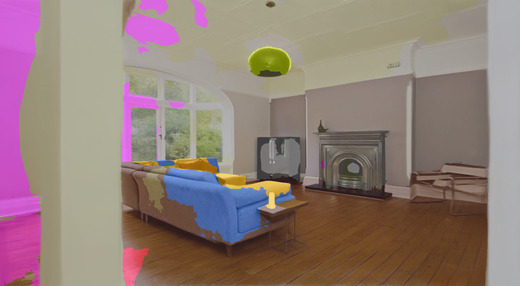}
& \includegraphics[width=\linewidth]{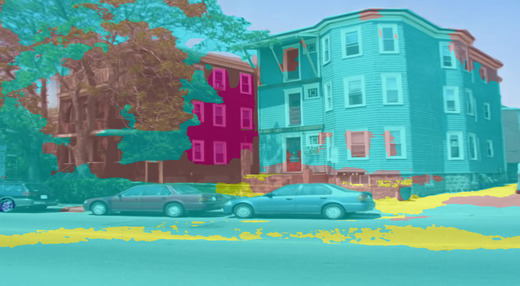}
& \includegraphics[width=\linewidth]{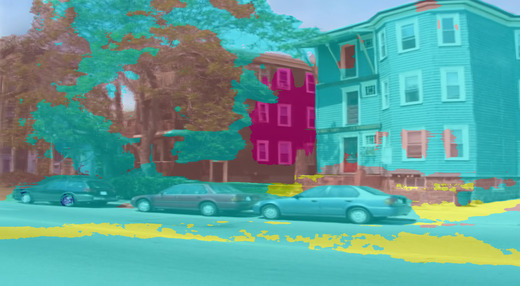}
& \includegraphics[width=\linewidth]{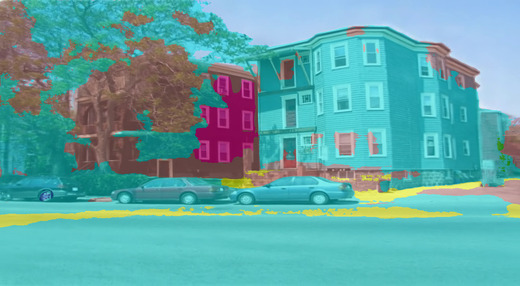}\\[-2.25pt]

\rotatebox[origin=lB]{90}{\hspace{-0.25em}{\shortstack{Lyra~+\\[-1pt]IGGT}}}
& \includegraphics[width=\linewidth]{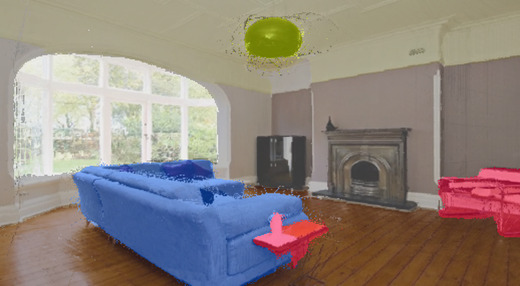}
& \includegraphics[width=\linewidth]{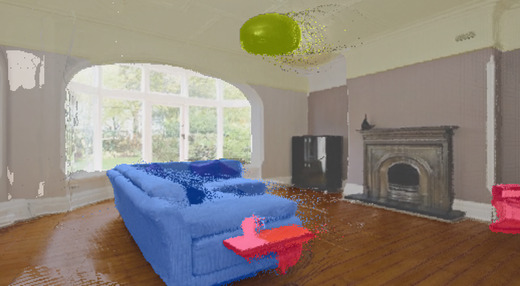}
& \includegraphics[width=\linewidth]{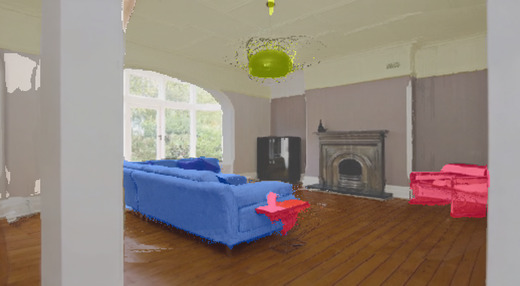}
& \includegraphics[width=\linewidth]{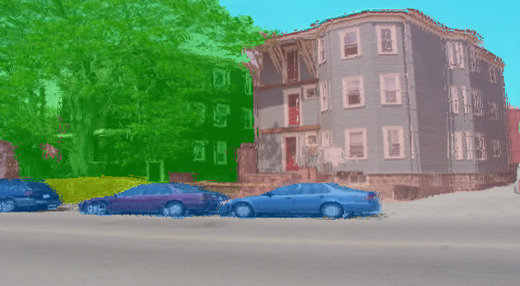}
& \includegraphics[width=\linewidth]{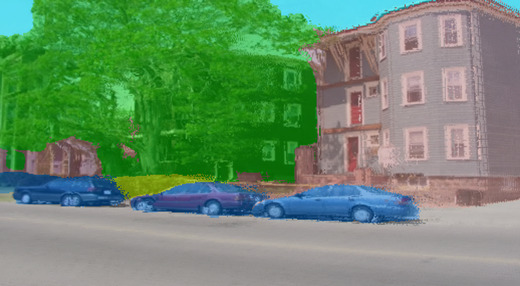}
& \includegraphics[width=\linewidth]{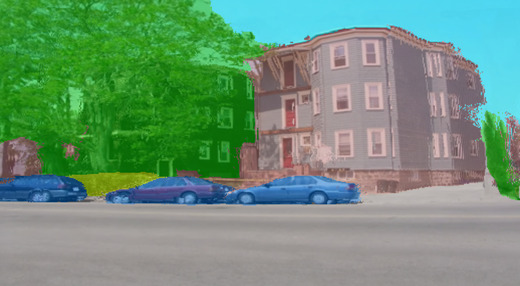}\\[-2.25pt]

\rotatebox[origin=lB]{90}{\hspace{-0.25em}{\MethodName}}
& \includegraphics[width=\linewidth]{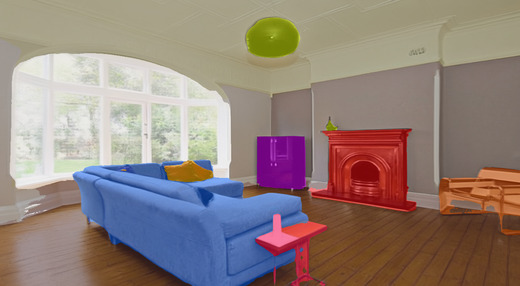}
& \includegraphics[width=\linewidth]{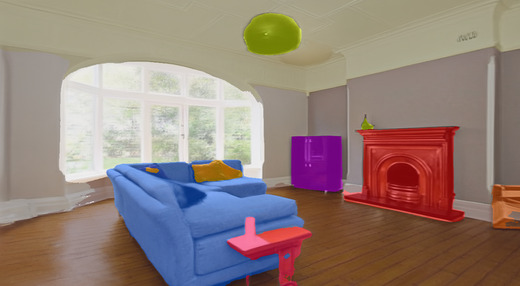}
& \includegraphics[width=\linewidth]{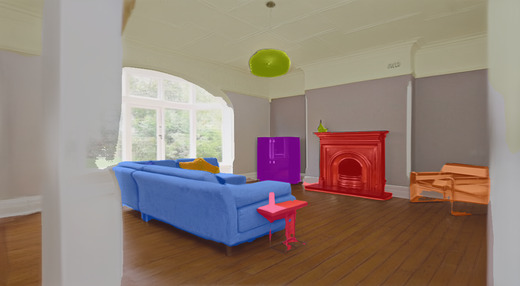}
& \includegraphics[width=\linewidth]{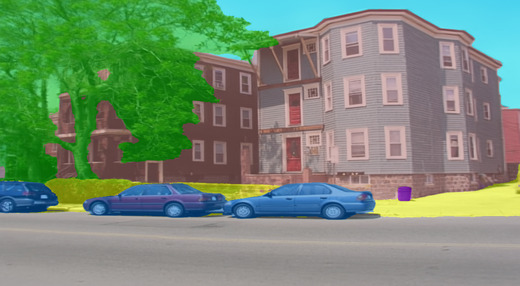}
& \includegraphics[width=\linewidth]{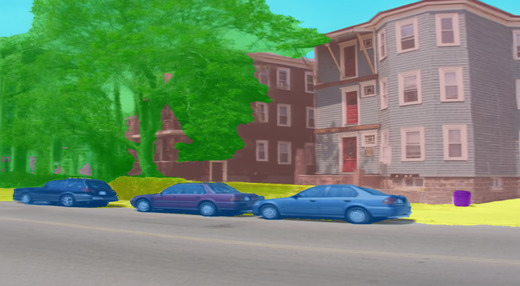}
& \includegraphics[width=\linewidth]{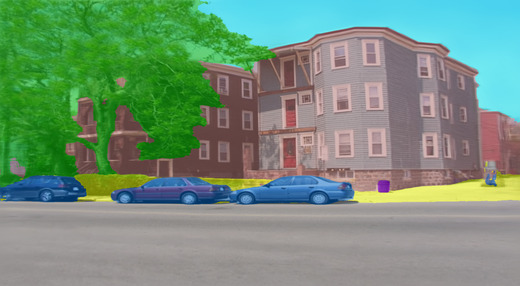}\\[-2.25pt]

\rotatebox[origin=lB]{90}{\hspace{-0.25em}{\MethodNamePlus}}
& \includegraphics[width=\linewidth]{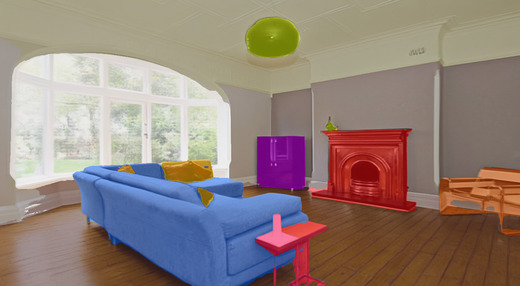}
& \includegraphics[width=\linewidth]{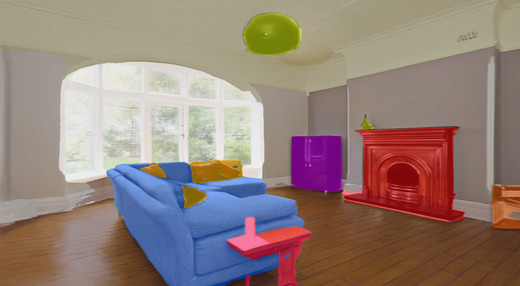}
& \includegraphics[width=\linewidth]{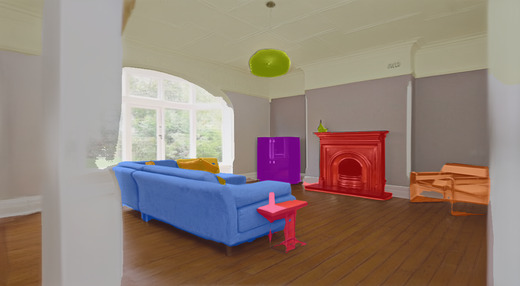}
& \includegraphics[width=\linewidth]{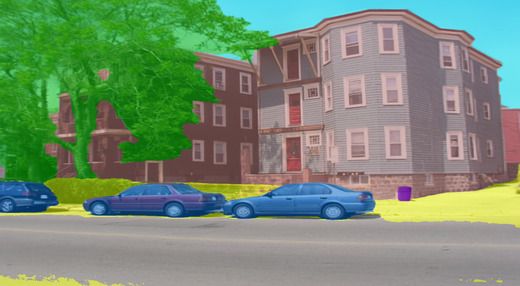}
& \includegraphics[width=\linewidth]{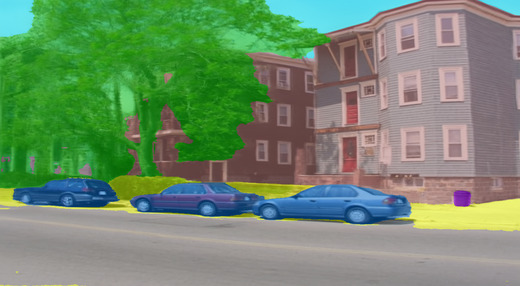}
& \includegraphics[width=\linewidth]{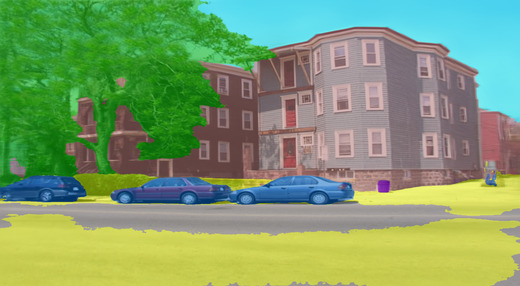}\\[1pt]

& \multicolumn{3}{@{}c@{\hspace{4pt}}}{%
    \begin{minipage}[t]{0.399\textwidth}%
    \vspace{0pt}%
    \begingroup
    \scriptsize
    \sffamily
    \setlength{\tabcolsep}{0pt}%
    \renewcommand{\arraystretch}{0.9}%
    \begin{tabularx}{\linewidth}{*{6}{>{\centering\arraybackslash}X}}
        \cellcolor{wall}\textcolor{white}{Wall} &
        \cellcolor{floor}\textcolor{white}{Floor} &
        \cellcolor{ceiling}\textcolor{white}{Ceiling} &
        \cellcolor{windowpane}\textcolor{black}{Window} &
        \cellcolor{cabinet}\textcolor{white}{Cabinet} &
        \cellcolor{table}\textcolor{white}{Table} \\
        \cellcolor{chair}\textcolor{white}{Chair} &
        \cellcolor{sofa}\textcolor{white}{Sofa} &
        \cellcolor{lamp}\textcolor{black}{Lamp} &
        \cellcolor{cushion}\textcolor{black}{Cushion} &
        \cellcolor{fireplace}\textcolor{white}{Fireplace} &
        \cellcolor{bottle}\textcolor{black}{Bottle} \\
    \end{tabularx}%
    \endgroup
    \end{minipage}%
}
& \multicolumn{3}{@{}c@{}}{%
    \begin{minipage}[t]{0.399\textwidth}%
    \vspace{0pt}%
    \begingroup
    \scriptsize
    \sffamily
    \setlength{\tabcolsep}{0pt}%
    \renewcommand{\arraystretch}{0.9}%
    \begin{tabularx}{\linewidth}{*{3}{>{\centering\arraybackslash}X}}
        \cellcolor{building}\textcolor{white}{Building} &
        \cellcolor{sky}\textcolor{black}{Sky} &
        \cellcolor{tree}\textcolor{white}{Tree} \\
        \cellcolor{road}\textcolor{white}{Road} &
        \cellcolor{sidewalk}\textcolor{black}{Sidewalk} &
        \cellcolor{plant}\textcolor{black}{Plant} \\
        \cellcolor{car}\textcolor{white}{Car} &
        \cellcolor{signboard}\textcolor{white}{Signboard} &
        \cellcolor{ashcan}\textcolor{white}{Trash can} \\
    \end{tabularx}%
    \endgroup
    \end{minipage}%
}\\

&\multicolumn{6}{c}{\rule[0.5ex]{4.5cm}{0.4pt}\textbf{~Instance Segmentation~}\rule[0.5ex]{4.5cm}{0.4pt}}\\

\rotatebox[origin=lB]{90}{\hspace{-0.25em}{RGB}}
& \includegraphics[width=\linewidth]{figures/supplement/ade20k_eval_qualitative/ADE_val_00001526/traj05_t120_rgb.jpg}
& \includegraphics[width=\linewidth]{figures/supplement/ade20k_eval_qualitative/ADE_val_00001526/traj01_t120_rgb.jpg}
& \includegraphics[width=\linewidth]{figures/supplement/ade20k_eval_qualitative/ADE_val_00001526/traj03_t120_rgb.jpg}
& \includegraphics[width=\linewidth]{figures/supplement/ade20k_eval_qualitative/ADE_val_00001805/traj05_t120_rgb.jpg}
& \includegraphics[width=\linewidth]{figures/supplement/ade20k_eval_qualitative/ADE_val_00001805/traj01_t120_rgb.jpg}
& \includegraphics[width=\linewidth]{figures/supplement/ade20k_eval_qualitative/ADE_val_00001805/traj03_t120_rgb.jpg}\\[-2.25pt]

\rotatebox[origin=lB]{90}{\hspace{-0.25em}{\shortstack{Ground\\[-1pt]Truth}}}
& \includegraphics[width=\linewidth]{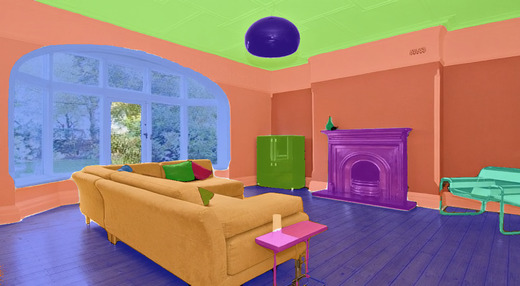}
& \includegraphics[width=\linewidth]{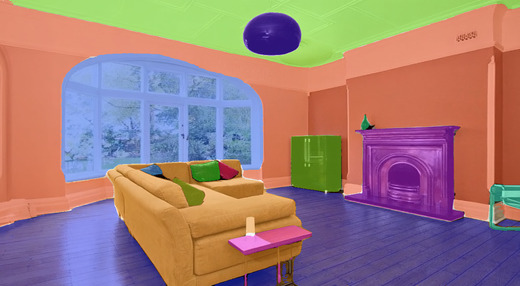}
& \includegraphics[width=\linewidth]{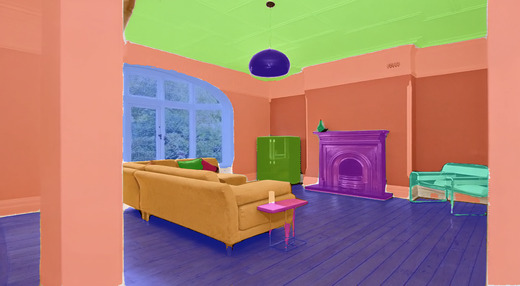}
& \includegraphics[width=\linewidth]{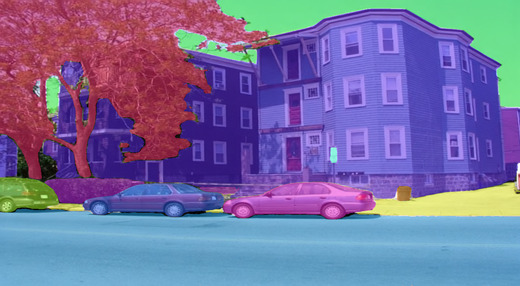}
& \includegraphics[width=\linewidth]{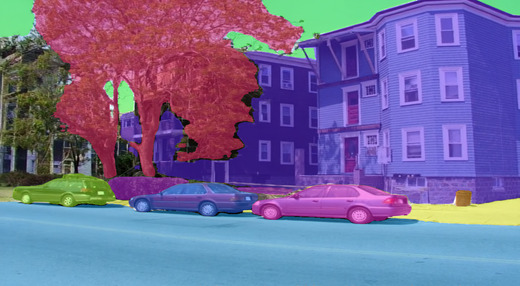}
& \includegraphics[width=\linewidth]{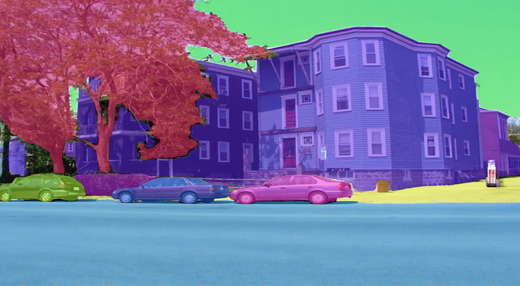}\\[-2.25pt]

\rotatebox[origin=lB]{90}{\hspace{-0.25em}{\shortstack{Lyra~+\\[-1pt]Lifted\\[-1pt]Teachers}}}
& \includegraphics[width=\linewidth]{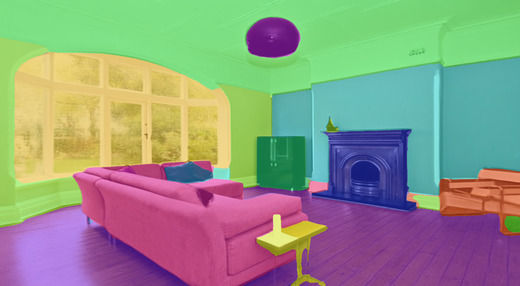}
& \includegraphics[width=\linewidth]{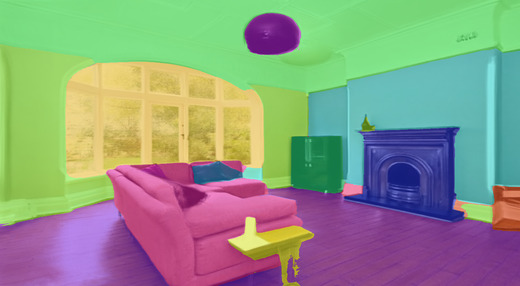}
& \includegraphics[width=\linewidth]{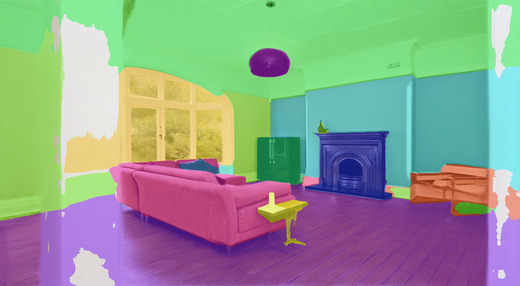}
& \includegraphics[width=\linewidth]{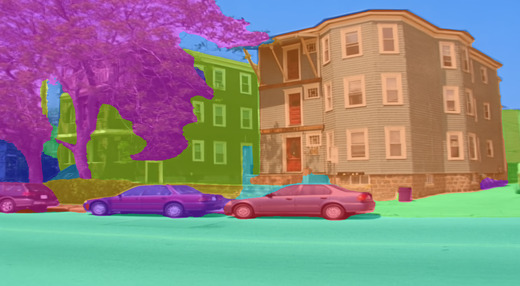}
& \includegraphics[width=\linewidth]{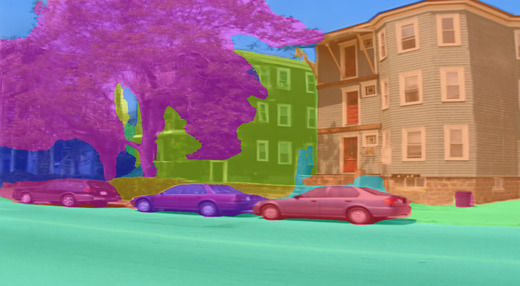}
& \includegraphics[width=\linewidth]{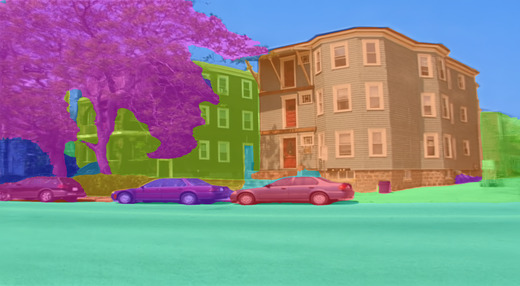}\\[-2.25pt]

\rotatebox[origin=lB]{90}{\hspace{-0.25em}{\shortstack{Lyra~+\\[-1pt]Unified-Lift}}}
& \includegraphics[width=\linewidth]{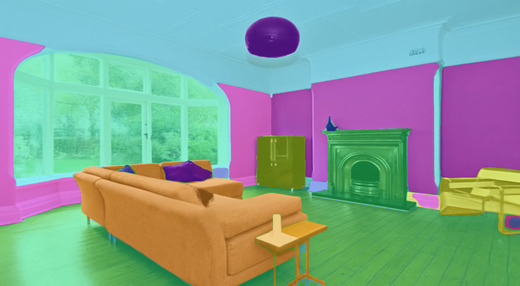}
& \includegraphics[width=\linewidth]{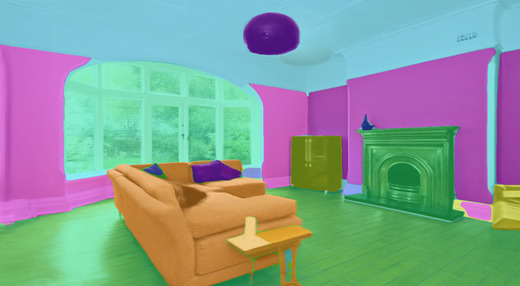}
& \includegraphics[width=\linewidth]{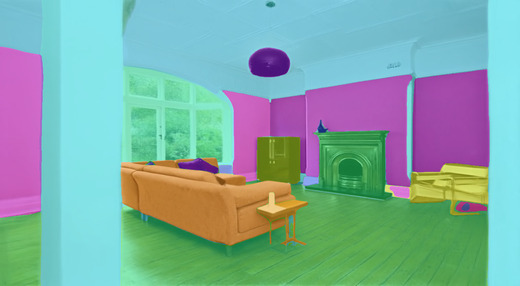}
& \includegraphics[width=\linewidth]{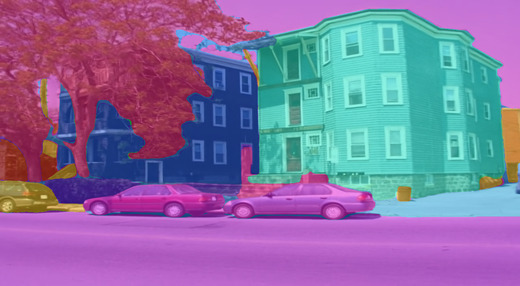}
& \includegraphics[width=\linewidth]{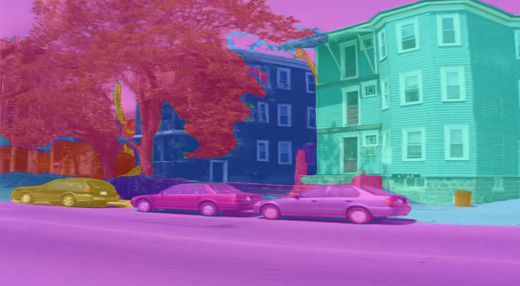}
& \includegraphics[width=\linewidth]{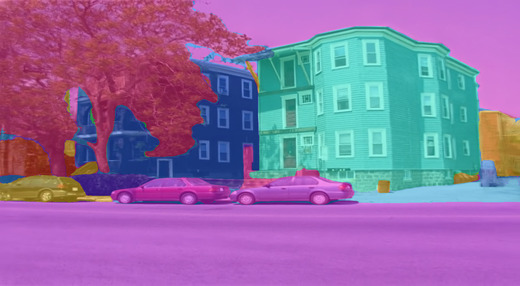}\\[-2.25pt]

\rotatebox[origin=lB]{90}{\hspace{-0.25em}{\shortstack{Lyra\textsuperscript{$\dagger$}~+\\[-1pt]IGGT}}}
& \includegraphics[width=\linewidth]{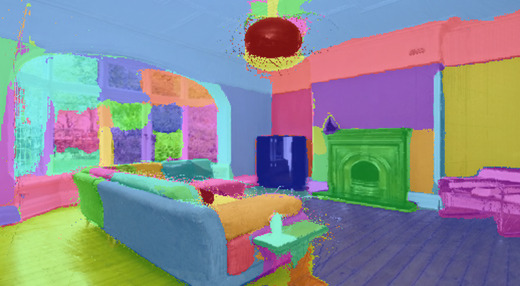}
& \includegraphics[width=\linewidth]{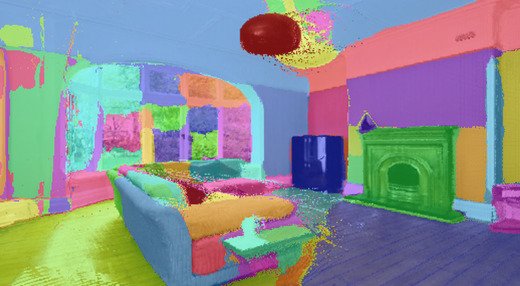}
& \includegraphics[width=\linewidth]{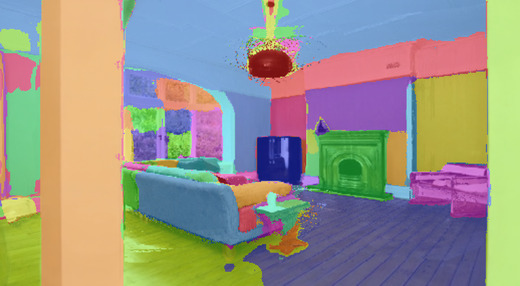}
& \includegraphics[width=\linewidth]{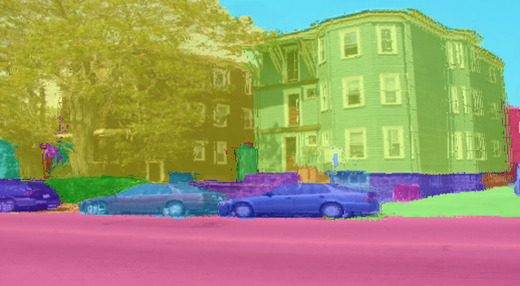}
& \includegraphics[width=\linewidth]{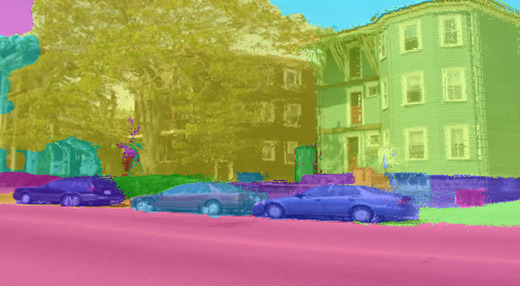}
& \includegraphics[width=\linewidth]{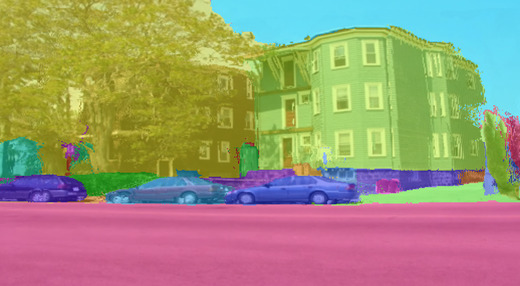}\\[-2.25pt]

\rotatebox[origin=lB]{90}{\hspace{-0.25em}{\MethodName}}
& \includegraphics[width=\linewidth]{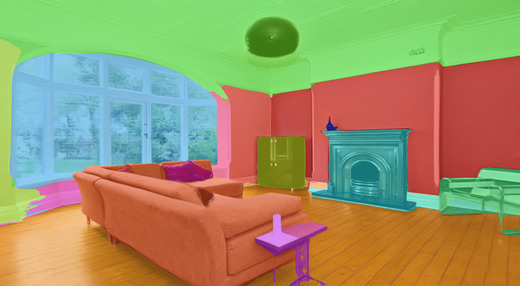}
& \includegraphics[width=\linewidth]{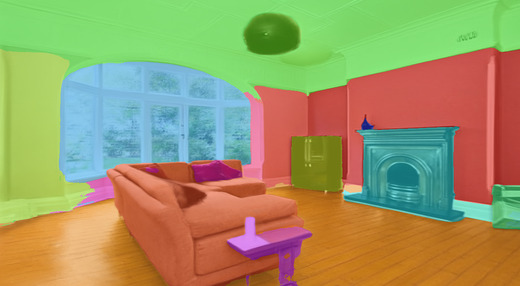}
& \includegraphics[width=\linewidth]{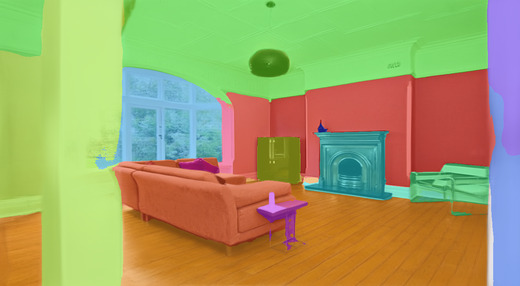}
& \includegraphics[width=\linewidth]{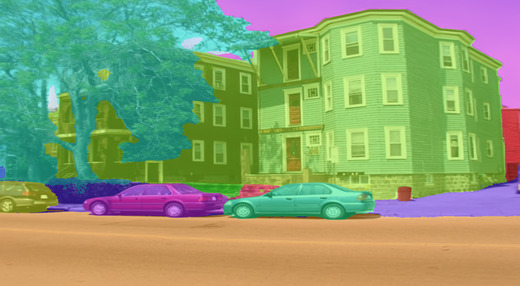}
& \includegraphics[width=\linewidth]{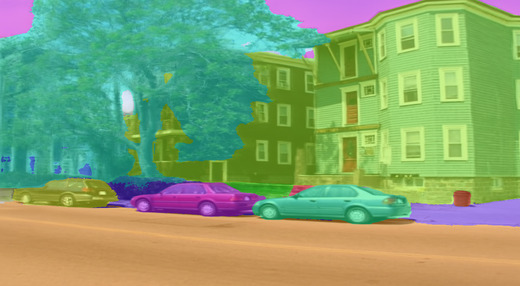}
& \includegraphics[width=\linewidth]{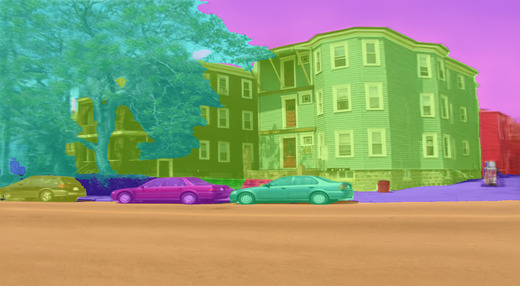}\\[-2.25pt]

\rotatebox[origin=lB]{90}{\hspace{-0.25em}{\MethodNamePlus}}
& \includegraphics[width=\linewidth]{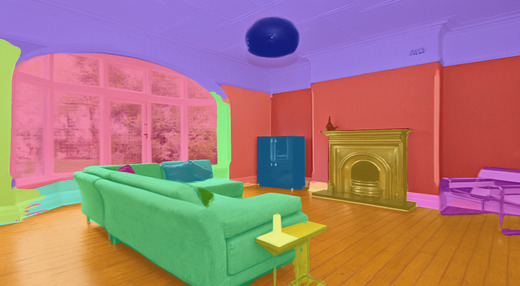}
& \includegraphics[width=\linewidth]{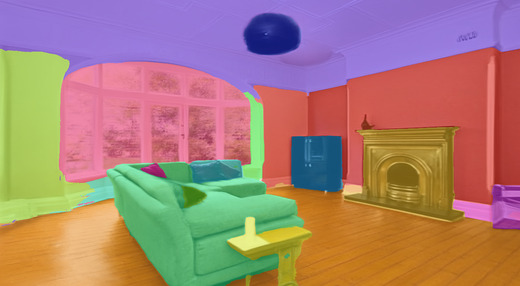}
& \includegraphics[width=\linewidth]{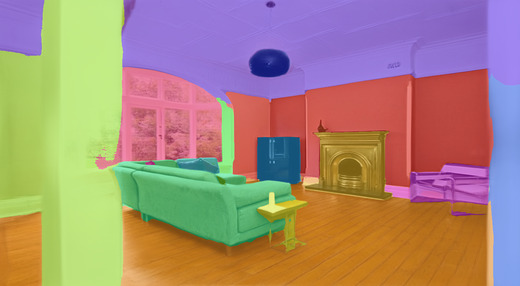}
& \includegraphics[width=\linewidth]{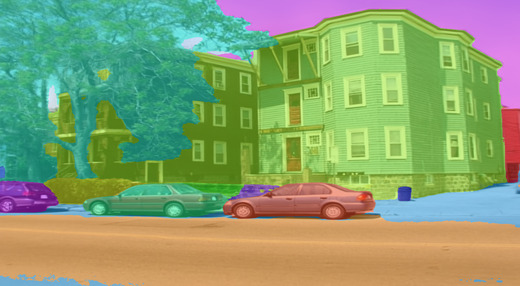}
& \includegraphics[width=\linewidth]{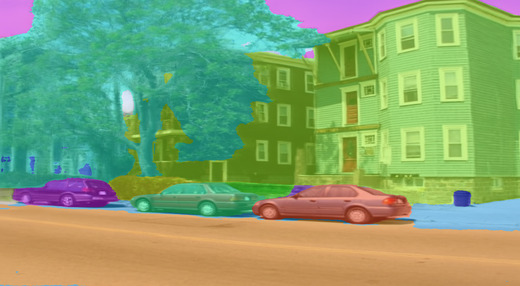}
& \includegraphics[width=\linewidth]{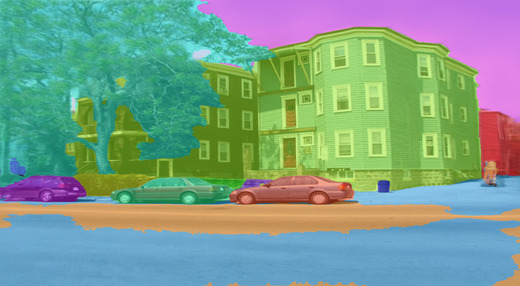}\\[-2.25pt]

\end{tabular}

    \caption{\textbf{Qualitative comparison on \MethodName-ADE20K static eval.}
    We show three viewpoints per scene, with semantic predictions above and instance
    predictions below. $\dagger$ denotes
    directly using generated multi-view RGB.}
    \label{fig:supp-qualitative-ade20k}
\end{figure}

\Cref{fig:supp-qualitative-sdg,fig:supp-qualitative-ade20k} compare semantic and instance predictions across both static benchmarks.

\inlinesection{Open-vocabulary semantics.}
LangSplatV2~\citep{li:2025:hd3l} and SFS~\citep{xiong:2026:sfs} recover broad semantic structure but exhibit class confusion and label leakage across object boundaries, with LangSplatV2 additionally showing flickering artifacts as the viewpoint changes. The feed-forward baselines Chorus~\citep{li:2026:mtph} and IGGT~\citep{li:2025:iggt} exhibit incomplete category coverage and noisy boundaries, respectively. We hypothesize that their reduced performance partly reflects a domain shift from real-world and synthetic 3D training data to our generatively reconstructed scenes. \MethodName preserves coherent semantic regions across viewpoints, approaching the lifted-teacher reference while correcting some of its assignment errors. \MethodNamePlus further refines semantic assignments without changing geometry or appearance.

\inlinesection{Instance segmentation.}
Unified-Lift~\citep{zhu:2025:rete} produces coherent instance regions but exhibits local label fragmentation and bleeding across object boundaries. IGGT shows blurred instance boundaries and a tendency to over-segment individual objects. Its reconstruction appears sensitive to cross-view inconsistencies in the generated videos, producing duplicated fine-scale structures that further blur segmentation predictions. Our methods preserve more coherent object-level partitions across viewpoints, while the lifted-teacher reference remains competitive. Since instance clustering is conditioned on predicted semantic classes, \MethodNamePlus can modify instance partitions without changing the learned instance embeddings.

\begin{figure}[p]
    \centering
    \begingroup
\newcommand{\dynamicBaselineTimedImage}[2]{%
    \begin{tikzpicture}[baseline=0pt]
        \node[anchor=south west,inner sep=0,outer sep=0] (timedimage) at (0,0) {#1};
        \begin{scope}[overlay,
            every node/.style={anchor=south west,inner sep=0,outer sep=0,
                font=\normalfont\scriptsize}]
            \coordinate (timelabel) at ([xshift=1.2pt,yshift=1.2pt]timedimage.south west);
            \foreach \contourangle in {0,22.5,...,337.5}{
                \node[text=white,
                    xshift={0.06mm*cos(\contourangle)},yshift={0.06mm*sin(\contourangle)}]
                    at (timelabel) {$t{=}\num{#2}$};
            }
            \node[text=black] at (timelabel) {$t{=}\num{#2}$};
        \end{scope}
    \end{tikzpicture}%
}
\newlength{\dynamicBaselinePanelHeight}
\setlength{\dynamicBaselinePanelHeight}{0.060500\linewidth}
\newcommand{\dynamicBaselineMethodLabel}[1]{%
    \raisebox{\dimexpr.5\dynamicBaselinePanelHeight-.5\height+.5\depth\relax}%
        [\dynamicBaselinePanelHeight][0pt]{\rotatebox[origin=c]{90}{\shortstack{#1}}}%
}
\scriptsize\sffamily
\newlength{\dynamicBaselineHeadingWidth}
\newlength{\dynamicBaselineHeadingRuleWidth}
\settowidth{\dynamicBaselineHeadingWidth}{\textbf{~Open-Vocabulary Semantic Segmentation~}}
\addtolength{\dynamicBaselineHeadingWidth}{8cm}
\newcommand{\dynamicBaselineHeading}[1]{%
    \settowidth{\dynamicBaselineHeadingRuleWidth}{\textbf{~#1~}}%
    \setlength{\dynamicBaselineHeadingRuleWidth}{\dimexpr(\dynamicBaselineHeadingWidth-\dynamicBaselineHeadingRuleWidth)/2\relax}%
    \rule[0.5ex]{\dynamicBaselineHeadingRuleWidth}{0.4pt}%
    \textbf{~#1~}%
    \rule[0.5ex]{\dynamicBaselineHeadingRuleWidth}{0.4pt}%
}

\setlength{\tabcolsep}{0pt}
\renewcommand{\arraystretch}{0.94}
\begin{tabular}{>{\centering\arraybackslash}m{0.038\linewidth}@{\hspace{0.5pt}}>{\centering\arraybackslash}m{0.110000\linewidth}@{\hspace{0.5pt}}>{\centering\arraybackslash}m{0.110000\linewidth}@{\hspace{0.5pt}}>{\centering\arraybackslash}m{0.110000\linewidth}@{\hspace{0.5pt}}>{\centering\arraybackslash}m{0.110000\linewidth}@{\hspace{4pt}}>{\centering\arraybackslash}m{0.110000\linewidth}@{\hspace{0.5pt}}>{\centering\arraybackslash}m{0.110000\linewidth}@{\hspace{0.5pt}}>{\centering\arraybackslash}m{0.110000\linewidth}@{\hspace{0.5pt}}>{\centering\arraybackslash}m{0.110000\linewidth}}
&\multicolumn{8}{c}{\dynamicBaselineHeading{RGB Reconstruction}}\\
\dynamicBaselineMethodLabel{Reference\\[-1.5pt]RGB} &
\dynamicBaselineTimedImage{\includegraphics[width=\linewidth]{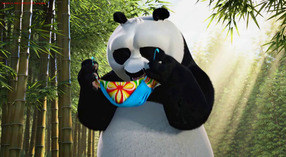}}{0} & \dynamicBaselineTimedImage{\includegraphics[width=\linewidth]{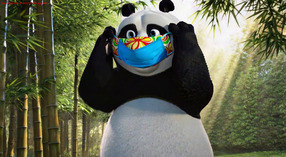}}{40} & \dynamicBaselineTimedImage{\includegraphics[width=\linewidth]{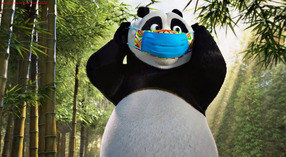}}{80} & \dynamicBaselineTimedImage{\includegraphics[width=\linewidth]{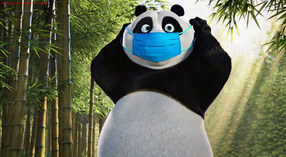}}{120} & \dynamicBaselineTimedImage{\includegraphics[width=\linewidth]{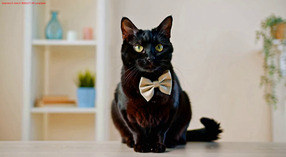}}{0} & \dynamicBaselineTimedImage{\includegraphics[width=\linewidth]{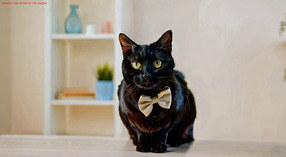}}{40} & \dynamicBaselineTimedImage{\includegraphics[width=\linewidth]{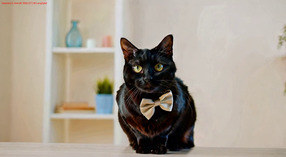}}{80} & \dynamicBaselineTimedImage{\includegraphics[width=\linewidth]{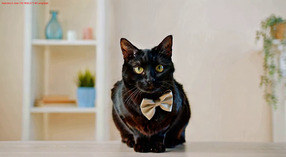}}{120} \\[-2.1pt]
\dynamicBaselineMethodLabel{Lyra} & \includegraphics[width=\linewidth]{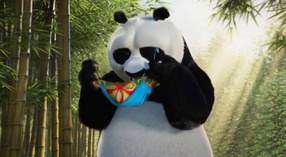} & \includegraphics[width=\linewidth]{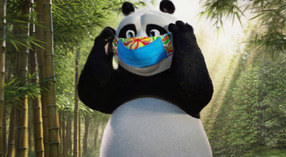} & \includegraphics[width=\linewidth]{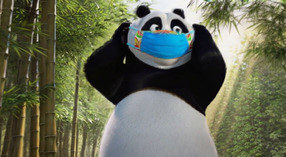} & \includegraphics[width=\linewidth]{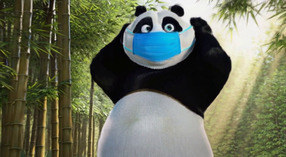} & \includegraphics[width=\linewidth]{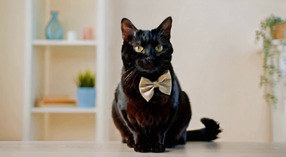} & \includegraphics[width=\linewidth]{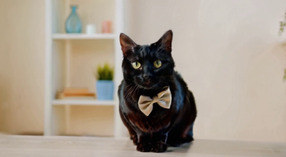} & \includegraphics[width=\linewidth]{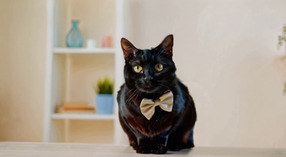} & \includegraphics[width=\linewidth]{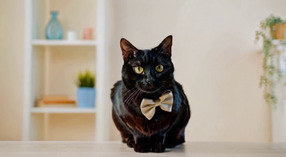} \\[-2.1pt]
\dynamicBaselineMethodLabel{4D\\[-1.5pt]LangSplat} & \includegraphics[width=\linewidth]{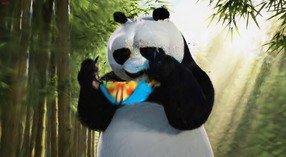} & \includegraphics[width=\linewidth]{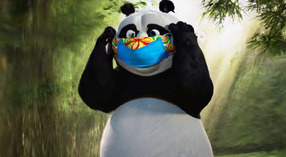} & \includegraphics[width=\linewidth]{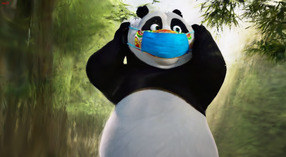} & \includegraphics[width=\linewidth]{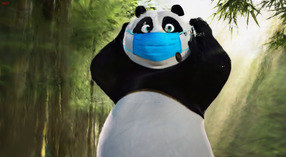} & \includegraphics[width=\linewidth]{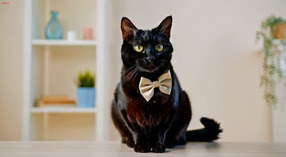} & \includegraphics[width=\linewidth]{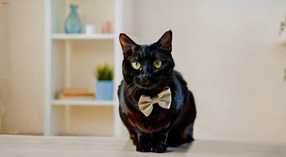} & \includegraphics[width=\linewidth]{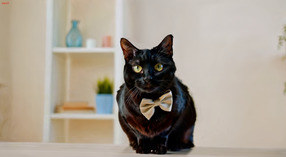} & \includegraphics[width=\linewidth]{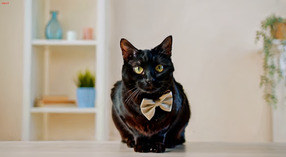} \\[-2.1pt]
\dynamicBaselineMethodLabel{TRASE} & \includegraphics[width=\linewidth]{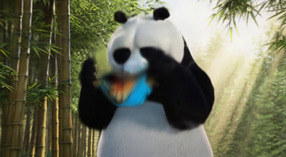} & \includegraphics[width=\linewidth]{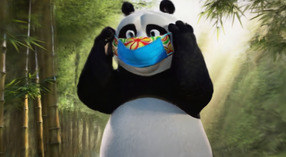} & \includegraphics[width=\linewidth]{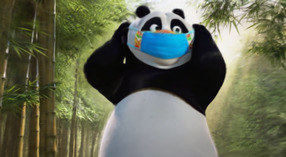} & \includegraphics[width=\linewidth]{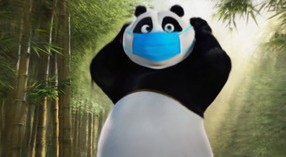} & \includegraphics[width=\linewidth]{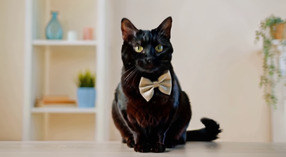} & \includegraphics[width=\linewidth]{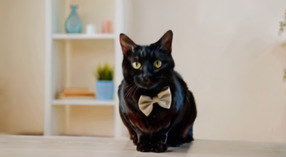} & \includegraphics[width=\linewidth]{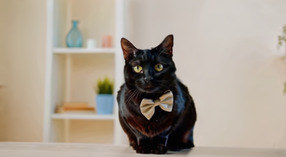} & \includegraphics[width=\linewidth]{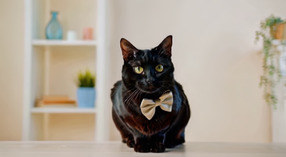} \\[-2.1pt]
\dynamicBaselineMethodLabel{Lucid\\[-1.5pt]Lyra} & \includegraphics[width=\linewidth]{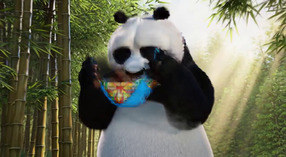} & \includegraphics[width=\linewidth]{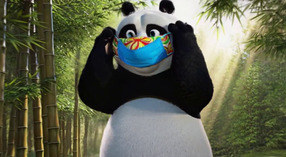} & \includegraphics[width=\linewidth]{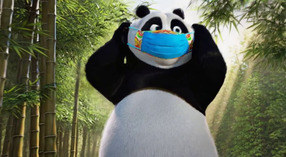} & \includegraphics[width=\linewidth]{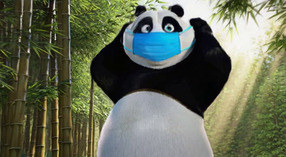} & \includegraphics[width=\linewidth]{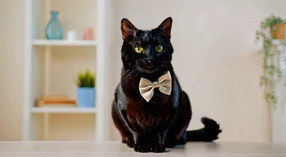} & \includegraphics[width=\linewidth]{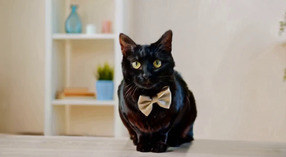} & \includegraphics[width=\linewidth]{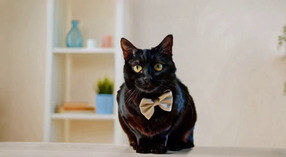} & \includegraphics[width=\linewidth]{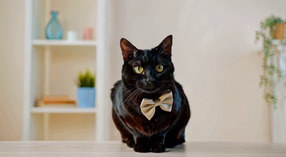} \\[-2.1pt]
\dynamicBaselineMethodLabel{Lucid\\[-1.5pt]Lyra+} & \includegraphics[width=\linewidth]{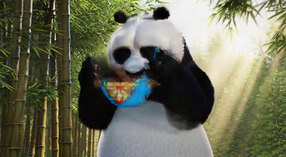} & \includegraphics[width=\linewidth]{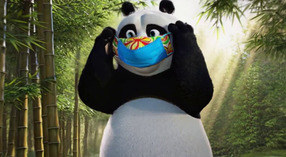} & \includegraphics[width=\linewidth]{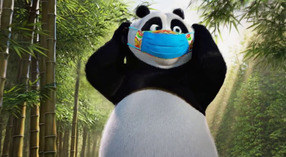} & \includegraphics[width=\linewidth]{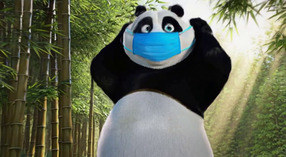} & \includegraphics[width=\linewidth]{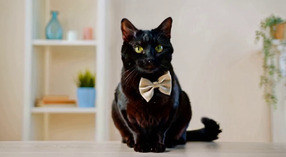} & \includegraphics[width=\linewidth]{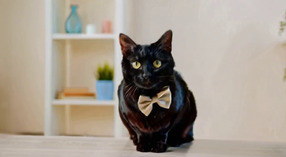} & \includegraphics[width=\linewidth]{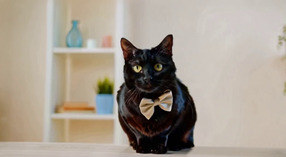} & \includegraphics[width=\linewidth]{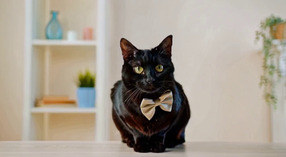} \\[-2.1pt]
\\[-2pt]
&\multicolumn{8}{c}{\dynamicBaselineHeading{Open-Vocabulary Semantic Segmentation}}\\[2pt]
\dynamicBaselineMethodLabel{Ground\\[-1.5pt]truth} & \dynamicBaselineTimedImage{\includegraphics[width=\linewidth]{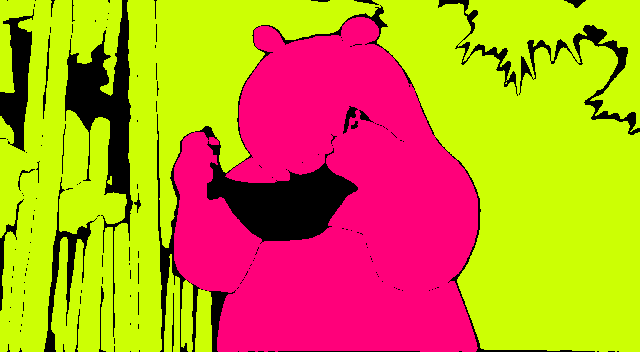}}{0} & \dynamicBaselineTimedImage{\includegraphics[width=\linewidth]{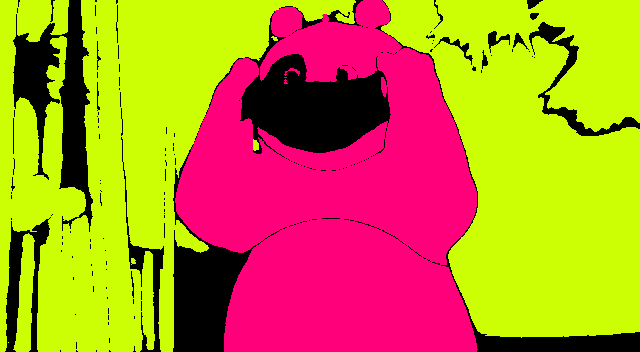}}{40} & \dynamicBaselineTimedImage{\includegraphics[width=\linewidth]{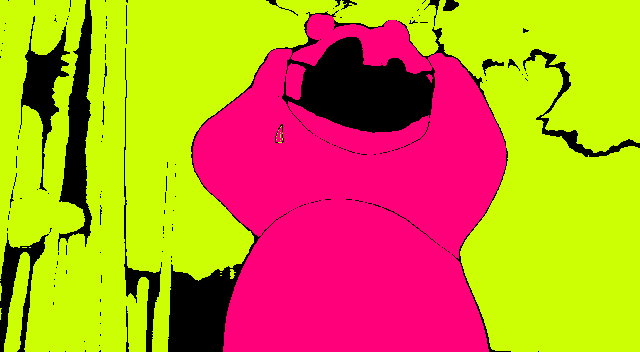}}{80} & \dynamicBaselineTimedImage{\includegraphics[width=\linewidth]{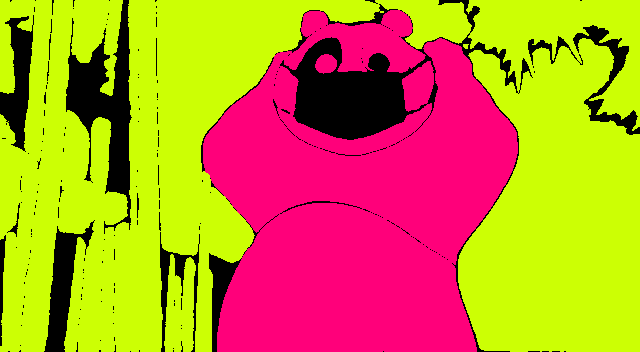}}{120} & \dynamicBaselineTimedImage{\includegraphics[width=\linewidth]{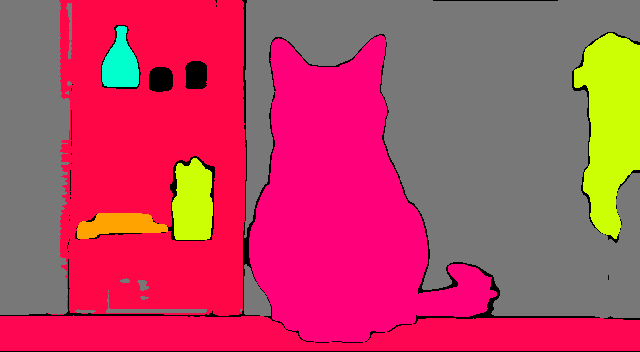}}{0} & \dynamicBaselineTimedImage{\includegraphics[width=\linewidth]{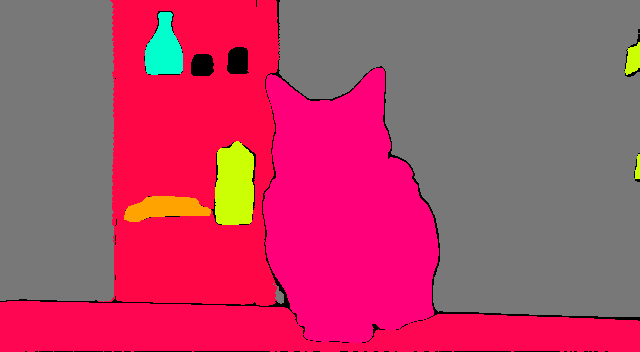}}{40} & \dynamicBaselineTimedImage{\includegraphics[width=\linewidth]{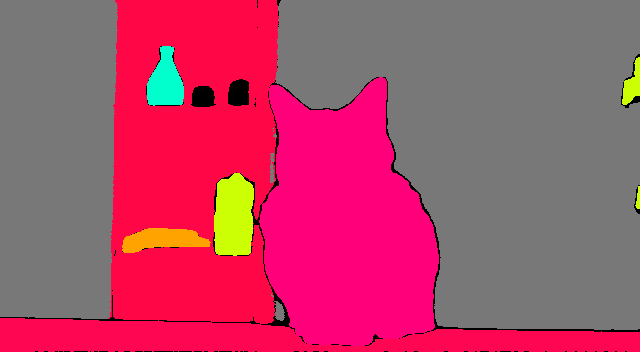}}{80} & \dynamicBaselineTimedImage{\includegraphics[width=\linewidth]{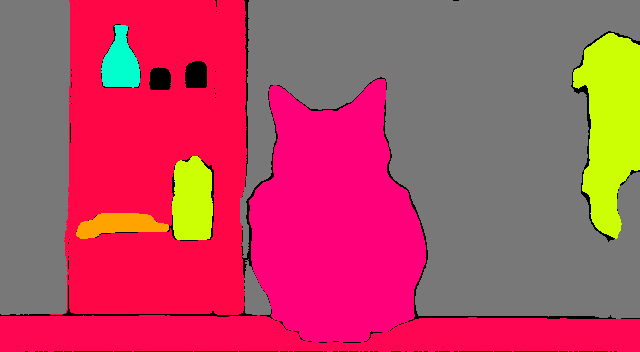}}{120} \\[-2.1pt]
\dynamicBaselineMethodLabel{Lifted\\[-1.5pt]Teachers} & \includegraphics[width=\linewidth]{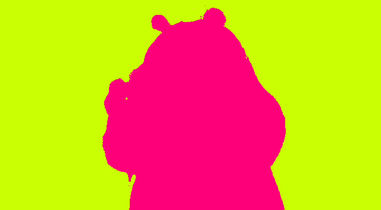} & \includegraphics[width=\linewidth]{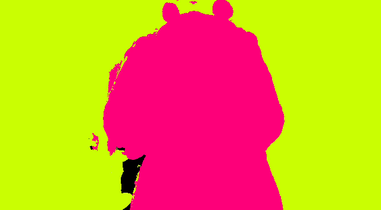} & \includegraphics[width=\linewidth]{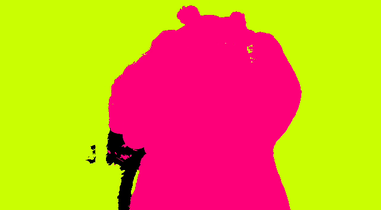} & \includegraphics[width=\linewidth]{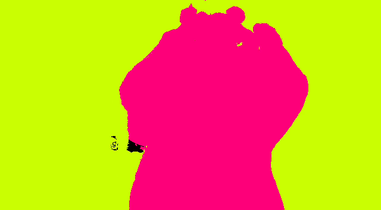} & \includegraphics[width=\linewidth]{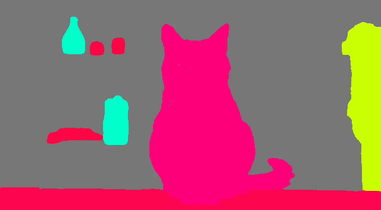} & \includegraphics[width=\linewidth]{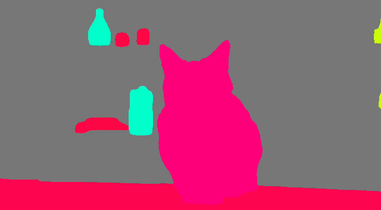} & \includegraphics[width=\linewidth]{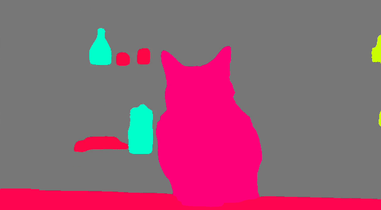} & \includegraphics[width=\linewidth]{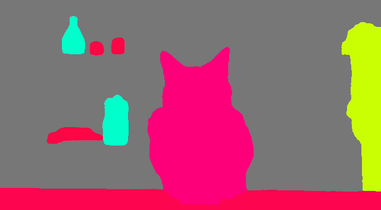} \\[-2.1pt]
\dynamicBaselineMethodLabel{4D~Lang-\\[-1.5pt]Splat} & \includegraphics[width=\linewidth]{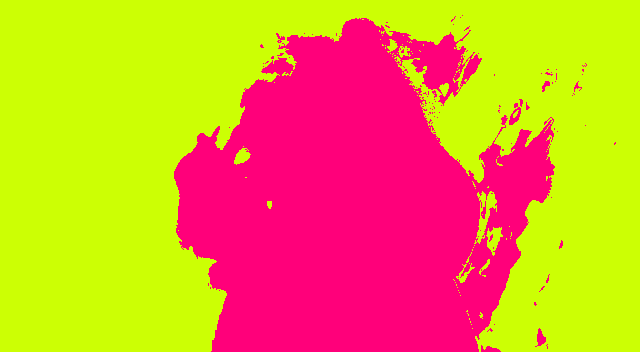} & \includegraphics[width=\linewidth]{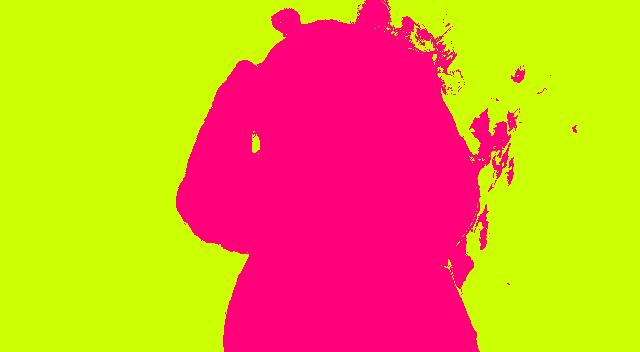} & \includegraphics[width=\linewidth]{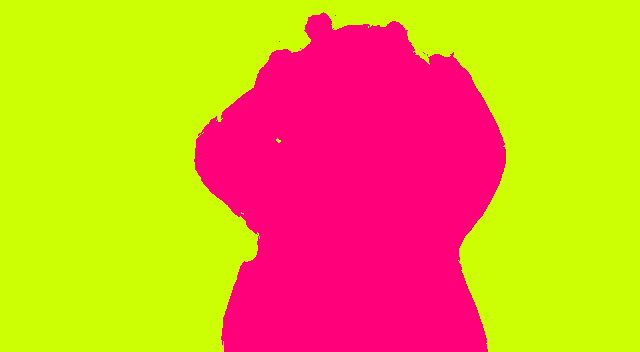} & \includegraphics[width=\linewidth]{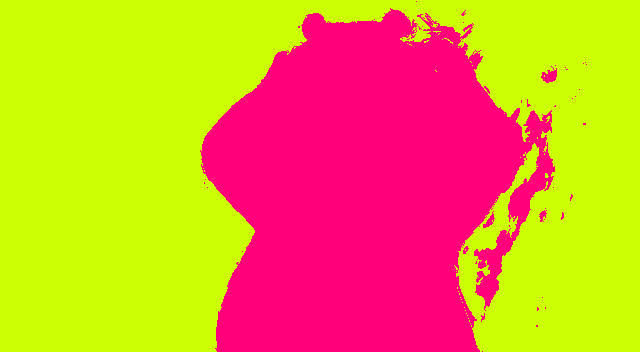} & \includegraphics[width=\linewidth]{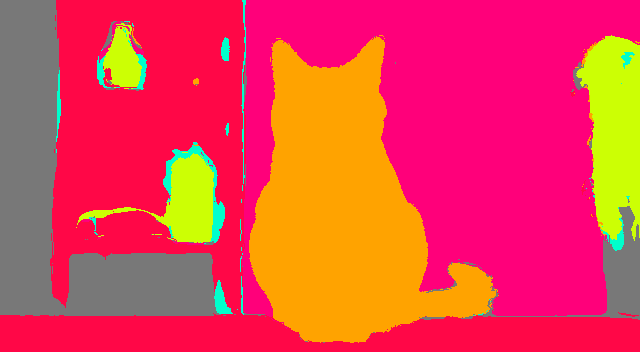} & \includegraphics[width=\linewidth]{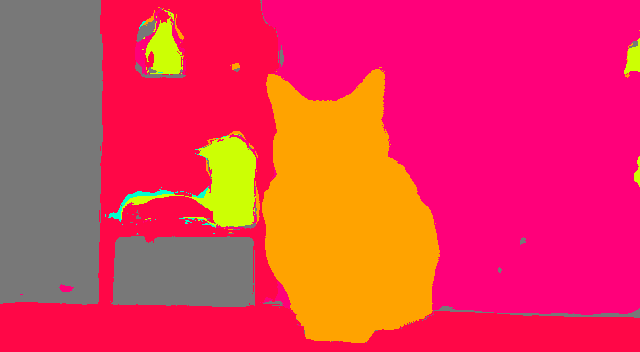} & \includegraphics[width=\linewidth]{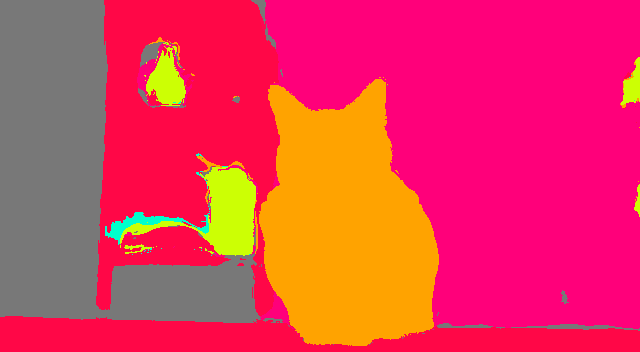} & \includegraphics[width=\linewidth]{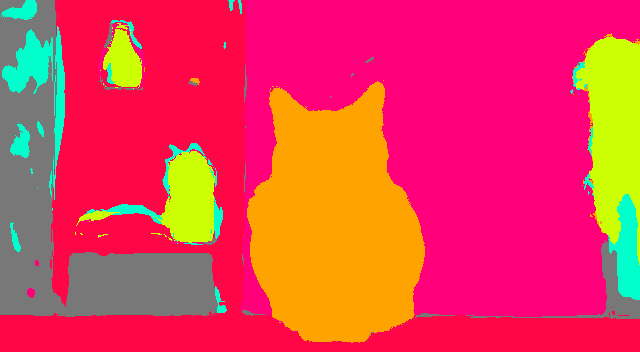} \\[-2.1pt]
\dynamicBaselineMethodLabel{IGGT4D\\[-1.5pt]+LSeg} & \includegraphics[width=\linewidth]{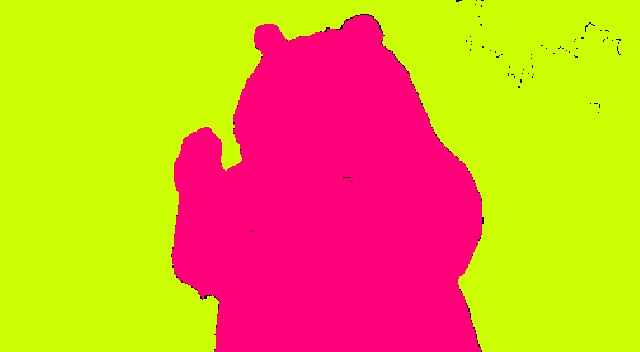} & \includegraphics[width=\linewidth]{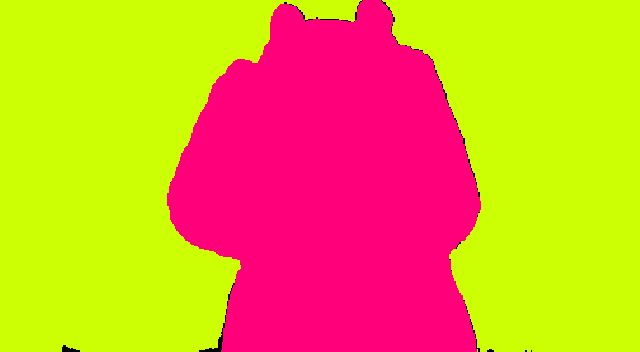} & \includegraphics[width=\linewidth]{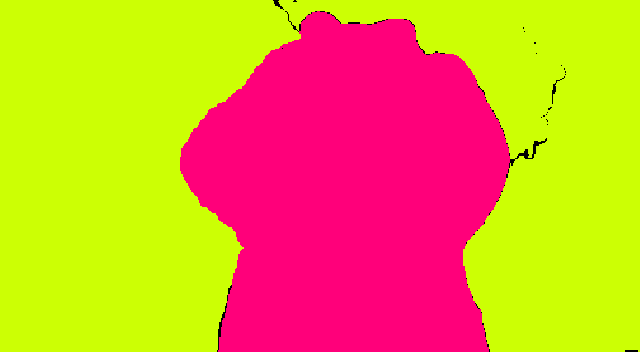} & \includegraphics[width=\linewidth]{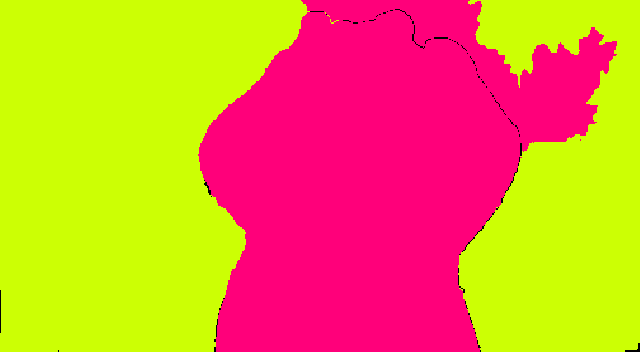} & \includegraphics[width=\linewidth]{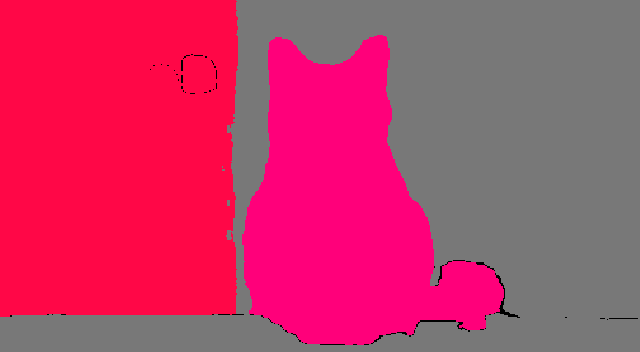} & \includegraphics[width=\linewidth]{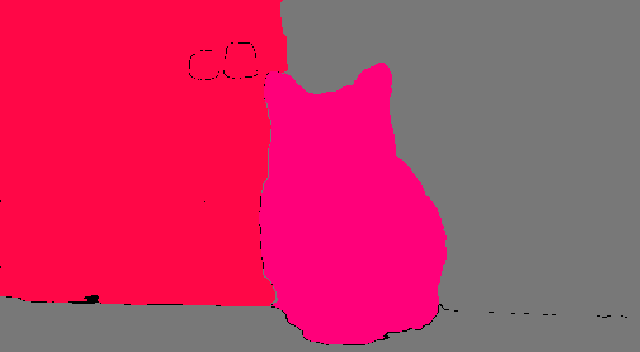} & \includegraphics[width=\linewidth]{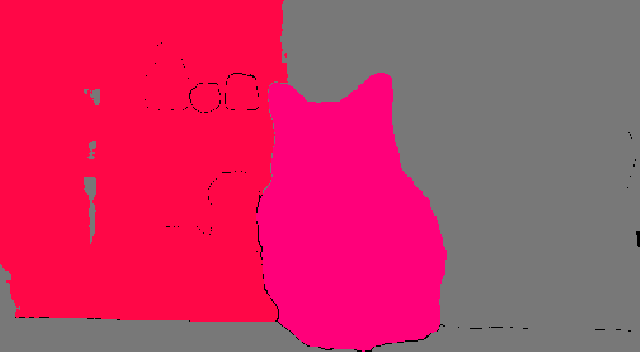} & \includegraphics[width=\linewidth]{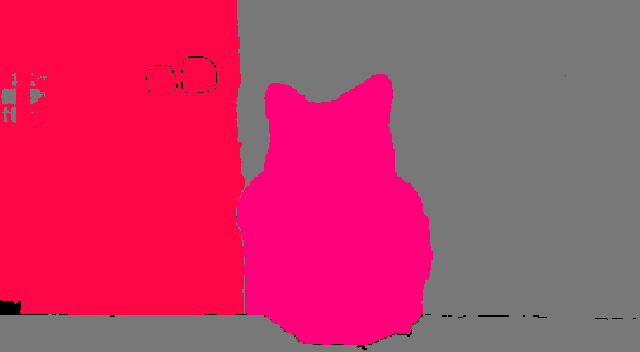} \\[-2.1pt]
\dynamicBaselineMethodLabel{Lucid\\[-1.5pt]Lyra} & \includegraphics[width=\linewidth]{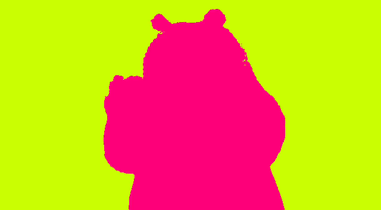} & \includegraphics[width=\linewidth]{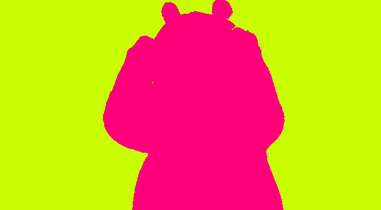} & \includegraphics[width=\linewidth]{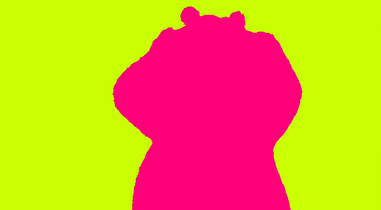} & \includegraphics[width=\linewidth]{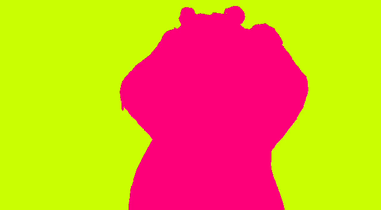} & \includegraphics[width=\linewidth]{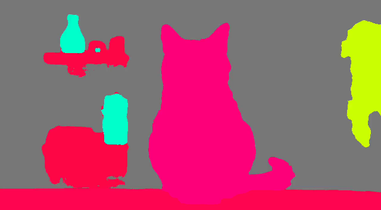} & \includegraphics[width=\linewidth]{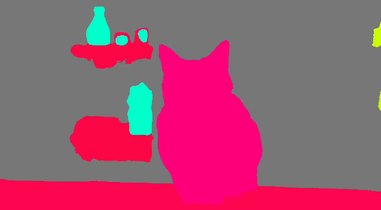} & \includegraphics[width=\linewidth]{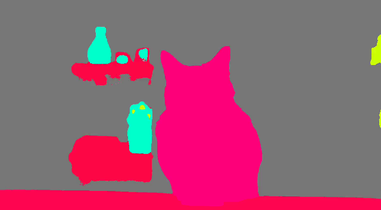} & \includegraphics[width=\linewidth]{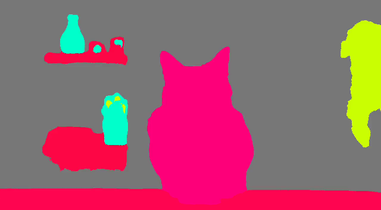} \\[-2.1pt]
\dynamicBaselineMethodLabel{Lucid\\[-1.5pt]Lyra+} & \includegraphics[width=\linewidth]{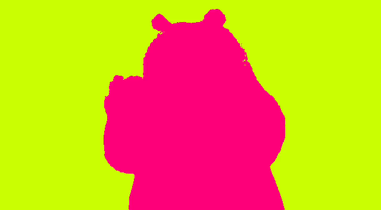} & \includegraphics[width=\linewidth]{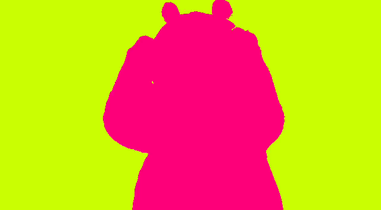} & \includegraphics[width=\linewidth]{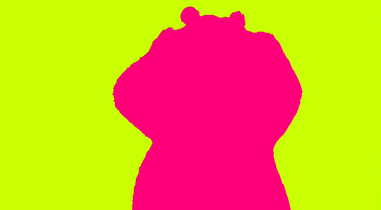} & \includegraphics[width=\linewidth]{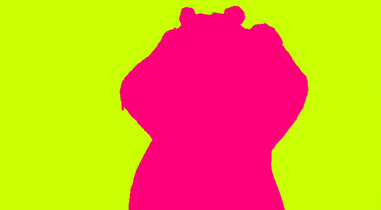} & \includegraphics[width=\linewidth]{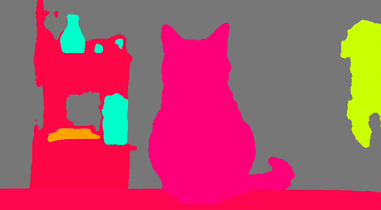} & \includegraphics[width=\linewidth]{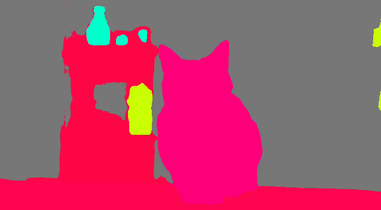} & \includegraphics[width=\linewidth]{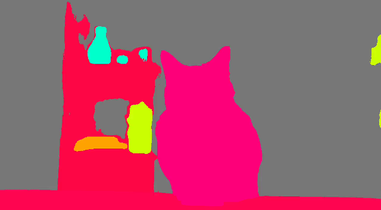} & \includegraphics[width=\linewidth]{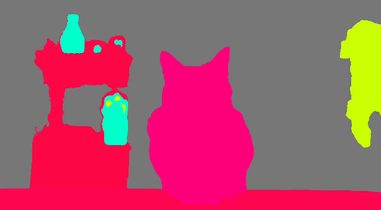} \\[1pt]
& \multicolumn{4}{@{}c@{\hspace{4pt}}}{%
\begin{minipage}[t]{\dimexpr0.440000\textwidth+1.5pt\relax}
\vspace{0pt}\begingroup\scriptsize\sffamily
\setlength{\tabcolsep}{0pt}\renewcommand{\arraystretch}{0.9}
\begin{tabularx}{\linewidth}{*{3}{>{\centering\arraybackslash}X}}
\cellcolor[RGB]{204,255,4}\textcolor{black}{Plant} & \cellcolor[RGB]{255,0,122}\textcolor{white}{Animal} & \cellcolor[RGB]{0,0,0}\textcolor{white}{Unlabeled} \\[-0.25pt]
\end{tabularx}\endgroup\end{minipage}
} & \multicolumn{4}{@{}c@{}}{%
\begin{minipage}[t]{\dimexpr0.440000\textwidth+1.5pt\relax}
\vspace{0pt}\begingroup\scriptsize\sffamily
\setlength{\tabcolsep}{0pt}\renewcommand{\arraystretch}{0.9}
\begin{tabularx}{\linewidth}{*{4}{>{\centering\arraybackslash}X}}
\cellcolor[RGB]{120,120,120}\textcolor{white}{Wall} & \cellcolor[RGB]{255,6,82}\textcolor{white}{Table} & \cellcolor[RGB]{204,255,4}\textcolor{black}{Plant} & \cellcolor[RGB]{255,7,71}\textcolor{white}{Shelf} \\[-0.25pt]
\cellcolor[RGB]{255,163,0}\textcolor{black}{Book} & \cellcolor[RGB]{255,0,122}\textcolor{white}{Animal} & \cellcolor[RGB]{0,255,204}\textcolor{black}{Vase} & \cellcolor[RGB]{0,0,0}\textcolor{white}{Unlabeled} \\[-0.25pt]
\end{tabularx}\endgroup\end{minipage}
} \\[1pt]
\\[-2pt]
&\multicolumn{8}{c}{\dynamicBaselineHeading{Instance Segmentation}}\\[2pt]
\dynamicBaselineMethodLabel{Ground\\[-1.5pt]truth} & \dynamicBaselineTimedImage{\includegraphics[width=\linewidth]{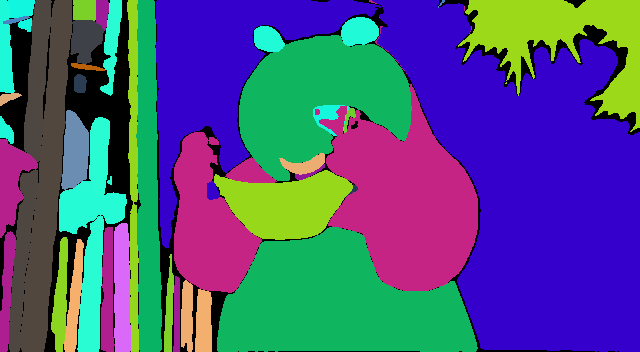}}{0} & \dynamicBaselineTimedImage{\includegraphics[width=\linewidth]{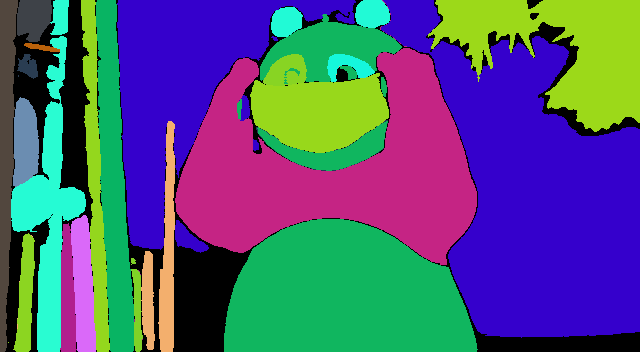}}{40} & \dynamicBaselineTimedImage{\includegraphics[width=\linewidth]{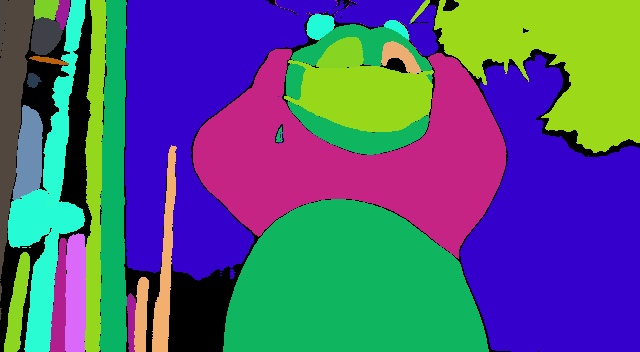}}{80} & \dynamicBaselineTimedImage{\includegraphics[width=\linewidth]{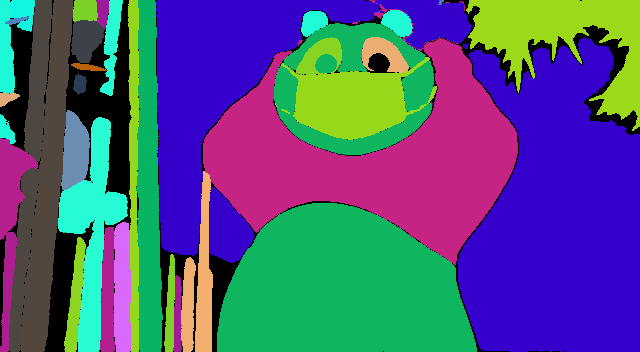}}{120} & \dynamicBaselineTimedImage{\includegraphics[width=\linewidth]{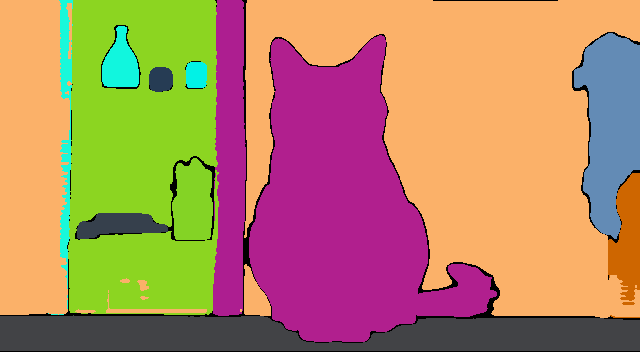}}{0} & \dynamicBaselineTimedImage{\includegraphics[width=\linewidth]{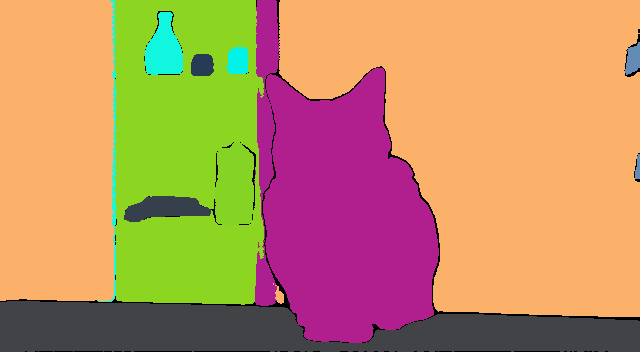}}{40} & \dynamicBaselineTimedImage{\includegraphics[width=\linewidth]{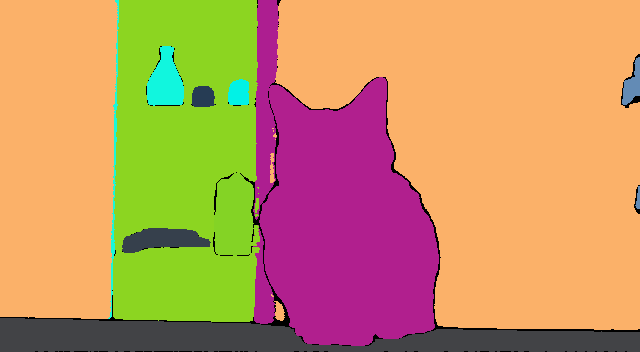}}{80} & \dynamicBaselineTimedImage{\includegraphics[width=\linewidth]{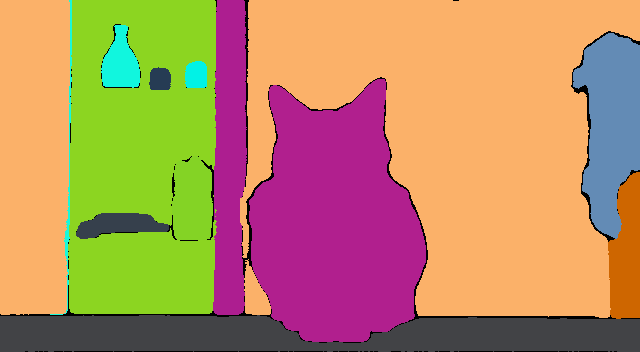}}{120} \\[-2.1pt]
\dynamicBaselineMethodLabel{Lifted\\[-1.5pt]teachers} & \includegraphics[width=\linewidth]{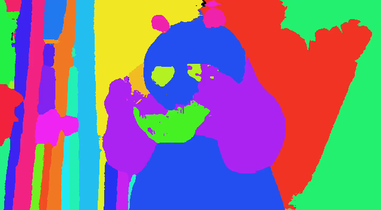} & \includegraphics[width=\linewidth]{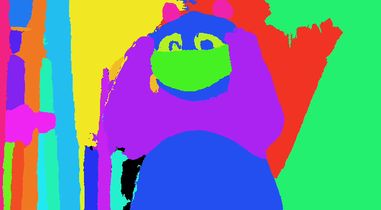} & \includegraphics[width=\linewidth]{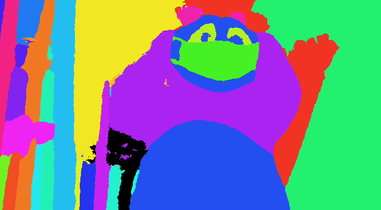} & \includegraphics[width=\linewidth]{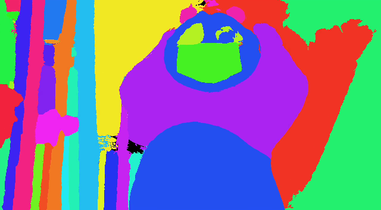} & \includegraphics[width=\linewidth]{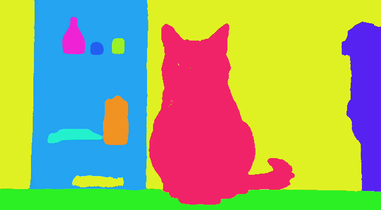} & \includegraphics[width=\linewidth]{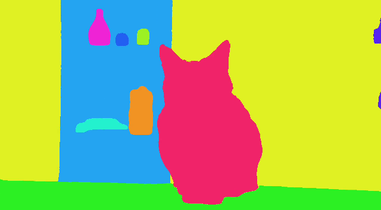} & \includegraphics[width=\linewidth]{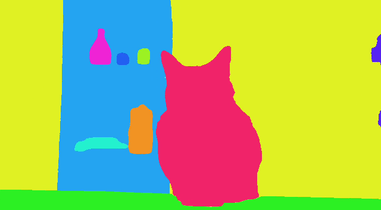} & \includegraphics[width=\linewidth]{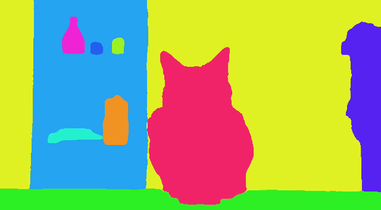} \\[-2.1pt]
\dynamicBaselineMethodLabel{TRASE} & \includegraphics[width=\linewidth]{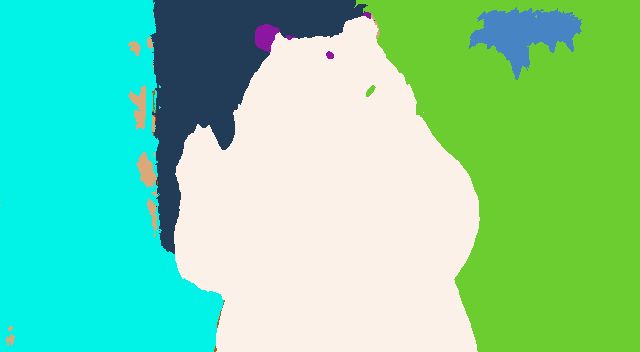} & \includegraphics[width=\linewidth]{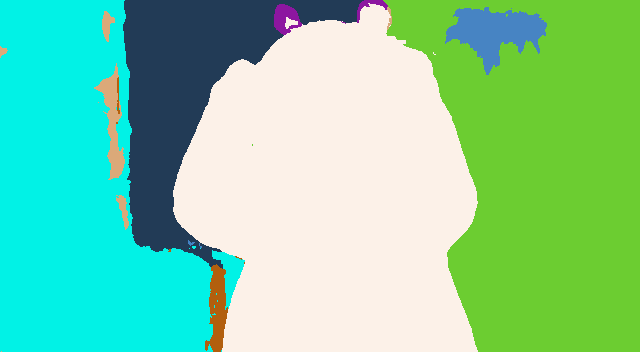} & \includegraphics[width=\linewidth]{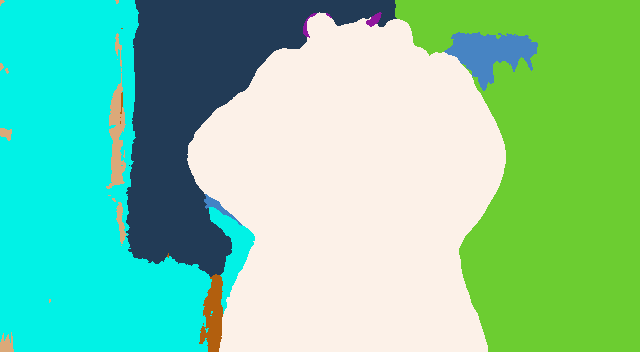} & \includegraphics[width=\linewidth]{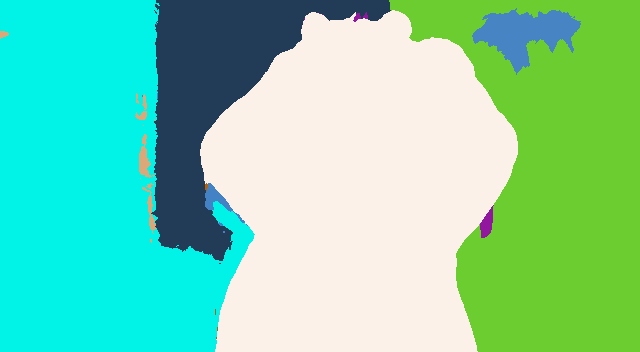} & \includegraphics[width=\linewidth]{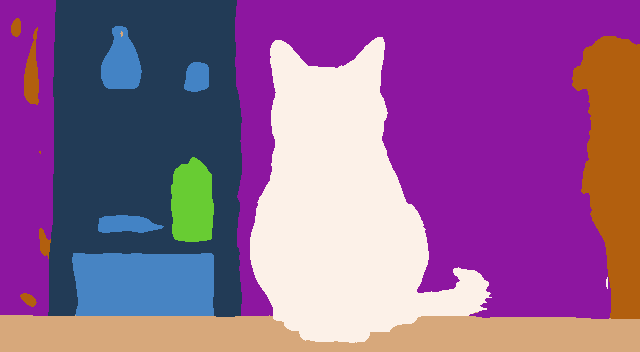} & \includegraphics[width=\linewidth]{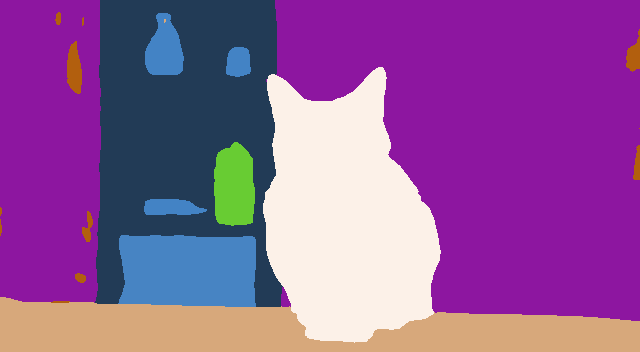} & \includegraphics[width=\linewidth]{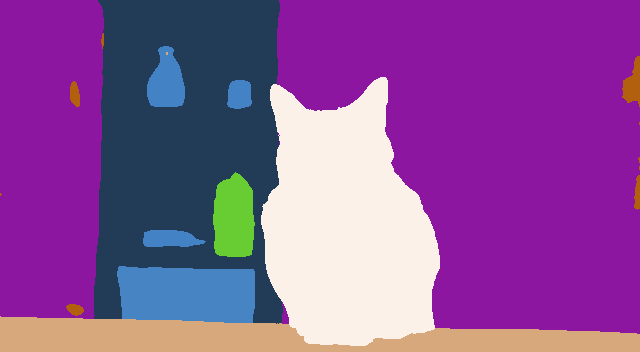} & \includegraphics[width=\linewidth]{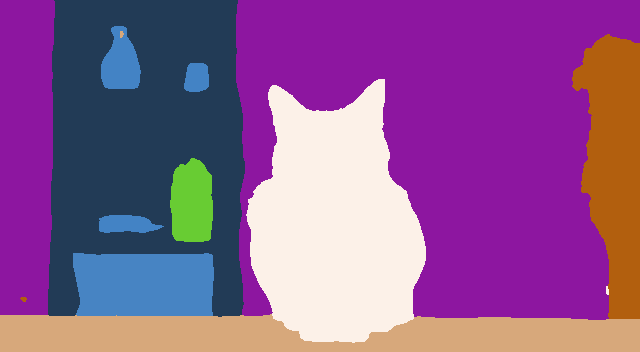} \\[-2.1pt]
\dynamicBaselineMethodLabel{IGGT4D} & \includegraphics[width=\linewidth]{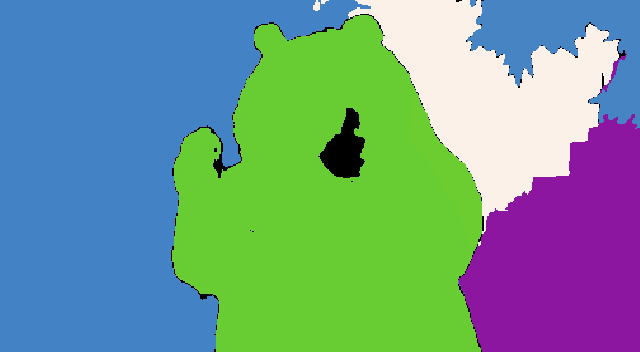} & \includegraphics[width=\linewidth]{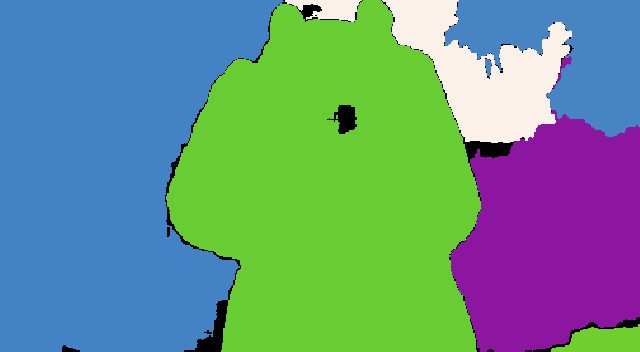} & \includegraphics[width=\linewidth]{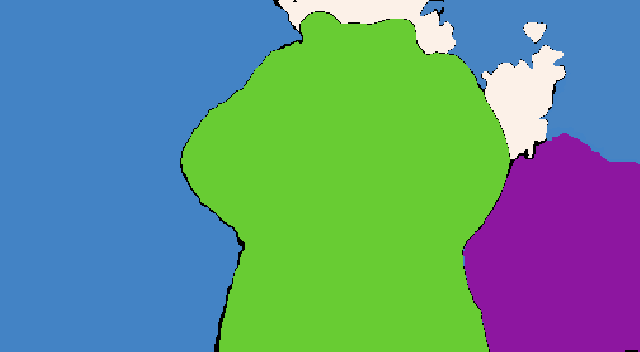} & \includegraphics[width=\linewidth]{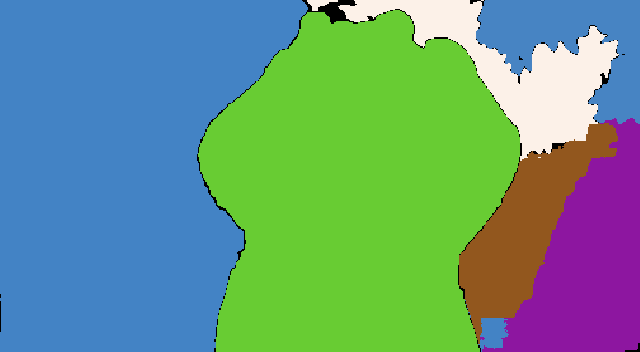} & \includegraphics[width=\linewidth]{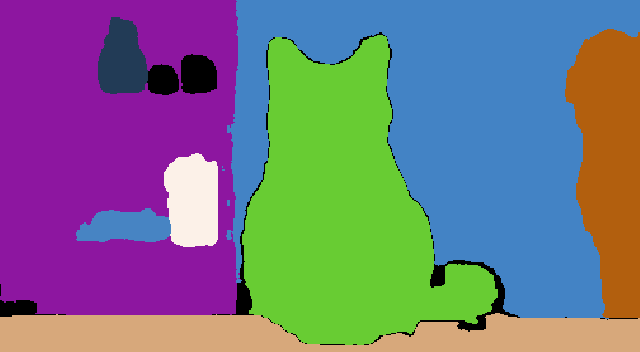} & \includegraphics[width=\linewidth]{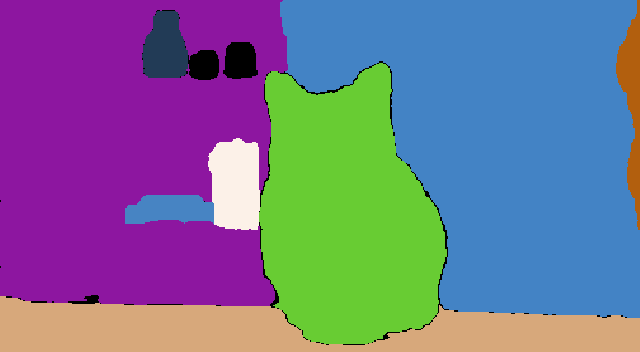} & \includegraphics[width=\linewidth]{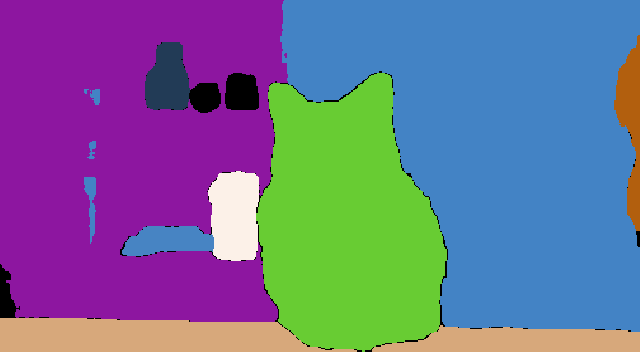} & \includegraphics[width=\linewidth]{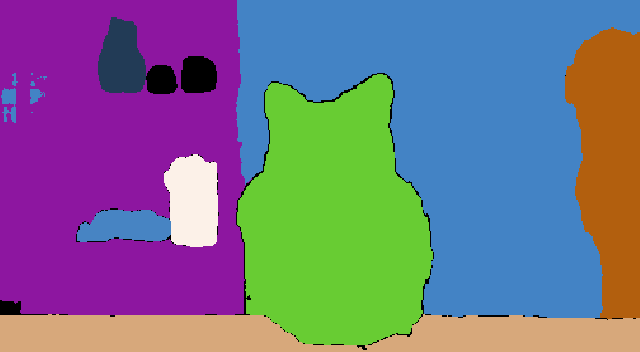} \\[-2.1pt]
\dynamicBaselineMethodLabel{Lucid\\[-1.5pt]Lyra} & \includegraphics[width=\linewidth]{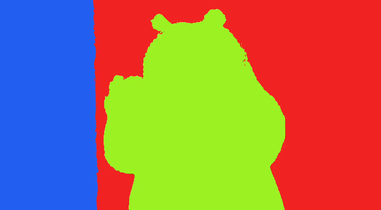} & \includegraphics[width=\linewidth]{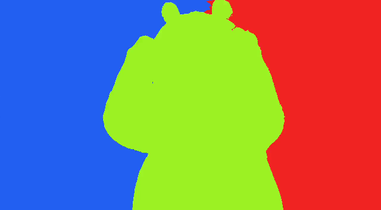} & \includegraphics[width=\linewidth]{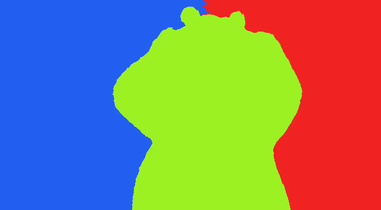} & \includegraphics[width=\linewidth]{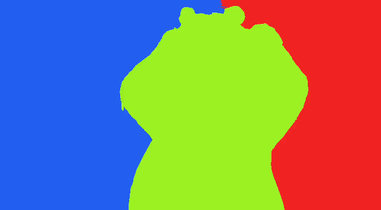} & \includegraphics[width=\linewidth]{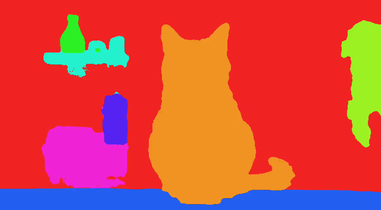} & \includegraphics[width=\linewidth]{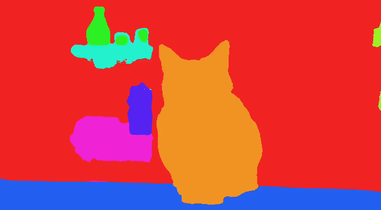} & \includegraphics[width=\linewidth]{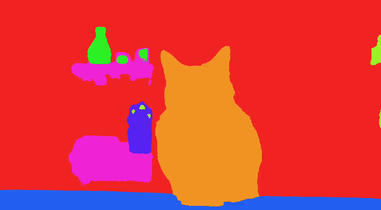} & \includegraphics[width=\linewidth]{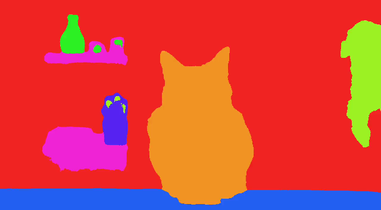} \\[-2.1pt]
\dynamicBaselineMethodLabel{Lucid\\[-1.5pt]Lyra+} & \includegraphics[width=\linewidth]{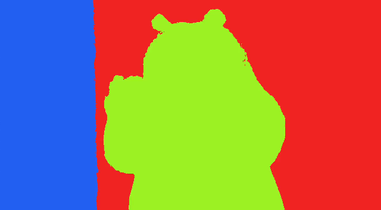} & \includegraphics[width=\linewidth]{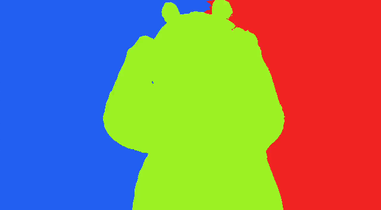} & \includegraphics[width=\linewidth]{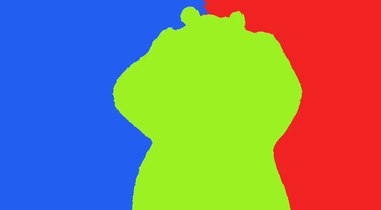} & \includegraphics[width=\linewidth]{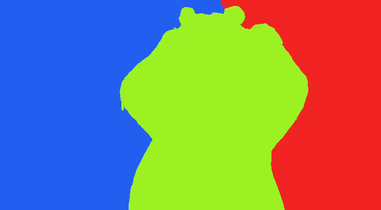} & \includegraphics[width=\linewidth]{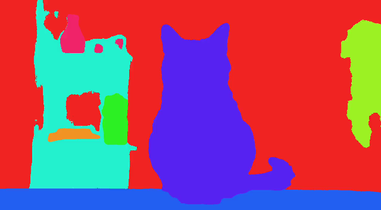} & \includegraphics[width=\linewidth]{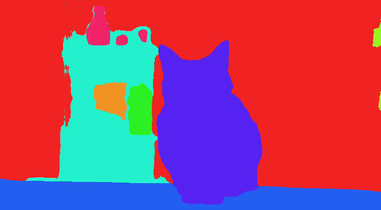} & \includegraphics[width=\linewidth]{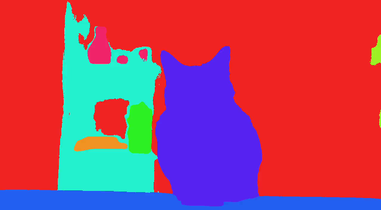} & \includegraphics[width=\linewidth]{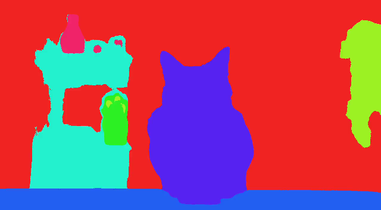} \\[-2.1pt]
\\[-2pt]
\end{tabular}
\endgroup

    \vspace{-0.5em}
    \caption{\textbf{Qualitative comparison on the dynamic evaluation set.} 
    We show times $t=\num{0},\num{40},\num{80},\num{120}$ along an orbiting camera trajectory. Both scene time and camera pose change. From top to bottom: RGB reconstruction, semantic segmentation, and instance segmentation. IGGT4D uses LSeg~\citep{li:2022:lss} for semantic predictions. Dynamic Lyra supplies the reconstruction used by the lifted-teacher baseline. Instance ID colors are random per method.}
    \label{fig:supp-qualitative-dynamic}
\end{figure}

\inlinesection{Dynamic scenes.}
In the dynamic setting, \MethodName better separates foreground objects from background surfaces and preserves finer semantic regions than 4D LangSplat~\citep{li:2025:4l4l} and IGGT4D~\citep{zou:2026:is4i} (\cref{fig:supp-qualitative-dynamic}). The dynamic baselines require different scene representations and cannot be directly applied to our independently decoded 3D Gaussian states. We therefore evaluate their respective reconstructions and report reconstruction and understanding metrics separately (\cref{tab:dynamic_main}).

For instance segmentation, \MethodName maintains coherent foreground masks across queried times, with fewer local gaps and cleaner boundaries than IGGT4D. TRASE~\citep{li:2026:tt4s} also recovers coherent foreground regions but produces coarser background partitions. The lifted-teacher reference remains competitive, particularly in separating smaller objects, while our method preserves complete foreground regions with less fragmentation.

\subsection{Hierarchical Instance Segmentation}
\begin{figure}[p]
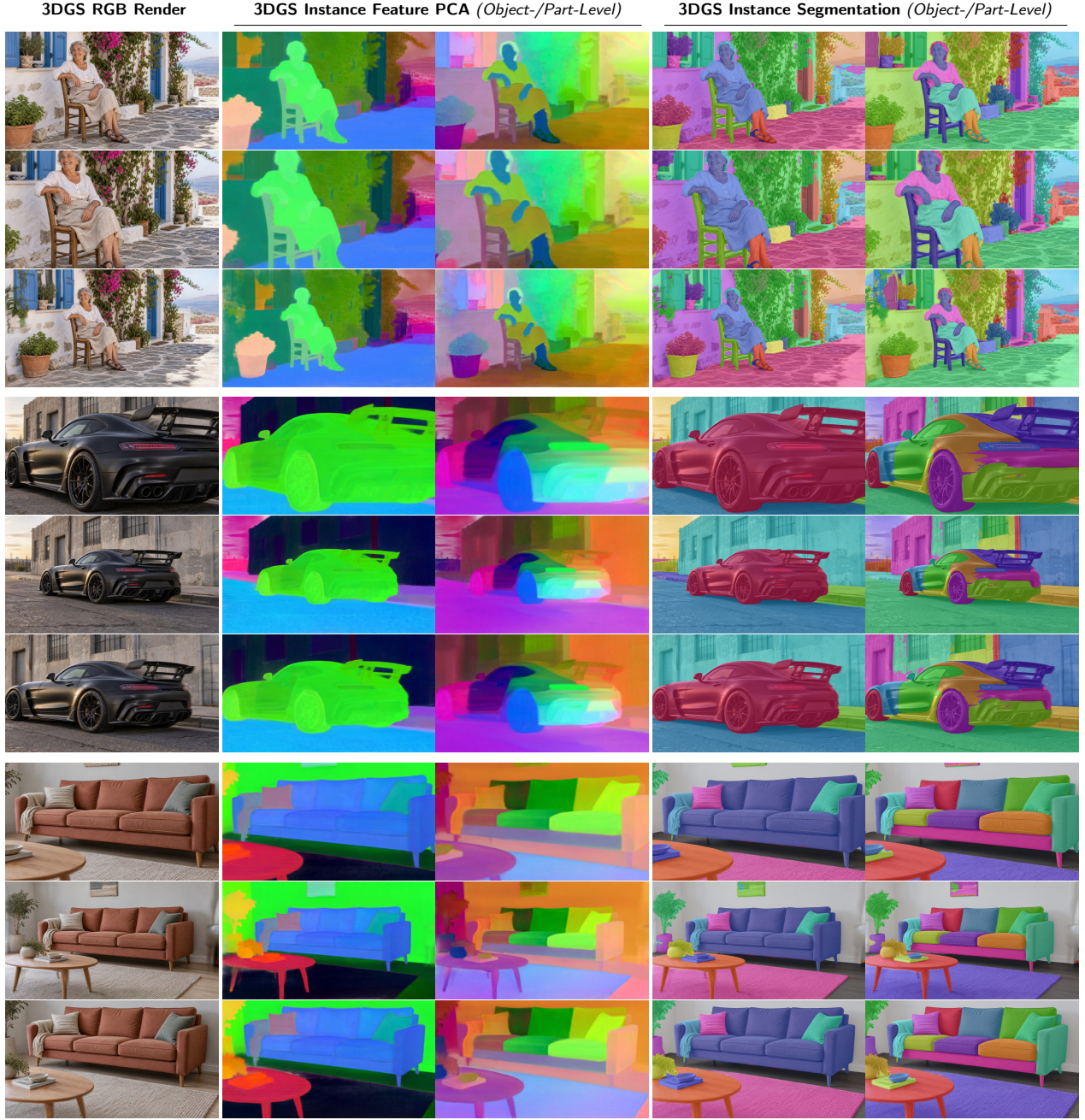

    \centering
    \scriptsize
\sffamily

\newcommand{\datapath}{%
    figures/supplement/hierarchical_instances/data/%
}

\setlength{\figgroupgap}{2pt}
\setlength{\figpanelwidth}{%
    \dimexpr\linewidth/5-2\figgroupgap/5\relax
}

\newcommand{\fullimage}[1]{%
    \includegraphics[width=\figpanelwidth]{\datapath#1}%
}

\setlength{\tabcolsep}{0pt}

\begin{tabular}{
    @{}c
    @{\hspace{\figgroupgap}}
    c@{}c
    @{\hspace{\figgroupgap}}
    c@{}c@{}
}
    \multicolumn{1}{c}{\textbf{3DGS RGB Render}} &
    \multicolumn{2}{c}{\textbf{3DGS Instance Feature PCA}~\emph{(Object-/Part-Level)}} &
    \multicolumn{2}{c}{\textbf{3DGS Instance Segmentation}~\emph{(Object-/Part-Level)}}
    \\
    \cmidrule(lr){2-3}
    \cmidrule(lr){4-5}

    \fullimage{semlyra_traj00_t000_rgb_recon.jpg} &
    \fullimage{semlyra_traj00_t000_instance_pca.jpg} &
    \fullimage{semlyra_traj00_t000_instance_part_pca.jpg} &
    \fullimage{semlyra_traj00_t000_instance_seg.png} &
    \fullimage{semlyra_traj00_t000_instance_part_clusters.png}
    \\[-2pt]

    \fullimage{semlyra_traj04_t120_rgb_recon.jpg} &
    \fullimage{semlyra_traj04_t120_instance_pca.jpg} &
    \fullimage{semlyra_traj04_t120_instance_part_pca.jpg} &
    \fullimage{semlyra_traj04_t120_instance_seg.png} &
    \fullimage{semlyra_traj04_t120_instance_part_clusters.png}
    \\[-2pt]

    \fullimage{semlyra_traj03_t120_rgb_recon.jpg} &
    \fullimage{semlyra_traj03_t120_instance_pca.jpg} &
    \fullimage{semlyra_traj03_t120_instance_part_pca.jpg} &
    \fullimage{semlyra_traj03_t120_instance_seg.png} &
    \fullimage{semlyra_traj03_t120_instance_part_clusters.png}
    \\[2pt]

    \fullimage{sports_car_street_09/semlyra_traj04_t120_rgb_recon.jpg} &
    \fullimage{sports_car_street_09/semlyra_traj04_t120_instance_pca.jpg} &
    \fullimage{sports_car_street_09/semlyra_traj04_t120_instance_part_pca.jpg} &
    \fullimage{sports_car_street_09/semlyra_traj04_t120_instance_seg.png} &
    \fullimage{sports_car_street_09/semlyra_traj04_t120_instance_part_clusters.png}
    \\[-2pt]

    \fullimage{sports_car_street_09/semlyra_traj03_t120_rgb_recon.jpg} &
    \fullimage{sports_car_street_09/semlyra_traj03_t120_instance_pca.jpg} &
    \fullimage{sports_car_street_09/semlyra_traj03_t120_instance_part_pca.jpg} &
    \fullimage{sports_car_street_09/semlyra_traj03_t120_instance_seg.png} &
    \fullimage{sports_car_street_09/semlyra_traj03_t120_instance_part_clusters.png}
    \\[-2pt]

    \fullimage{sports_car_street_09/semlyra_traj00_t100_rgb_recon.jpg} &
    \fullimage{sports_car_street_09/semlyra_traj00_t100_instance_pca.jpg} &
    \fullimage{sports_car_street_09/semlyra_traj00_t100_instance_part_pca.jpg} &
    \fullimage{sports_car_street_09/semlyra_traj00_t100_instance_seg.png} &
    \fullimage{sports_car_street_09/semlyra_traj00_t100_instance_part_clusters.png}
    \\[2pt]

    \fullimage{segmentable_couch_living_room_03/semlyra_traj04_t120_rgb_recon.jpg} &
    \fullimage{segmentable_couch_living_room_03/semlyra_traj04_t120_instance_pca.jpg} &
    \fullimage{segmentable_couch_living_room_03/semlyra_traj04_t120_instance_part_pca.jpg} &
    \fullimage{segmentable_couch_living_room_03/semlyra_traj04_t120_instance_seg.png} &
    \fullimage{segmentable_couch_living_room_03/semlyra_traj04_t120_instance_part_clusters.png}
    \\[-2pt]

    \fullimage{segmentable_couch_living_room_03/semlyra_traj03_t120_rgb_recon.jpg} &
    \fullimage{segmentable_couch_living_room_03/semlyra_traj03_t120_instance_pca.jpg} &
    \fullimage{segmentable_couch_living_room_03/semlyra_traj03_t120_instance_part_pca.jpg} &
    \fullimage{segmentable_couch_living_room_03/semlyra_traj03_t120_instance_seg.png} &
    \fullimage{segmentable_couch_living_room_03/semlyra_traj03_t120_instance_part_clusters.png}
    \\[-2pt]

    \fullimage{segmentable_couch_living_room_03/semlyra_traj02_t120_rgb_recon.jpg} &
    \fullimage{segmentable_couch_living_room_03/semlyra_traj02_t120_instance_pca.jpg} &
    \fullimage{segmentable_couch_living_room_03/semlyra_traj02_t120_instance_part_pca.jpg} &
    \fullimage{segmentable_couch_living_room_03/semlyra_traj02_t120_instance_seg.png} &
    \fullimage{segmentable_couch_living_room_03/semlyra_traj02_t120_instance_part_clusters.png}
    \\
\end{tabular}

    \vspace{-0.5em}
    \caption{\textbf{Object- and part-level instance segmentation.}
    We visualize \num{3} views per scene, showing 3DGS RGB render,
    object- and part-level 3DGS feature PCA, 3DGS object clusters, and within-object part
    clusters. Random colors denote cluster IDs.}
    \label{fig:hierarchical-instances}
\end{figure}
\label{ssec:supp-hierarchical-instances}
\MethodName infers two levels of instance representations that we aim to analyze qualitatively in \cref{fig:hierarchical-instances}.
The main object-level instance features group scene
entities, while additional part-level features expose finer regions within the recovered objects.
The qualitative results reveal a consistent feature hierarchy across all examples: object-level instance features remain similar over the spatial extent of an object, enabling clustering to recover a single instance mask, whereas part-level features preserve finer variations that distinguish individual object components. Because these representations form continuous feature fields, occasional grouping errors may still occur. While open-vocabulary querying and object-level instance segmentation provide the foundation for interacting with our generated scenes, part-level features offer a lightweight extension that enables finer-grained control over individual object components.

\section{Downstream Applications}
\label{sec:supp-downstream-applications}
\MethodName generates structured, queryable scenes that support diverse downstream applications. In the following section, we demonstrate this capability through examples of robotic simulation, 3D scene editing, and 3D asset mining. See our project website for video demos.

\subsection{Robot Simulation}
\label{ssec:supp-robot-simulation}

\begin{figure}[!htbp]
    \centering
    \input{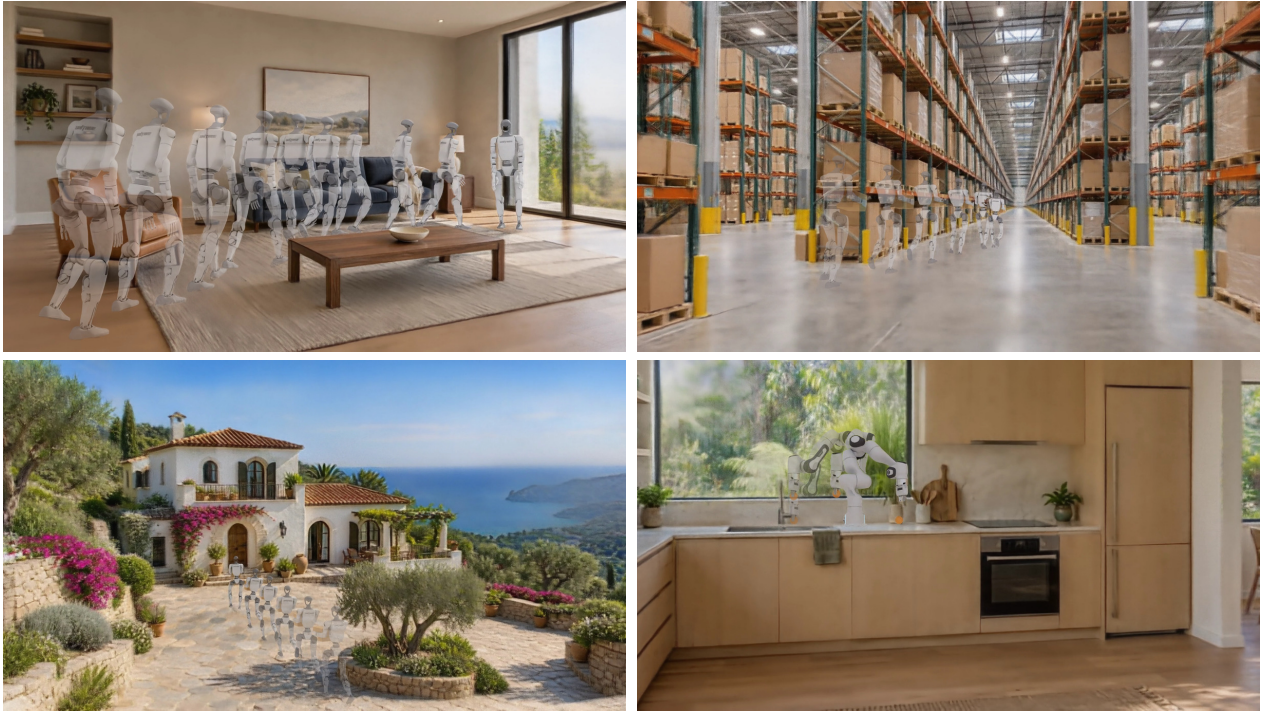}
    \caption{\textbf{Robot simulation in NVIDIA Isaac Sim using \MethodName scenes.}
    Four different robot-simulation scenarios built on \MethodName reconstructions.
    Extracted geometry provides collision surfaces, semantic labels support task planning,
    and Gaussian-splat rendering preserves the scene appearance.
    Overlaid robot poses show motion over time, with increasing opacity indicating temporal progression.}
    \label{fig:supp-robot-simulation}
\end{figure}

We extract a mesh from the \MethodName depth maps and camera parameters by computing a signed distance field on a hierarchy of sparse grids. We use the generated scene, extracted mesh and the semantic labels to set up a simulation in NVIDIA IsaacSim \citep{NVIDIA_Isaac_Sim}. For navigation, the semantic labels partition the scene into free space and obstacles, over which we build an occupancy grid and plan a route to a semantically specified goal using a clearance-weighted Dijkstra search. A robot (Unitree G1 humanoid) then follows this route with a pure-pursuit controller while a learned rough-terrain locomotion policy drives its joints under full rigid-body physics.
The humanoid also carries a box rigidly attached near its torso to demonstrate loco-manipulation. For manipulation, the \MethodName semantic labeling localizes a kitchen counter and sink in a separate reconstruction, from which we derive a mount pose and reach-constrained pick-and-place targets for a stationary Franka robotic arm. The arm is simulated with full articulated-body dynamics and a friction-based parallel-jaw grasp under differential inverse-kinematics control, picks a small synthetic object (a sphere) from the sink and places it on the counter. The extracted mesh serves as the collision proxy during simulation, while the original Gaussian-splat reconstruction~\citep{kerbl:2023:3gsf} is rendered for visual fidelity. \Cref{fig:supp-robot-simulation} visualizes the simulation examples.

\subsection{3D Scene Editing}
We implement a simple browser-based editor. The user can select objects through text queries or clicks. Text selection intersects the predicted semantic class with an instance ID, thereby separating objects of the same category, while click selection retrieves the semantic--instance pair of the Gaussian center nearest the camera ray. Users can further refine the selection using separate similarity thresholds for Gaussian semantic and instance features relative to prototypes computed from the initial support.

\begin{figure}[!htbp]
    \centering
\begingroup
\scriptsize\sffamily
\noindent
\begin{minipage}[c]{0.35\linewidth}
  \centering
  \textbf{\MethodName Editing GUI}\par\smallskip
  \includegraphics[width=\linewidth]{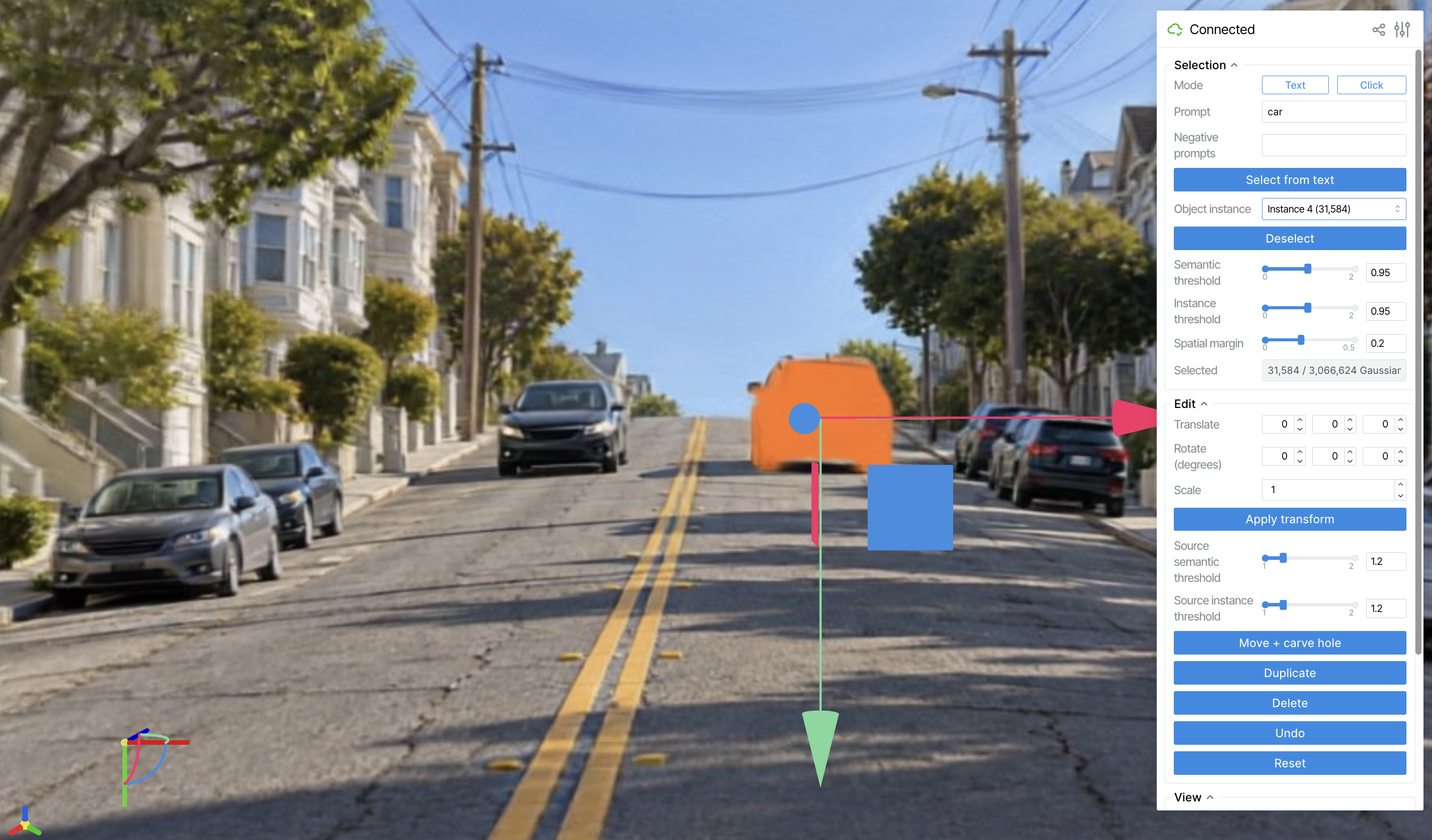}
\end{minipage}\hfill
\begin{minipage}[c]{0.63\linewidth}
  \centering
  \setlength{\tabcolsep}{1pt}
  \begin{tabular}{@{}*{3}{>{\centering\arraybackslash}m{0.32\linewidth}}@{}}
    \textbf{Initial 3DGS Scene} & \textbf{Edits} & \textbf{Edited 3DGS Scene} \\[2pt]
    \includegraphics[width=\linewidth]{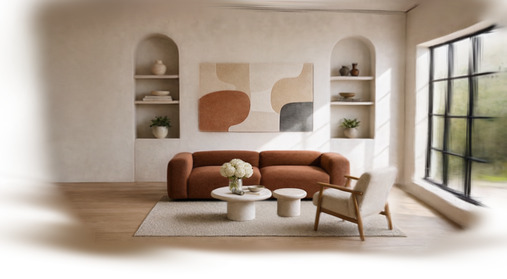} &
    \includegraphics[width=\linewidth]{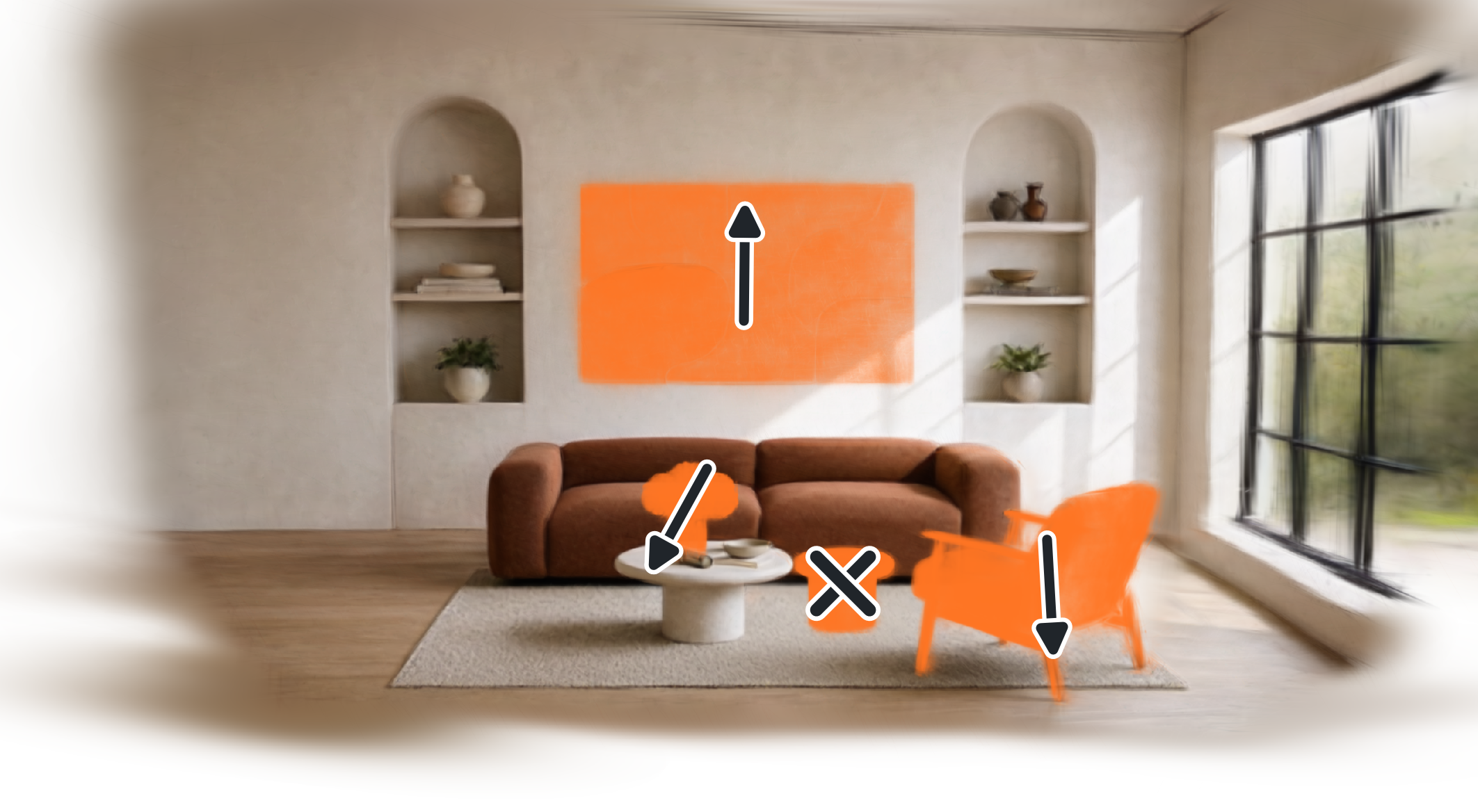} &
    \includegraphics[width=\linewidth]{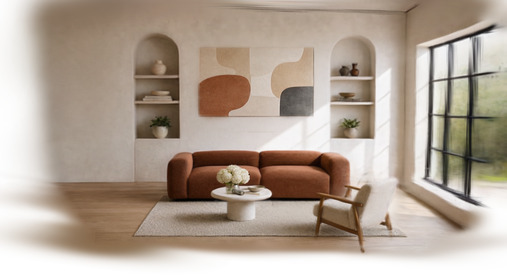} \\
    \includegraphics[width=\linewidth]{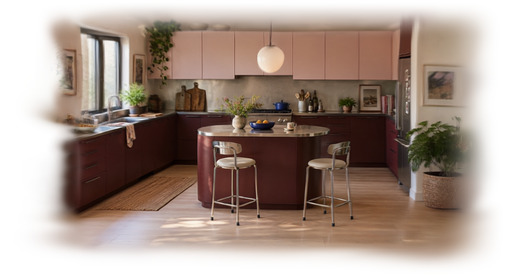} &
    \includegraphics[width=\linewidth]{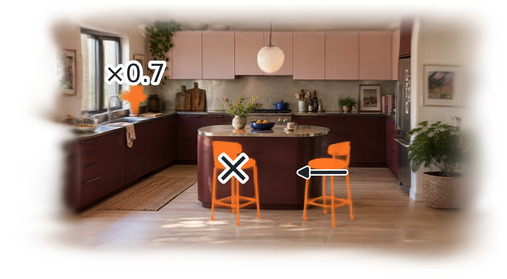} &
    \includegraphics[width=\linewidth]{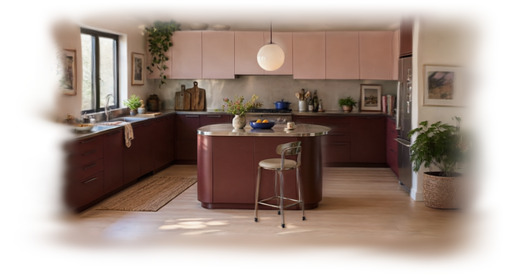} \\
  \end{tabular}
\end{minipage}
\endgroup

    \caption{\textbf{\MethodName 3DGS Scene Editing.} \emph{Left:} Our GUI provides an interface for natural interaction with the generated Gaussian scenes.
    Gaussians are selected via text prompts or via clicking and can then be manipulated interactively. 
    \emph{Right:} Editing examples showing the initial 3DGS scene, the applied edits and the final scene.}
    \label{fig:supp-scene-editing}
\end{figure}

Translation, rotation, scaling, duplication, and deletion operate directly on the selected Gaussian parameters. Transformations are applied around the selection centroid and update positions, anisotropic scales, and orientation quaternions, while color, opacity, and learned features remain attached to each primitive. Duplication appends a transformed copy, whereas deletion removes the selected primitives.
Moving or deleting an object can expose unobserved regions. After the user commits all edits, OmniPaint~\citep{yu:2025:omnipaint} inpaints the original object support, and a subsequent \MethodName forward pass reconstructs a clean background scene. We retain only reconstructed Gaussians near the edit whose projections fall within the exposed source support, excluding the transformed object's footprint. Opacity tapering converts these Gaussians into a smoothly blended local background patch. \Cref{fig:supp-scene-editing} visualizes our GUI as well as editing examples.

\subsection{3D Asset Mining}
\label{ssec:supp-articulated-assets}

Much like in 3D scene editing, we can make use of the generated appearance from multiple views, together with the instance and semantic segmentation masks, to mine high-quality 3D assets from the generated scenes.

\begin{figure*}[!t]
  \centering
  \setlength{\tabcolsep}{0pt}
  \captionsetup[subfigure]{
    font=scriptsize,
    skip=2pt,
    justification=centering,
    singlelinecheck=false
  }

  \newcommand{\assetrowlabels}[3]{%
    \raisebox{-0.5\height}{%
      \vbox{\offinterlineskip
        \vbox to #1{\vfil\hbox{\rotatebox{90}{#2}}\vfil}%
        \vbox to #1{\vfil\hbox{\rotatebox{90}{#3}}\vfil}%
      }%
    }%
  }

  \begin{subfigure}{\linewidth}
    \centering
    \includegraphics[width=\linewidth]{%
      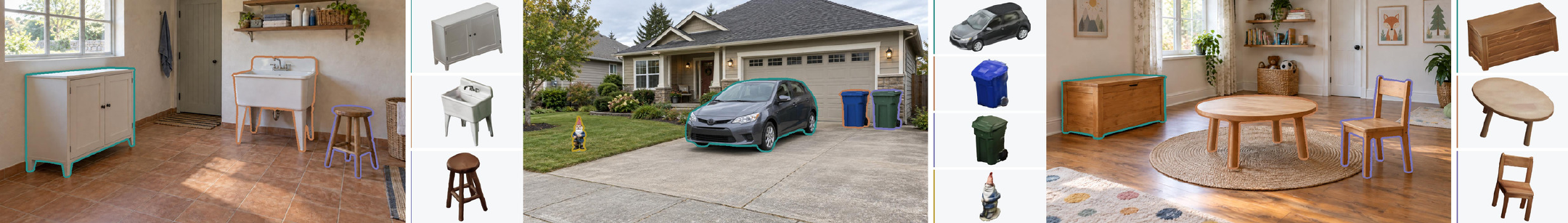}
    \caption{Asset selection and extraction.}
    \label{fig:supp-scene-assets}
  \end{subfigure}

  \par\vspace{6pt}

  \begin{subfigure}{\linewidth}
    \centering
    \begingroup
      \scriptsize\sffamily
      \begin{tabular}{@{}ccccc@{}}
        & \makebox[.2425\linewidth]{Sink}
        & \makebox[.2425\linewidth]{Cabinet}
        & \makebox[.2425\linewidth]{Garden gnome}
        & \makebox[.2425\linewidth]{Chair} \\[2pt]
        \makebox[.03\linewidth]{%
          \assetrowlabels{.095\linewidth}{Single-view}{Multi-view}%
        }
        & \multicolumn{4}{@{}c@{}}{%
          \raisebox{-0.5\height}{%
            \includegraphics[width=.97\linewidth]{%
              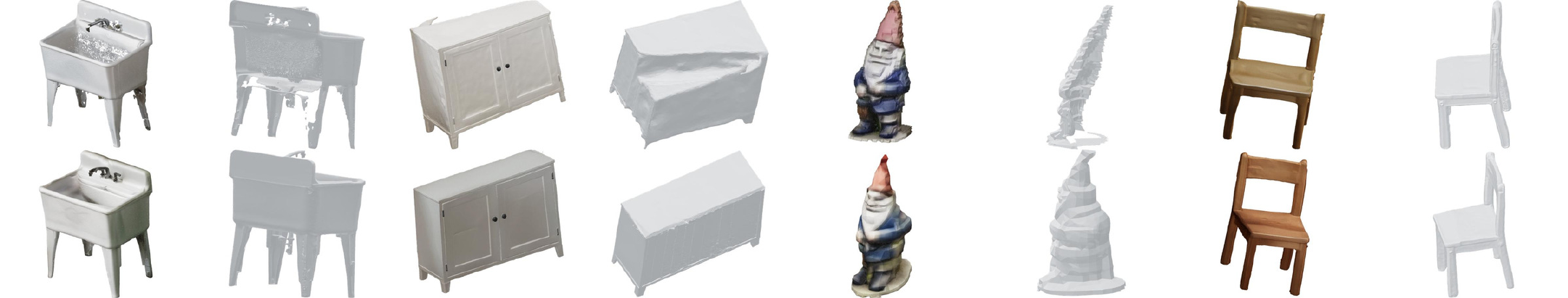}%
          }%
        }
      \end{tabular}
      \par
    \endgroup
    \caption{Single-view vs.\ multi-view reconstruction.}
    \label{fig:supp-views-assets}
  \end{subfigure}

  \par\vspace{6pt}

  \begin{subfigure}{\linewidth}
    \centering
    \begingroup
      \scriptsize\sffamily
      \begin{tabular}{@{}cc@{}}
        \makebox[.07\linewidth]{%
          \assetrowlabels{.155\linewidth}{Texture}{Parts}%
        }
        & \raisebox{-0.5\height}{%
          \includegraphics[width=.93\linewidth]{%
            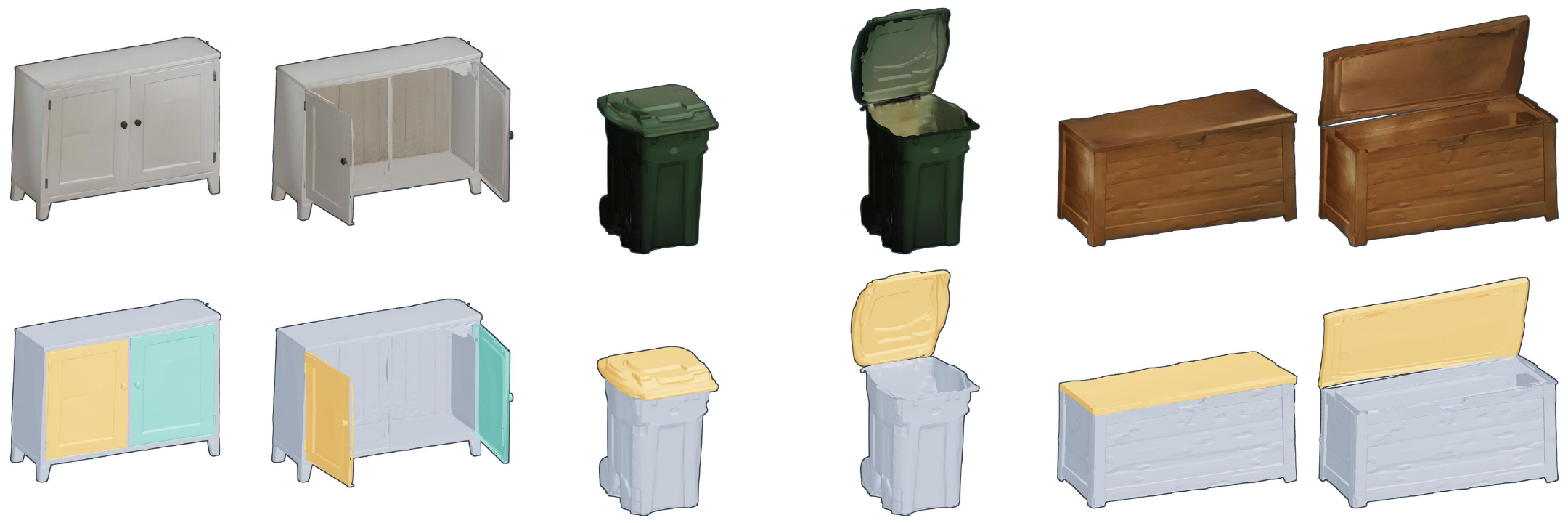}%
        }
      \end{tabular}
      \par
    \endgroup
    \caption{Estimated articulation of extracted assets.}
    \label{fig:supp-articulated-assets}
  \end{subfigure}

  \caption{%
    \textbf{Asset extraction and articulation.}
    \emph{(a)} Selected instances alongside their textured reconstructions.
    Display scales are independent.
    \emph{(b)} Selected single-view (top) and multi-view (bottom) Pixal3D
    results. Each result pairs a textured view with an alternate
    neutral-material view.
    \emph{(c)} Articulated assets shown closed and open, with original
    textures and part segmentation color. Part boundaries and hinges
    are obtained through PARTICULATE~\citep{Li:2026:particulate} on the
    generated asset meshes. Doors open to $\num{85}^{\circ}$ and lids
    to $\num{75}^{\circ}$ for illustration, not as measured joint limits.
  }
  \label{fig:supp-asset-extraction}
\end{figure*}

For each selected object, we extract six segmented views, prioritizing viewpoint diversity and object coverage without occlusion. The segmented RGB observations together with the corresponding camera parameters are provided to the pretrained Pixal3D~\citep{Li:2026:pixal} multi-view reconstruction model to produce a textured mesh. \Cref{fig:supp-scene-assets} presents assets extracted from \MethodName scenes. Comparisons in \cref{fig:supp-views-assets} show that the incorporation of multi-view observations can substantially improve structural consistency compared to single-view reconstruction without \MethodName.

Furthermore, the generated meshes can also be post-processed by mesh-to-articulation pipelines. In \cref{fig:supp-articulated-assets}, we employ PARTICULATE~\citep{Li:2026:particulate} to estimate moving parts and the corresponding articulation kinematic tree. 

\section{Datasets and Annotations}
\label{sec:supp-datasets}
We augment the PhysicalAI-SpatialIntelligence-Lyra-SDG~\citep{bahmani:2026:g3sr}
with semantic and instance targets. For evaluation, we introduce three benchmarks based on PhysicalAI-SpatialIntelligence-Lyra-SDG and ADE20K~\citep{zhou:2019:sust}.

\subsection{Static Datasets}

\begin{figure}[t]
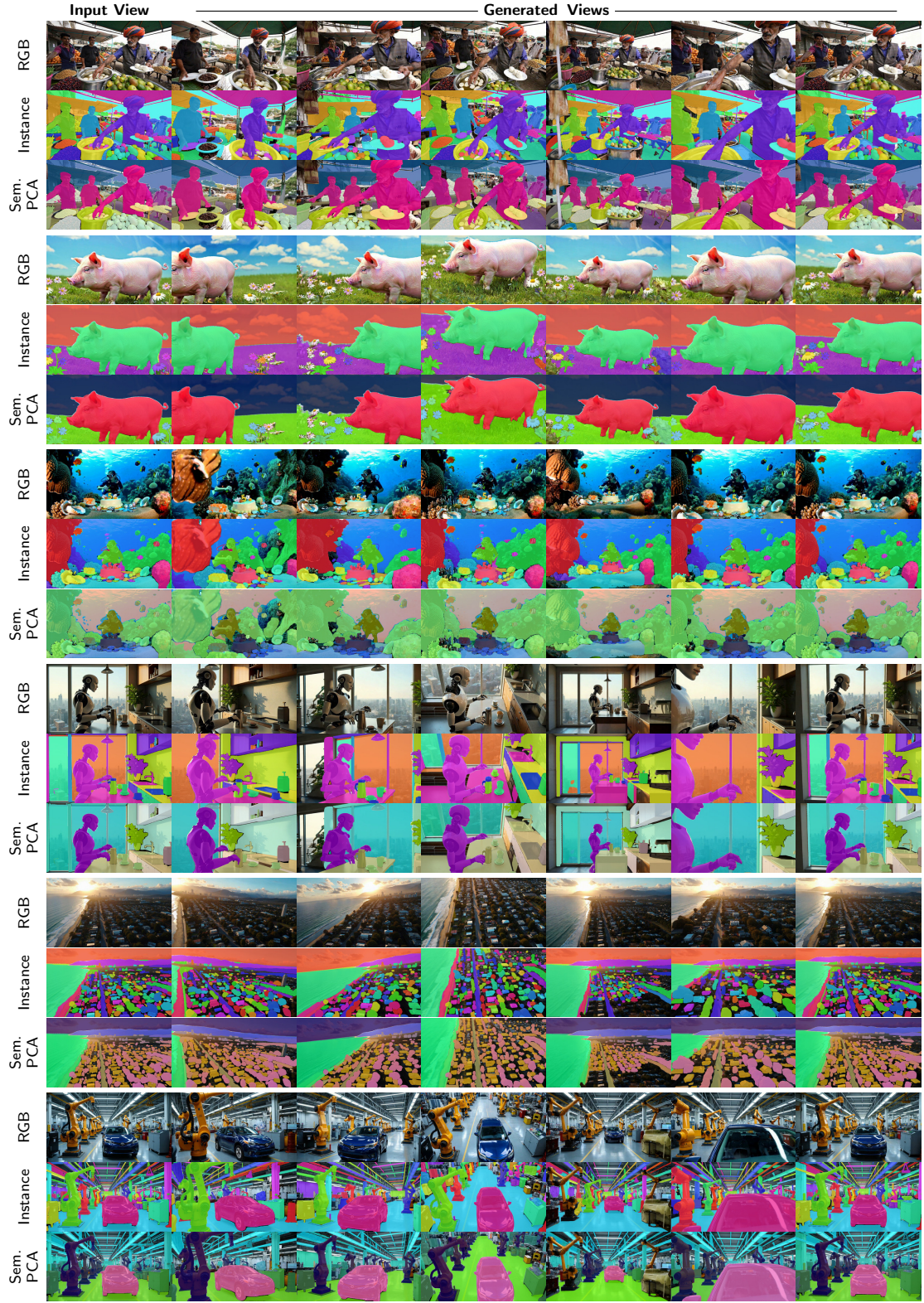

    \centering
    \begin{adjustbox}{max totalsize={\linewidth}{0.98\textheight},center}
    \scriptsize
\sffamily
\setlength{\tabcolsep}{0pt}
\newcommand{\lyratrainimgwidth}{0.12}
\newcommand{\lyratrainimage}[2]{%
    \includegraphics[width=\linewidth]{figures/supplement/lyra_train_static/#1/#2}%
}
\newcommand{\lyratrainscene}[1]{%
\rotatebox[origin=lB]{90}{RGB}
& \lyratrainimage{#1}{traj00_t000_rgb.jpg}
& \lyratrainimage{#1}{traj00_t120_rgb.jpg}
& \lyratrainimage{#1}{traj01_t120_rgb.jpg}
& \lyratrainimage{#1}{traj02_t120_rgb.jpg}
& \lyratrainimage{#1}{traj03_t120_rgb.jpg}
& \lyratrainimage{#1}{traj04_t120_rgb.jpg}
& \lyratrainimage{#1}{traj05_t100_rgb.jpg}\\[-2.25pt]
\rotatebox[origin=lB]{90}{Instance}
& \lyratrainimage{#1}{traj00_t000_instance_supervision}
& \lyratrainimage{#1}{traj00_t120_instance_supervision}
& \lyratrainimage{#1}{traj01_t120_instance_supervision}
& \lyratrainimage{#1}{traj02_t120_instance_supervision}
& \lyratrainimage{#1}{traj03_t120_instance_supervision}
& \lyratrainimage{#1}{traj04_t120_instance_supervision}
& \lyratrainimage{#1}{traj05_t100_instance_supervision}\\[-2.25pt]
\rotatebox[origin=lB]{90}{\shortstack{Sem.\\PCA}}
& \lyratrainimage{#1}{traj00_t000_semantic_supervision_pca.jpg}
& \lyratrainimage{#1}{traj00_t120_semantic_supervision_pca.jpg}
& \lyratrainimage{#1}{traj01_t120_semantic_supervision_pca.jpg}
& \lyratrainimage{#1}{traj02_t120_semantic_supervision_pca.jpg}
& \lyratrainimage{#1}{traj03_t120_semantic_supervision_pca.jpg}
& \lyratrainimage{#1}{traj04_t120_semantic_supervision_pca.jpg}
& \lyratrainimage{#1}{traj05_t100_semantic_supervision_pca.jpg}\\[6pt]
}

\begin{tabular}{
>{\centering\arraybackslash} m{0.045\textwidth}
>{\centering\arraybackslash} m{\lyratrainimgwidth\textwidth}
>{\centering\arraybackslash} m{\lyratrainimgwidth\textwidth}
>{\centering\arraybackslash} m{\lyratrainimgwidth\textwidth}
>{\centering\arraybackslash} m{\lyratrainimgwidth\textwidth}
>{\centering\arraybackslash} m{\lyratrainimgwidth\textwidth}
>{\centering\arraybackslash} m{\lyratrainimgwidth\textwidth}
>{\centering\arraybackslash} m{\lyratrainimgwidth\textwidth}
}
& \textbf{Input View}
& \multicolumn{6}{c}{\rule[0.5ex]{4.5cm}{0.4pt}~\textbf{Generated\phantom{p}Views}~\rule[0.5ex]{4.5cm}{0.4pt}} \\
\lyratrainscene{0_000025_fbf9a39bfd7a454cb5b538d475651fb9}
\lyratrainscene{0_000046_0867a1e46fc542af9f44932bb164263b}
\lyratrainscene{1_000048_134224e7ad2349c0bcbcbde1061d91cc}
\lyratrainscene{6_0_00170}
\lyratrainscene{11_0_00033}
\lyratrainscene{12_0_00294}
\end{tabular}

    \end{adjustbox}

    \caption{\textbf{\MethodName-SDG static training data.}
    RGB frames, instance tracks, and PCA visualizations of semantic
    teacher features across generated camera trajectories.}
    \label{fig:lyra-static-train-examples}
\end{figure}

\inlinesection{Training data.}
The static Lyra-SDG corpus comprises \num{59031} image-conditioned scenes spanning highly diverse indoor and outdoor environments, people, animals, and various imaginative settings.
Each scene provides six generated camera trajectories of \num{121} frames at
$\num{704}{\times}\num{1280}$, together with depth, camera parameters, and video latents.
\Cref{fig:lyra-static-train-examples} visualizes random training data samples.

\begin{figure}[t]
    \centering
    
\scriptsize
\sffamily
\setlength{\tabcolsep}{0pt}
\newcommand{\imgwidth}{0.139}

\begin{tabular}{
>{\centering\arraybackslash} m{0.027\textwidth}
>{\centering\arraybackslash} m{\imgwidth\textwidth}
>{\centering\arraybackslash} m{\imgwidth\textwidth}
>{\centering\arraybackslash} m{\imgwidth\textwidth}
>{\centering\arraybackslash} m{\imgwidth\textwidth}
>{\centering\arraybackslash} m{\imgwidth\textwidth}
>{\centering\arraybackslash} m{\imgwidth\textwidth}
>{\centering\arraybackslash} m{\imgwidth\textwidth}
}

& \textbf{Input View}
& \multicolumn{6}{c}{\rule[0.5ex]{4.5cm}{0.4pt}~\textbf{Generated\phantom{p}Views}~\rule[0.5ex]{4.5cm}{0.4pt}} \\

\rotatebox[origin=lB]{90}{\hspace{-0.25em}{RGB}}
& \includegraphics[width=\linewidth]{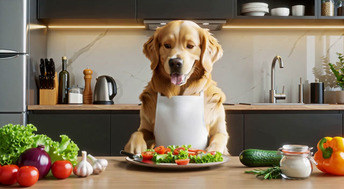} 
& \includegraphics[width=\linewidth]{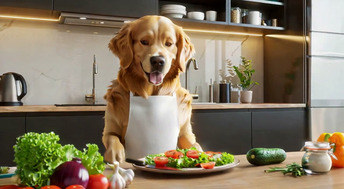} 
& \includegraphics[width=\linewidth]{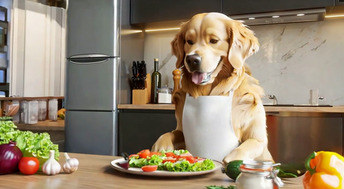} 
& \includegraphics[width=\linewidth]{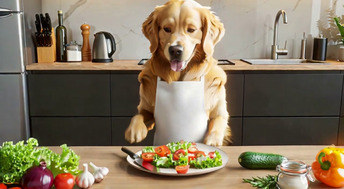} 
& \includegraphics[width=\linewidth]{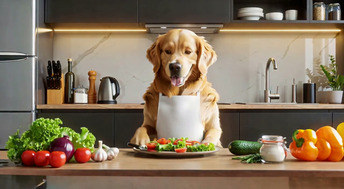} 
& \includegraphics[width=\linewidth]{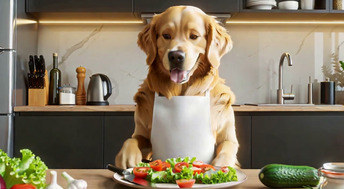} 
& \includegraphics[width=\linewidth]{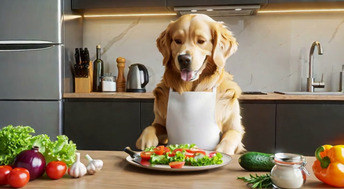}\\[-2.25pt]

\rotatebox[origin=lB]{90}{\hspace{-0.25em}{Instance}}
& \includegraphics[width=\linewidth]{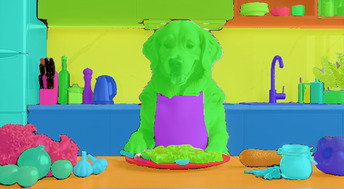}
& \includegraphics[width=\linewidth]{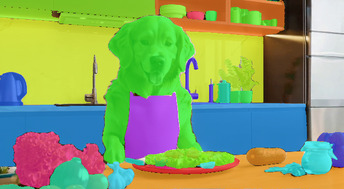}
& \includegraphics[width=\linewidth]{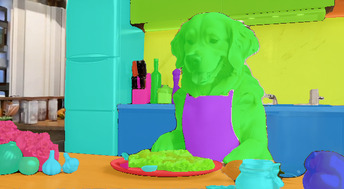}
& \includegraphics[width=\linewidth]{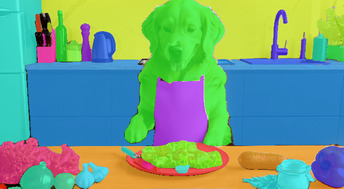}
& \includegraphics[width=\linewidth]{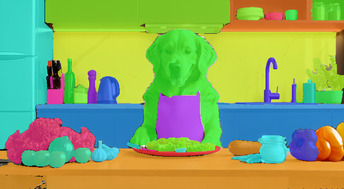}
& \includegraphics[width=\linewidth]{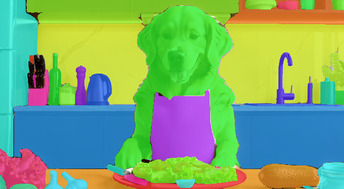}
& \includegraphics[width=\linewidth]{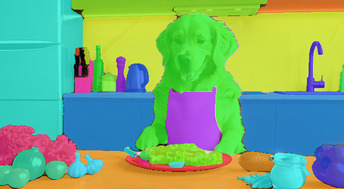}\\[-2.25pt]

\rotatebox[origin=lB]{90}{\hspace{-0.25em}{Semantic}}
& \includegraphics[width=\linewidth]{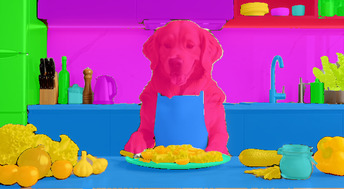}
& \includegraphics[width=\linewidth]{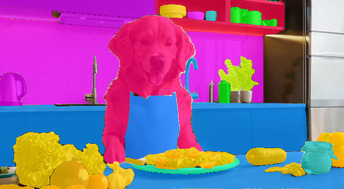}
& \includegraphics[width=\linewidth]{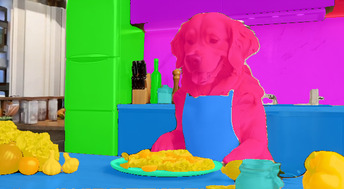}
& \includegraphics[width=\linewidth]{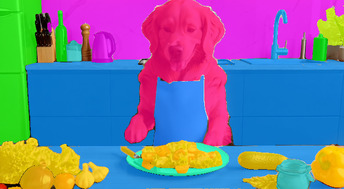}
& \includegraphics[width=\linewidth]{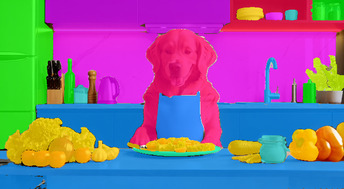}
& \includegraphics[width=\linewidth]{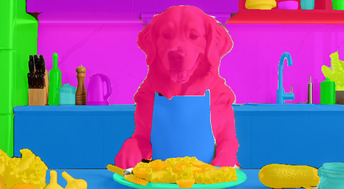}
& \includegraphics[width=\linewidth]{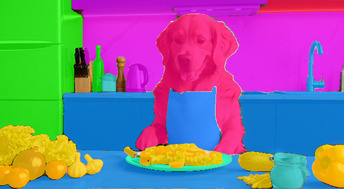}\\[-2.5pt]
& \multicolumn{7}{c}{%
    \begingroup
    \scriptsize
    \sffamily
    \setlength{\tabcolsep}{0pt}%
    \renewcommand{\arraystretch}{0.9}%
    \begin{tabularx}{0.973\textwidth}{*{4}{>{\centering\arraybackslash}X}>{\centering\arraybackslash}p{6em}*{8}{>{\centering\arraybackslash}X}}
        \cellcolor{cabinet}\textcolor{white}{Cabinet} &
        \cellcolor{plant}\textcolor{white}{Plant} &
        \cellcolor{shelf}\textcolor{white}{Shelf} &
        \cellcolor{sink}\textcolor{white}{Sink} &
        \cellcolor{refrigerator}\textcolor{white}{Refrigerator} &
        \cellcolor{countertop}\textcolor{white}{Counter} &
        \cellcolor{apparel}\textcolor{white}{Apparel} &
        \cellcolor{bottle}\textcolor{white}{Bottle} &
        \cellcolor{food}\textcolor{white}{Food} &
        \cellcolor{pot}\textcolor{white}{Pot} &
        \cellcolor{animal}\textcolor{white}{Animal} &
        \cellcolor{plate}\textcolor{white}{Plate} &
        \cellcolor{glass}\textcolor{white}{Glass}
    \end{tabularx}%
    \endgroup
}\\[1em]

\rotatebox[origin=lB]{90}{\hspace{-0.25em}{RGB}}
& \includegraphics[width=\linewidth]{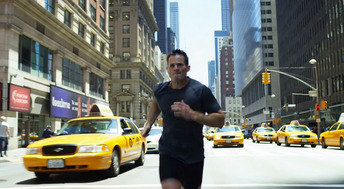} 
& \includegraphics[width=\linewidth]{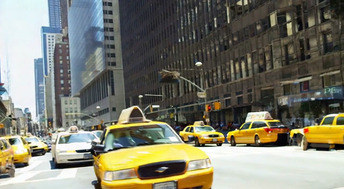} 
& \includegraphics[width=\linewidth]{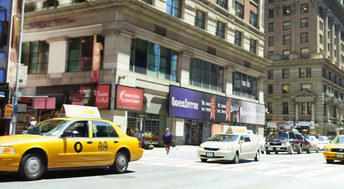} 
& \includegraphics[width=\linewidth]{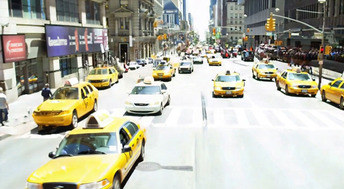} 
& \includegraphics[width=\linewidth]{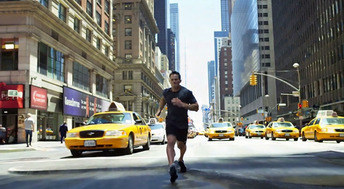} 
& \includegraphics[width=\linewidth]{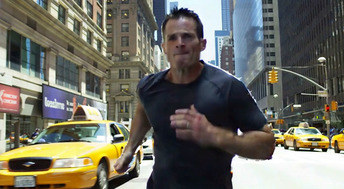} 
& \includegraphics[width=\linewidth]{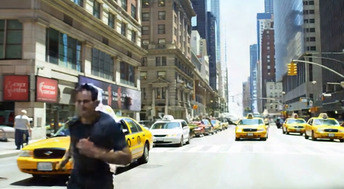}\\[-2.25pt]

\rotatebox[origin=lB]{90}{\hspace{-0.25em}{Instance}}
& \includegraphics[width=\linewidth]{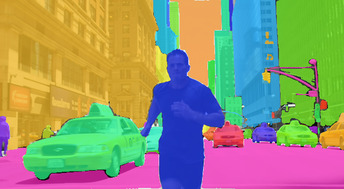}
& \includegraphics[width=\linewidth]{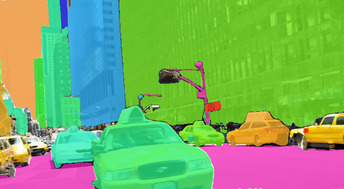}
& \includegraphics[width=\linewidth]{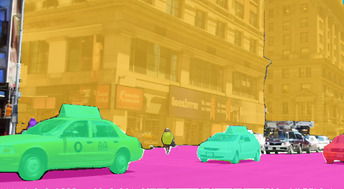}
& \includegraphics[width=\linewidth]{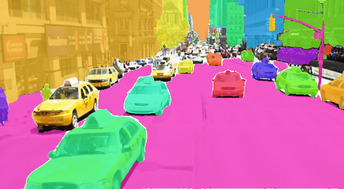}
& \includegraphics[width=\linewidth]{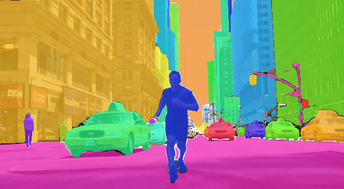}
& \includegraphics[width=\linewidth]{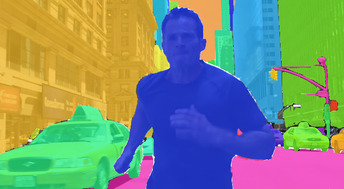}
& \includegraphics[width=\linewidth]{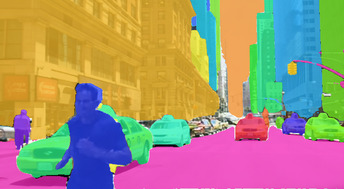}\\[-2.25pt]

\rotatebox[origin=lB]{90}{\hspace{-0.25em}{Semantic}}
& \includegraphics[width=\linewidth]{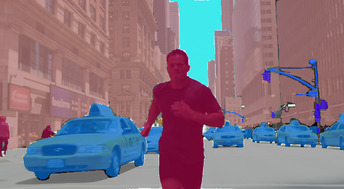}
& \includegraphics[width=\linewidth]{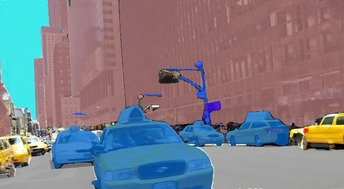}
& \includegraphics[width=\linewidth]{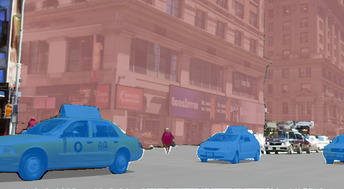}
& \includegraphics[width=\linewidth]{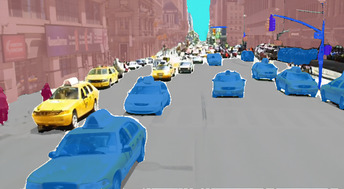}
& \includegraphics[width=\linewidth]{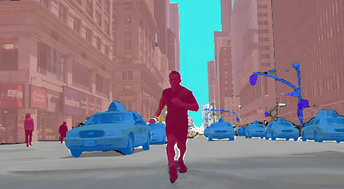}
& \includegraphics[width=\linewidth]{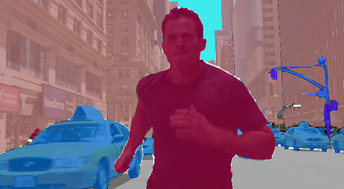}
& \includegraphics[width=\linewidth]{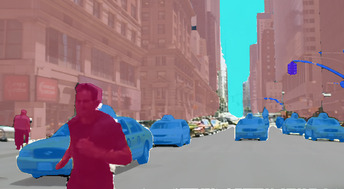}\\[-2.5pt]
& \multicolumn{7}{c}{%
    \begingroup
    \scriptsize
    \sffamily
    \setlength{\tabcolsep}{0pt}%
    \renewcommand{\arraystretch}{0.9}%
    \begin{tabularx}{0.973\textwidth}{*{7}{>{\centering\arraybackslash}X}}
        \cellcolor{building}\textcolor{white}{Building} &
        \cellcolor{sky}\textcolor{white}{Sky} &
        \cellcolor{road}\textcolor{white}{Road} &
        \cellcolor{person}\textcolor{white}{Person} &
        \cellcolor{car}\textcolor{white}{Car} &
        \cellcolor{streetlight}\textcolor{white}{Streetlight} &
        \cellcolor{trafficlight}\textcolor{white}{Traffic light}
    \end{tabularx}%
    \endgroup
}\\[1em]

\end{tabular}

    \vspace{-1em}
    \caption{\textbf{\MethodName-SDG static evaluation data.}
    RGB frames and reference semantic and instance annotations for
    the conditioning view and generated viewpoints.}
    \label{fig:lyra-static-eval-examples}
\end{figure}

\inlinesection{Evaluation data.}
\MethodName-SDG contains \num{25} manually curated diverse scenes held out from the training data.
To obtain high-quality reference annotations, we use an ensembled version of our SAM~\citep{kirillov:2023:sam}/SAM2~\citep{ravi:2025:saiv} instance preprocessing, combined with human review, refinement, and assignment of ADE20K semantic labels. The resulting labels and persistent identities provide references across all six generated trajectories. \Cref{fig:lyra-static-eval-examples} shows annotated examples.

\begin{figure}[t]
    \centering
    
\scriptsize
\sffamily
\setlength{\tabcolsep}{0pt}
\newcommand{\imgwidth}{0.139}

\begin{tabular}{
>{\centering\arraybackslash} m{0.027\textwidth}
>{\centering\arraybackslash} m{\imgwidth\textwidth}
>{\centering\arraybackslash} m{\imgwidth\textwidth}
>{\centering\arraybackslash} m{\imgwidth\textwidth}
>{\centering\arraybackslash} m{\imgwidth\textwidth}
>{\centering\arraybackslash} m{\imgwidth\textwidth}
>{\centering\arraybackslash} m{\imgwidth\textwidth}
>{\centering\arraybackslash} m{\imgwidth\textwidth}
}

& \textbf{Input View}
& \multicolumn{6}{c}{\rule[0.5ex]{4.5cm}{0.4pt}~\textbf{Generated\phantom{p}Views}~\rule[0.5ex]{4.5cm}{0.4pt}} \\

\rotatebox[origin=lB]{90}{\hspace{-0.25em}{RGB}}
& \includegraphics[width=\linewidth]{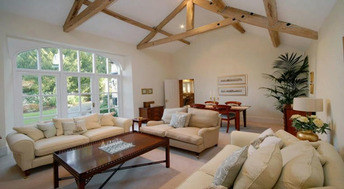}
& \includegraphics[width=\linewidth]{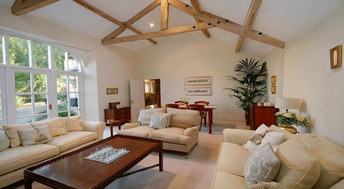}
& \includegraphics[width=\linewidth]{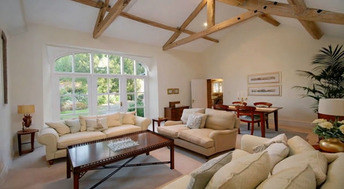}
& \includegraphics[width=\linewidth]{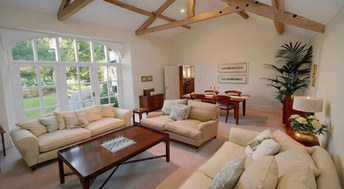}
& \includegraphics[width=\linewidth]{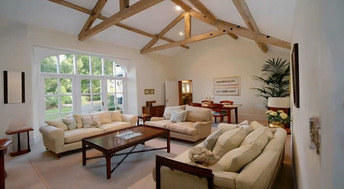}
& \includegraphics[width=\linewidth]{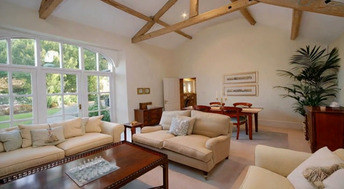}
& \includegraphics[width=\linewidth]{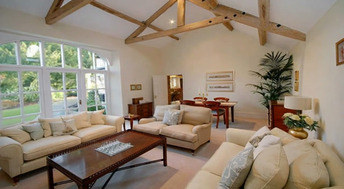}\\[-2.25pt]

\rotatebox[origin=lB]{90}{\hspace{-0.25em}{Instance}}
& \includegraphics[width=\linewidth]{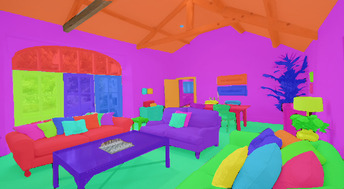}
& \includegraphics[width=\linewidth]{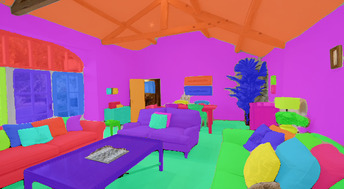}
& \includegraphics[width=\linewidth]{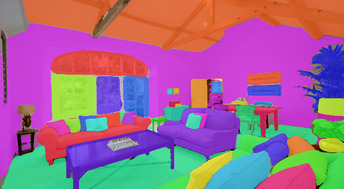}
& \includegraphics[width=\linewidth]{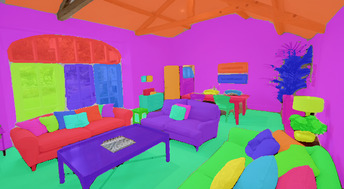}
& \includegraphics[width=\linewidth]{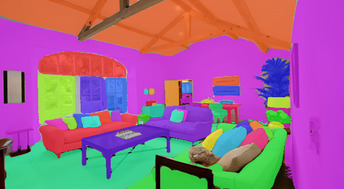}
& \includegraphics[width=\linewidth]{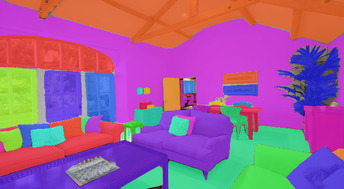}
& \includegraphics[width=\linewidth]{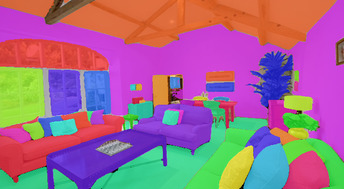}\\[-2.25pt]

\rotatebox[origin=lB]{90}{\hspace{-0.25em}{Semantic}}
& \includegraphics[width=\linewidth]{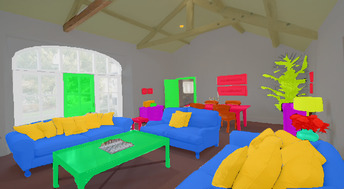}
& \includegraphics[width=\linewidth]{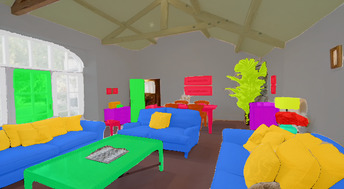}
& \includegraphics[width=\linewidth]{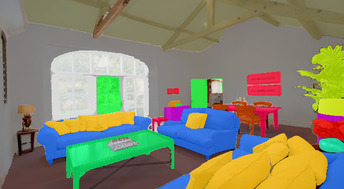}
& \includegraphics[width=\linewidth]{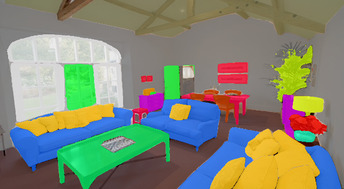}
& \includegraphics[width=\linewidth]{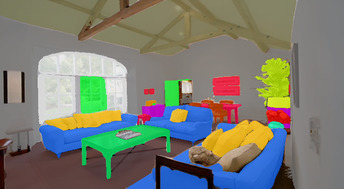}
& \includegraphics[width=\linewidth]{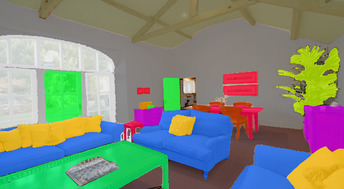}
& \includegraphics[width=\linewidth]{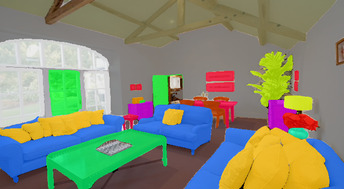}\\[-2.5pt]
& \multicolumn{7}{c}{%
    \begingroup
    \scriptsize
    \sffamily
    \setlength{\tabcolsep}{0pt}%
    \renewcommand{\arraystretch}{0.9}%
    \begin{tabularx}{0.973\textwidth}{*{10}{>{\centering\arraybackslash}X}}
        \cellcolor{wall}\textcolor{white}{Wall} &
        \cellcolor{floor}\textcolor{white}{Floor} &
        \cellcolor{ceiling}\textcolor{white}{Ceiling} &
        \cellcolor{windowpane}\textcolor{white}{Window} &
        \cellcolor{cabinet}\textcolor{white}{Cabinet} &
        \cellcolor{door}\textcolor{white}{Door} &
        \cellcolor{table}\textcolor{white}{Table} &
        \cellcolor{plant}\textcolor{white}{Plant} &
        \cellcolor{chair}\textcolor{white}{Chair} &
        \cellcolor{painting}\textcolor{white}{Painting} \\[-0.25pt]
        \cellcolor{sofa}\textcolor{white}{Sofa} &
        \cellcolor{mirror}\textcolor{white}{Mirror} &
        \cellcolor{lamp}\textcolor{white}{Lamp} &
        \cellcolor{cushion}\textcolor{white}{Cushion} &
        \cellcolor{box}\textcolor{white}{Box} &
        \cellcolor{coffeetable}\textcolor{white}{Coffee table} &
        \cellcolor{flower}\textcolor{white}{Flower} &
        \cellcolor{book}\textcolor{white}{Book} &
        \cellcolor{light}\textcolor{white}{Light} &
        \cellcolor{vase}\textcolor{white}{Vase}
    \end{tabularx}%
    \endgroup
}\\[1em]

\rotatebox[origin=lB]{90}{\hspace{-0.25em}{RGB}}
& \includegraphics[width=\linewidth]{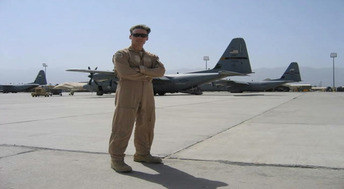}
& \includegraphics[width=\linewidth]{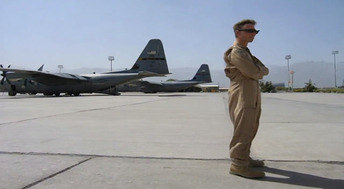}
& \includegraphics[width=\linewidth]{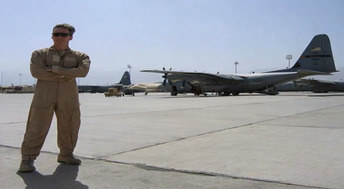}
& \includegraphics[width=\linewidth]{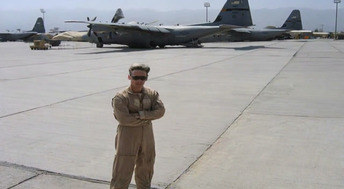}
& \includegraphics[width=\linewidth]{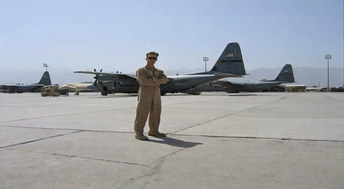}
& \includegraphics[width=\linewidth]{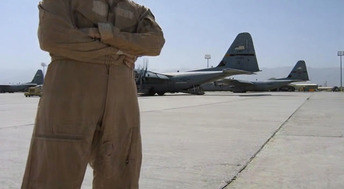}
& \includegraphics[width=\linewidth]{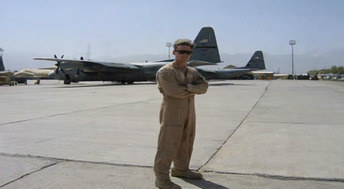}\\[-2.25pt]

\rotatebox[origin=lB]{90}{\hspace{-0.25em}{Instance}}
& \includegraphics[width=\linewidth]{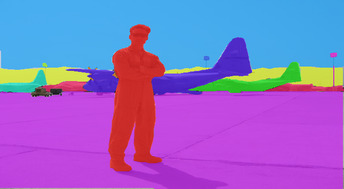}
& \includegraphics[width=\linewidth]{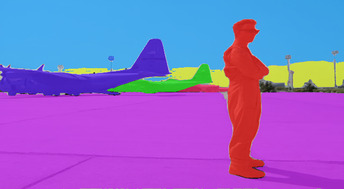}
& \includegraphics[width=\linewidth]{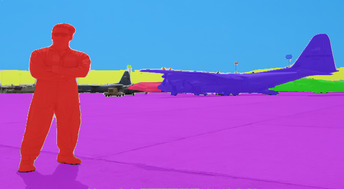}
& \includegraphics[width=\linewidth]{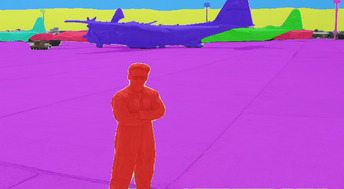}
& \includegraphics[width=\linewidth]{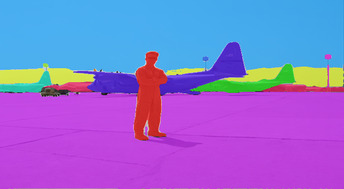}
& \includegraphics[width=\linewidth]{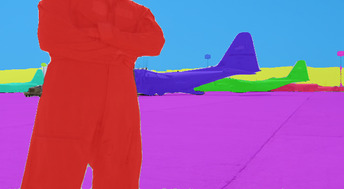}
& \includegraphics[width=\linewidth]{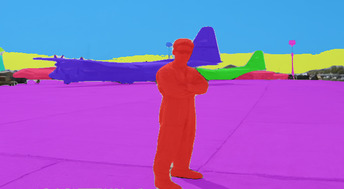}\\[-2.25pt]

\rotatebox[origin=lB]{90}{\hspace{-0.25em}{Semantic}}
& \includegraphics[width=\linewidth]{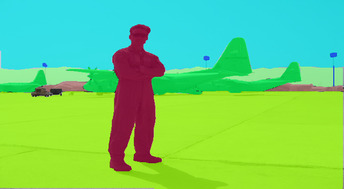}
& \includegraphics[width=\linewidth]{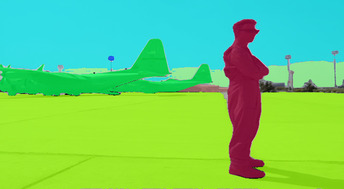}
& \includegraphics[width=\linewidth]{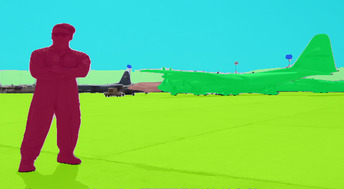}
& \includegraphics[width=\linewidth]{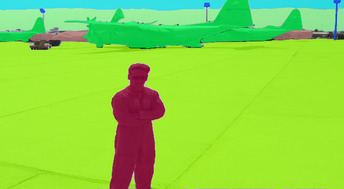}
& \includegraphics[width=\linewidth]{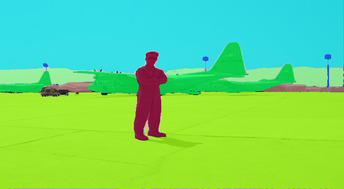}
& \includegraphics[width=\linewidth]{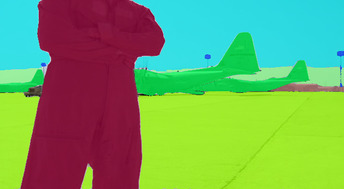}
& \includegraphics[width=\linewidth]{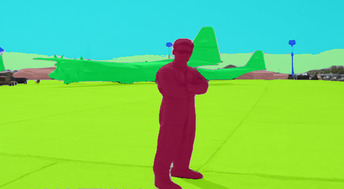}\\[-2.5pt]
& \multicolumn{7}{c}{%
    \begingroup
    \scriptsize
    \sffamily
    \setlength{\tabcolsep}{0pt}%
    \renewcommand{\arraystretch}{0.9}%
    \begin{tabularx}{0.973\textwidth}{*{7}{>{\centering\arraybackslash}X}}
        \cellcolor{building}\textcolor{white}{Building} &
        \cellcolor{sky}\textcolor{white}{Sky} &
        \cellcolor{person}\textcolor{white}{Person} &
        \cellcolor{mountain}\textcolor{white}{Mountain} &
        \cellcolor{runway}\textcolor{white}{Runway} &
        \cellcolor{streetlight}\textcolor{white}{Streetlight} &
        \cellcolor{airplane}\textcolor{white}{Airplane}
    \end{tabularx}%
    \endgroup
}\\[1em]

\end{tabular}

    \vspace{-1em}
    \caption{\textbf{\MethodName-ADE20K evaluation data.}
    Real conditioning images and generated viewpoints with semantic
    and instance annotations propagated from official ADE20K labels.}
    \label{fig:ade20k-eval-examples}
\end{figure}

\MethodName-ADE20K uses \num{25} diverse ADE20K validation images~\citep{zhou:2019:sust}, selected for scene-category diversity. Gen3C~\citep{ren:2025:3wvg} expands each image into six \num{121}-frame videos. We use a SAM2 ensemble to propagate the original semantic and instance ground-truth labels through all camera trajectories, followed by human review and refinement. \Cref{fig:ade20k-eval-examples} shows example scenes.

\subsection{Dynamic Dataset}

\begin{figure}[t]
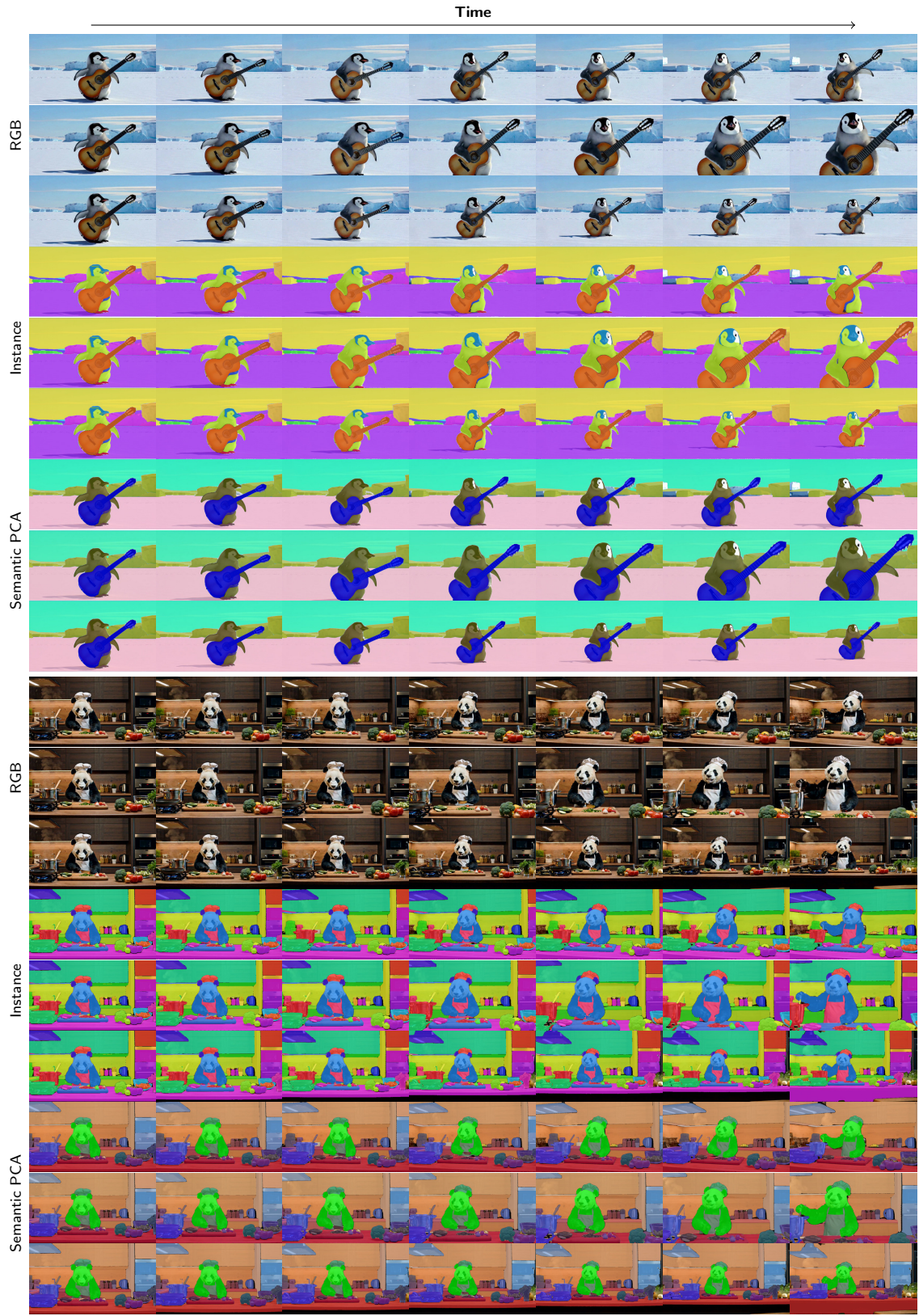

    \centering
    \scriptsize
\sffamily
\setlength{\tabcolsep}{0pt}
\newcommand{\lyradynamictrainimgwidth}{0.125}
\newcommand{\lyradynamictrainimage}[4]{%
    \includegraphics[width=\linewidth]{figures/supplement/lyra_train_dynamic/#1/traj#2_t#3_#4}%
}
\newcommand{\lyradynamictrainrow}[3]{%
\lyradynamictrainimage{#1}{#2}{000}{#3}
& \lyradynamictrainimage{#1}{#2}{020}{#3}
& \lyradynamictrainimage{#1}{#2}{040}{#3}
& \lyradynamictrainimage{#1}{#2}{060}{#3}
& \lyradynamictrainimage{#1}{#2}{080}{#3}
& \lyradynamictrainimage{#1}{#2}{100}{#3}
& \lyradynamictrainimage{#1}{#2}{120}{#3}
}
\newcommand{\lyradynamicmodality}[1]{%
    \smash{\raisebox{-30pt}{\rotatebox[origin=c]{90}{#1}}}%
}

\begin{tabular}{
>{\centering\arraybackslash} m{0.027\textwidth}
>{\centering\arraybackslash} m{\lyradynamictrainimgwidth\textwidth}
>{\centering\arraybackslash} m{\lyradynamictrainimgwidth\textwidth}
>{\centering\arraybackslash} m{\lyradynamictrainimgwidth\textwidth}
>{\centering\arraybackslash} m{\lyradynamictrainimgwidth\textwidth}
>{\centering\arraybackslash} m{\lyradynamictrainimgwidth\textwidth}
>{\centering\arraybackslash} m{\lyradynamictrainimgwidth\textwidth}
>{\centering\arraybackslash} m{\lyradynamictrainimgwidth\textwidth}
}
& \multicolumn{7}{c}{%
    \tikz[baseline=-0.5ex]{\draw[->, line width=0.4pt] (0,0) -- node[above=1pt] {\textbf{Time}} (12.5cm,0);}%
} \\
\lyradynamicmodality{RGB} & \lyradynamictrainrow{0_0_eafb4a8ecedd4667ac92464e871e35f1}{01}{rgb} \\[-2.25pt]
& \lyradynamictrainrow{0_0_eafb4a8ecedd4667ac92464e871e35f1}{04}{rgb} \\[-2.25pt]
& \lyradynamictrainrow{0_0_eafb4a8ecedd4667ac92464e871e35f1}{03}{rgb} \\[-2.25pt]
\lyradynamicmodality{Instance} & \lyradynamictrainrow{0_0_eafb4a8ecedd4667ac92464e871e35f1}{01}{instance_supervision} \\[-2.25pt]
& \lyradynamictrainrow{0_0_eafb4a8ecedd4667ac92464e871e35f1}{04}{instance_supervision} \\[-2.25pt]
& \lyradynamictrainrow{0_0_eafb4a8ecedd4667ac92464e871e35f1}{03}{instance_supervision} \\[-2.25pt]
\lyradynamicmodality{Semantic PCA} & \lyradynamictrainrow{0_0_eafb4a8ecedd4667ac92464e871e35f1}{01}{semantic_supervision_pca} \\[-2.25pt]
& \lyradynamictrainrow{0_0_eafb4a8ecedd4667ac92464e871e35f1}{04}{semantic_supervision_pca} \\[-2.25pt]
& \lyradynamictrainrow{0_0_eafb4a8ecedd4667ac92464e871e35f1}{03}{semantic_supervision_pca} \\[6pt]
\lyradynamicmodality{RGB} & \lyradynamictrainrow{0_0_2da9632f71d849328d2c96b186128711}{01}{rgb} \\[-2.25pt]
& \lyradynamictrainrow{0_0_2da9632f71d849328d2c96b186128711}{04}{rgb} \\[-2.25pt]
& \lyradynamictrainrow{0_0_2da9632f71d849328d2c96b186128711}{03}{rgb} \\[-2.25pt]
\lyradynamicmodality{Instance} & \lyradynamictrainrow{0_0_2da9632f71d849328d2c96b186128711}{01}{instance_supervision} \\[-2.25pt]
& \lyradynamictrainrow{0_0_2da9632f71d849328d2c96b186128711}{04}{instance_supervision} \\[-2.25pt]
& \lyradynamictrainrow{0_0_2da9632f71d849328d2c96b186128711}{03}{instance_supervision} \\[-2.25pt]
\lyradynamicmodality{Semantic PCA} & \lyradynamictrainrow{0_0_2da9632f71d849328d2c96b186128711}{01}{semantic_supervision_pca} \\[-2.25pt]
& \lyradynamictrainrow{0_0_2da9632f71d849328d2c96b186128711}{04}{semantic_supervision_pca} \\[-2.25pt]
& \lyradynamictrainrow{0_0_2da9632f71d849328d2c96b186128711}{03}{semantic_supervision_pca} \\[-2.25pt]
\end{tabular}

    \vspace{-0.5em}
    \caption{\textbf{\MethodName-SDG dynamic training data.}
    RGB frames, instance tracks, and PCA visualizations of semantic
    teacher features across viewpoints and time.}
    \label{fig:lyra-dynamic-train-examples}
\end{figure}

\inlinesection{Training data.}
The dynamic Lyra-SDG corpus comprises \num{7378} video-conditioned scenes.
Generation from the original and temporally reversed input video
provides two complementary six-trajectory batches, each containing
\num{121} synchronized frames per trajectory.
Targets are constructed within each batch, keeping their independently
initialized track IDs separate.
\Cref{fig:lyra-dynamic-train-examples} illustrates training scenes across viewpoints and time.

\begin{figure}[t]
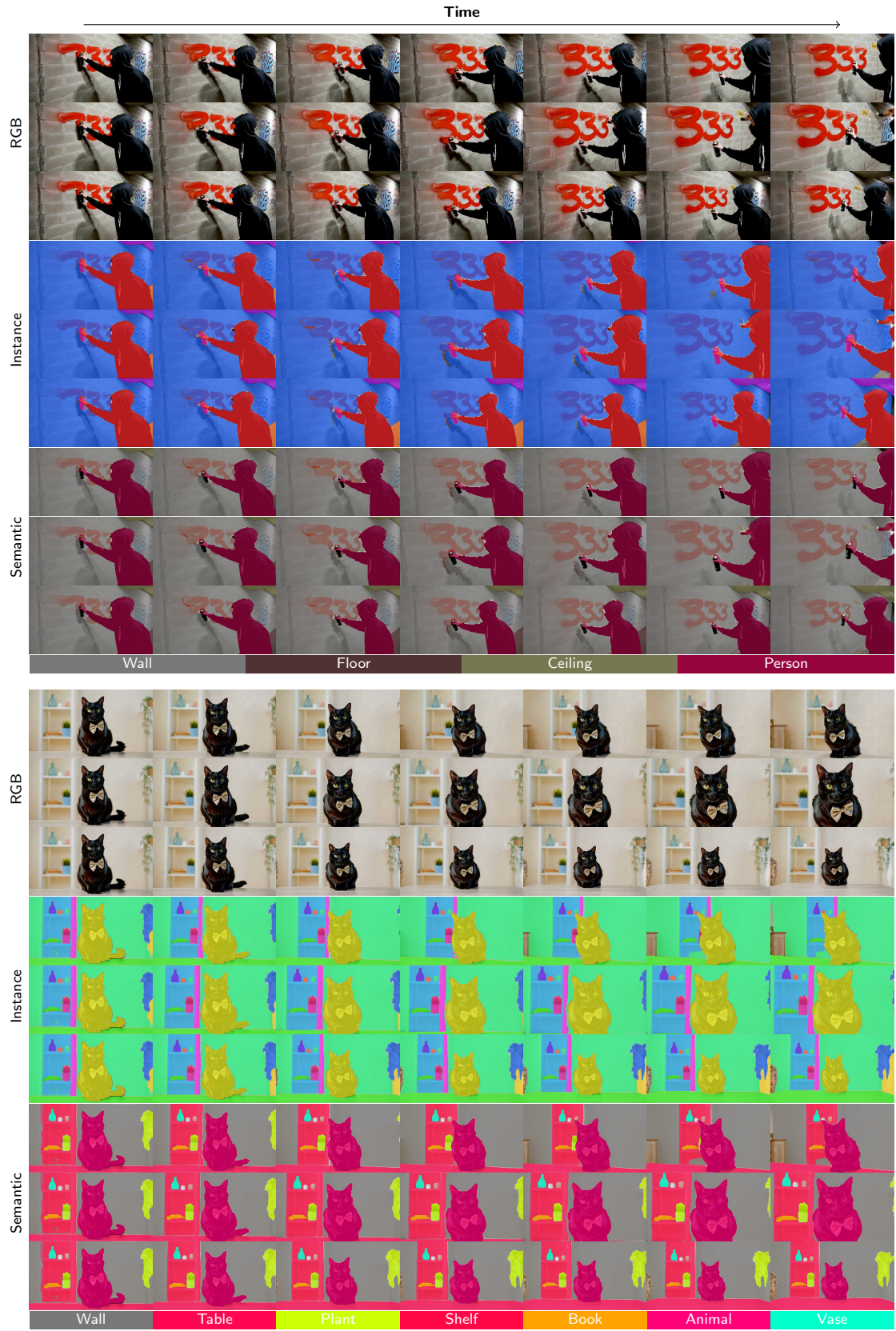

    \centering
    \scriptsize
\sffamily
\setlength{\tabcolsep}{0pt}
\newcommand{\lyradynamicevalimgwidth}{0.123}
\newcommand{\lyradynamicevalimage}[4]{%
    \includegraphics[width=\linewidth]{figures/supplement/lyra_eval_dynamic/#1/traj#2_t#3_#4}%
}
\newcommand{\lyradynamicevalrow}[3]{%
\lyradynamicevalimage{#1}{#2}{000}{#3}
& \lyradynamicevalimage{#1}{#2}{020}{#3}
& \lyradynamicevalimage{#1}{#2}{040}{#3}
& \lyradynamicevalimage{#1}{#2}{060}{#3}
& \lyradynamicevalimage{#1}{#2}{080}{#3}
& \lyradynamicevalimage{#1}{#2}{100}{#3}
& \lyradynamicevalimage{#1}{#2}{120}{#3}
}
\newcommand{\lyradynamicevalmodality}[1]{%
    \smash{\raisebox{-30pt}{\rotatebox[origin=c]{90}{#1}}}%
}

\begin{tabular}{
>{\centering\arraybackslash} m{0.027\textwidth}
>{\centering\arraybackslash} m{\lyradynamicevalimgwidth\textwidth}
>{\centering\arraybackslash} m{\lyradynamicevalimgwidth\textwidth}
>{\centering\arraybackslash} m{\lyradynamicevalimgwidth\textwidth}
>{\centering\arraybackslash} m{\lyradynamicevalimgwidth\textwidth}
>{\centering\arraybackslash} m{\lyradynamicevalimgwidth\textwidth}
>{\centering\arraybackslash} m{\lyradynamicevalimgwidth\textwidth}
>{\centering\arraybackslash} m{\lyradynamicevalimgwidth\textwidth}
}
& \multicolumn{7}{c}{%
    \tikz[baseline=-0.5ex]{\draw[->, line width=0.4pt] (0,0) -- node[above=1pt] {\textbf{Time}} (12.5cm,0);}%
} \\
\lyradynamicevalmodality{RGB} & \lyradynamicevalrow{0_0_0c9c1959ab874fbcb32b219781501458}{01}{rgb} \\[-2.25pt]
& \lyradynamicevalrow{0_0_0c9c1959ab874fbcb32b219781501458}{04}{rgb} \\[-2.25pt]
& \lyradynamicevalrow{0_0_0c9c1959ab874fbcb32b219781501458}{03}{rgb} \\[-2.25pt]
\lyradynamicevalmodality{Instance} & \lyradynamicevalrow{0_0_0c9c1959ab874fbcb32b219781501458}{01}{instance_gt} \\[-2.25pt]
& \lyradynamicevalrow{0_0_0c9c1959ab874fbcb32b219781501458}{04}{instance_gt} \\[-2.25pt]
& \lyradynamicevalrow{0_0_0c9c1959ab874fbcb32b219781501458}{03}{instance_gt} \\[-2.25pt]
\lyradynamicevalmodality{Semantic} & \lyradynamicevalrow{0_0_0c9c1959ab874fbcb32b219781501458}{01}{semantic_gt} \\[-2.25pt]
& \lyradynamicevalrow{0_0_0c9c1959ab874fbcb32b219781501458}{04}{semantic_gt} \\[-2.25pt]
& \lyradynamicevalrow{0_0_0c9c1959ab874fbcb32b219781501458}{03}{semantic_gt} \\[-2.5pt]
& \multicolumn{7}{c}{%
    \begingroup
    \scriptsize
    \sffamily
    \setlength{\tabcolsep}{0pt}%
    \renewcommand{\arraystretch}{0.9}%
    \begin{tabularx}{0.861\textwidth}{*{4}{>{\centering\arraybackslash}X}}
        \cellcolor{wall}\textcolor{white}{Wall} &
        \cellcolor{floor}\textcolor{white}{Floor} &
        \cellcolor{ceiling}\textcolor{white}{Ceiling} &
        \cellcolor{person}\textcolor{white}{Person}
    \end{tabularx}%
    \endgroup
}\\[1em]

\lyradynamicevalmodality{RGB} & \lyradynamicevalrow{0_0_da11c4a113124350bacf409e03830923}{01}{rgb} \\[-2.25pt]
& \lyradynamicevalrow{0_0_da11c4a113124350bacf409e03830923}{04}{rgb} \\[-2.25pt]
& \lyradynamicevalrow{0_0_da11c4a113124350bacf409e03830923}{03}{rgb} \\[-2.25pt]
\lyradynamicevalmodality{Instance} & \lyradynamicevalrow{0_0_da11c4a113124350bacf409e03830923}{01}{instance_gt} \\[-2.25pt]
& \lyradynamicevalrow{0_0_da11c4a113124350bacf409e03830923}{04}{instance_gt} \\[-2.25pt]
& \lyradynamicevalrow{0_0_da11c4a113124350bacf409e03830923}{03}{instance_gt} \\[-2.25pt]
\lyradynamicevalmodality{Semantic} & \lyradynamicevalrow{0_0_da11c4a113124350bacf409e03830923}{01}{semantic_gt} \\[-2.25pt]
& \lyradynamicevalrow{0_0_da11c4a113124350bacf409e03830923}{04}{semantic_gt} \\[-2.25pt]
& \lyradynamicevalrow{0_0_da11c4a113124350bacf409e03830923}{03}{semantic_gt} \\[-2.5pt]
& \multicolumn{7}{c}{%
    \begingroup
    \scriptsize
    \sffamily
    \setlength{\tabcolsep}{0pt}%
    \renewcommand{\arraystretch}{0.9}%
    \begin{tabularx}{0.861\textwidth}{*{7}{>{\centering\arraybackslash}X}}
        \cellcolor{wall}\textcolor{white}{Wall} &
        \cellcolor{table}\textcolor{white}{Table} &
        \cellcolor{plant}\textcolor{white}{Plant} &
        \cellcolor{shelf}\textcolor{white}{Shelf} &
        \cellcolor{book}\textcolor{white}{Book} &
        \cellcolor{animal}\textcolor{white}{Animal} &
        \cellcolor{vase}\textcolor{white}{Vase}
    \end{tabularx}%
    \endgroup
}\\[1em]
\end{tabular}

    \vspace{-1.5em}
    \caption{\textbf{\MethodName-SDG dynamic evaluation data.}
    RGB frames with reference semantic and instance annotations
    from dynamic evaluation scenes.}
    \label{fig:lyra-dynamic-eval-examples}
\end{figure}

\inlinesection{Evaluation data.}
\MethodName-SDG Dynamic contains \num{25} diverse curated scenes excluded from the training set. Annotation follows the static setup. \Cref{fig:lyra-dynamic-eval-examples} visualizes annotated examples.

\section{Related Work}

Below, we discuss further related work not covered in the main paper.

\inlinesection{Dynamic representations}

Dynamic semantic and instance representations have likewise focused on captured
video, spanning per-sequence optimization and feed-forward reconstruction.
4D LangSplat~\citep{li:2025:4l4l} fits language features to deformable 4D
Gaussians, whereas TRASE~\citep{li:2026:tt4s} learns an instance field from
per-frame masks. Related methods distill foundation-model features or temporal
cues into scene-specific dynamic representations
\citep{labe:2024:dd3g,fiebelman:2025:44le,zhou:2025:fbam,
ji:2024:sa4g,hu:2025:s4ds,wu:2026:cifd,jiang:2026:digs}.
Feed-forward approaches amortize language-aligned dynamic reconstruction
\citep{wu:2025:44lg,qiu:2026:ssal}, while IGGT4D~\citep{zou:2026:is4i}
jointly predicts streaming geometry and persistent instances.

\end{document}